\newcommand\SEKIusersusepackages{
\usepackage{graphicx}
\usepackage{tikz}
\usepackage{tikz-qtree}
\usetikzlibrary{matrix,decorations.pathreplacing}
\input headerhot
\def\proof         {\relax}
\def\proofqed      {\relax}
\def\proofparsep   {\relax}
\def\proofparsepqed{\relax}
\def\proofshort    {\relax}
\newcommand\Proofof{Proof of}
\renewenvironment{proof}[1]{\yestop\begin{sloppypar}\noindent
{\bf\Proofof\ {#1}}\\}{\end{sloppypar}}
\renewenvironment{proofqed}[1]
{\begin{sloppypar}\def\fooqed{#1}\noindent{\bf\Proofof\ \fooqed}}
{\QEDbf\fooqed\end{sloppypar}}
\renewenvironment{proofparsep}[1]{\parindent=0pt\begin
{sloppypar}\noindent{\bf\Proofof\ {#1}}\nopagebreak\par}
{\end{sloppypar}}
\renewenvironment{proofparsepqed}[1]{\parindent=0pt\begin
{sloppypar}\def\fooqed{#1}\noindent{\bf\Proofof\ \fooqed}\nopagebreak\par}
{\nopagebreak\QEDbf\fooqed\end{sloppypar}}
\renewenvironment{proofshort}[1]{\begin{sloppypar}\noindent
{\bf\Proofof\ {#1}:}}{\end{sloppypar}}

\mathcommand\ident[1]{\mathsf{#1}}
\newcommand\plussymbol  {\ident{+}}
\newcommand\minussymbol {\ident{-}}
\newcommand\dividesymbol{\ident{/}}
\newcommand\timessymbol {\ident{*}}

\newcommand\eins    {\ident{1}}
\newcommand\nat     {\ident{nat}}
\newcommand\posnat  {\ident{posnat}}
\newcommand\zeronat {\ident{zero}}
\newcommand\ord     {\ident{ord}}
\newcommand\ORD     {\ident{ORD}}
\newcommand\lists   {\ident{list}}
\newcommand\set     {\ident{set}}
\newcommand\stream  {\ident{stream}}
\newcommand\term    {\ident{term}}
\newcommand\nonemptylists{\ident{nonemptylist}}
\newcommand\emptylists   {\ident{emptylist}}
\newcommand\bool    {\ident{bool}}
\newcommand\intsort {\ident{int}}
\newcommand\funsort [2]{{#1\math\rightarrow#2}}
\newcommand\pairsort[2]{{#1\math{\!\times\!}#2}}
\newcommand\sumsort [2]{{#1\math{+}#2}}
\newcommand\ortreesort{\ident{or\mbox-tree}}
\newcommand\andtreesort{\ident{and\mbox-tree}}

\newcommand\naturalssymbol{\ident{naturals}}
\newcommand\gensymsymbol{\ident{gensym}}
\mathcommand\mbpsymbol{\ident{m\hspace{-0.055em}b\hspace{-0.045em}p}}
\newcommand\appsymbol   {\ident a}
\newcommand\csymbol     {\ident c}
\newcommand\esymbol     {\ident e}
\newcommand\fsymbol     {\ident f}
\newcommand\gsymbol     {\ident g}
\newcommand\hsymbol     {\ident h}
\newcommand\ksymbol     {\ident k}
\newcommand\psymbol     {\ident p}
\newcommand\ssymbol     {\ident s}
\newcommand\Everysymbol {\ident{Every}}
\newcommand\Permsymbol {\ident{Perm}}
\newcommand\RExistssymbol{\ident{Rexists}}
\newcommand\invertsymbol{\ident{invert}}
\newcommand\invsymbol{\ident{inv}}
\newcommand\abssymbol   {\ident{abs}}
\newcommand\cnssymbol   {\ident{cons}}
\mathcommand\cnsindexsymbol[1]{\ident{cons}_{#1}}
\newcommand\carsymbol   {\ident{car}}
\newcommand\cardsymbol  {\ident{card}}
\newcommand\cdrsymbol   {\ident{cdr}}
\newcommand\lengthsymbol{\ident{length}}
\newcommand\sizesymbol{\ident{size}}
\newcommand\dlsymbol    {\ident{dl}}
\newcommand\dloncesymbol{\ident{delfirst}}
\newcommand\rcsymbol    {\ident{rc}}
\newcommand\brsymbol    {\ident{br}}
\newcommand\revtailsymbol{\ident{revtail}}
\newcommand\revsymbol{\ident{rev}}
\newcommand\appendsymbol {\ident{append}}
\newcommand\zeropredicatesymbol{\ident{zerop}}
\newcommand\eqsymbol        {\ident{eq}}
\newcommand\ifthensymbol    {\mbox{\ident{If{}Then}}}
\newcommand\ifthenelsesymbol{\mbox{\ident{If{}ThenElse}}}
\mathcommand\eqindexsymbol        [1]{\eqsymbol        _{#1}}
\mathcommand\ifthenindexsymbol    [1]{\ifthensymbol    _{#1}}
\mathcommand\ifthenelseindexsymbol[1]{\ifthenelsesymbol_{#1}}
\newcommand\orsymbol    {\ident{or}}
\newcommand\andsymbol   {\ident{and}}
\newcommand\leqsymbol   {\ident{leq}}
\newcommand\lessymbol   {\ident{less}}
\newcommand\lexlessymbol{\ident{lexless}}
\newcommand\lexlimlessymbol{\ident{lexlimless}}
\newcommand\lexsymbol   {\ident{lex}}
\newcommand\acksymbol   {\ident{ack}}
\newcommand\switchsymbol{\ident{switch}}
\newcommand\swatchsymbol{\ident{swatch}}
\newcommand\diveinssymbol{\ident{div1}}
\newcommand\divzweisymbol{\ident{div2}}
\newcommand\divrestsymbol{\ident{divrest}}
\newcommand\diveinstailsymbol{\ident{div1tail}}
\newcommand\divzweitailsymbol{\ident{div2tail}}
\newcommand\remsymbol{\ident{rem}}
\newcommand\divsymbol{\ident{div}}

\newcommand\turingmachinesymbol{\ident T}
\newcommand\terminatespsymbol  {\ident{terminatesp}}
\newcommand\statesymbol        {\ident{state}}
\newcommand\cmdsymbol          {\ident{cmd}}
\newcommand\nthsymbol          {\ident{nth}}
\newcommand\doublesymbol       {\ident{double}}

\newcommand\ppsymbol           {\ident{p}}
\newcommand\qpsymbol           {\ident{q}}
\newcommand\Epsymbol           {\ident{E}}
\newcommand\Ppsymbol           {\ident{P}}
\newcommand\Qpsymbol           {\ident{Q}}
\newcommand\Marriessymbol      {\ident{Marries}}
\newcommand\Lovessymbol        {\ident{Loves}}
\newcommand\StolenBysymbol     {\ident{StolenBy}}
\newcommand\Humansymbol        {\ident{Human}}
\newcommand\Evensymbol         {\ident{Even}}
\newcommand\Oddsymbol          {\ident{Odd}}
\newcommand\Primesymbol        {\ident{Prime}}
\newcommand\EveryPairsymbol   {\ident{EveryPair}}
\newcommand\Givesymbol         {\ident{Give}}
\newcommand\Fathersymbol       {\ident{Father}}
\newcommand\Elephantpsymbol    {\ident{Elephant}}
\newcommand\Flowerpsymbol    {\ident{Flower}}
\newcommand\Germanpsymbol      {\ident{German}}
\newcommand\Bicyclepsymbol     {\ident{Bicycle}}
\newcommand\Hugepsymbol        {\ident{Huge}}
\newcommand\Animalpsymbol      {\ident{Animal}}
\newcommand\Malepsymbol        {\ident{Male}}
\newcommand\Boypsymbol         {\ident{Boy}}
\newcommand\Girlpsymbol        {\ident{Girl}}
\newcommand\Femalepsymbol      {\ident{Female}}
\newcommand\Roundpsymbol       {\ident{Round}}
\newcommand\Quadrangularpsymbol{\ident{Quadrangular}}
\newcommand\Metpsymbol         {\ident{Met}}
\newcommand\Kissedpsymbol      {\ident{Kissed}}
\newcommand\Bishopsymbol       {\ident{Bishop}}
\newcommand\mindexsymbol[1]{\existsvari w{#1}}

\newcommand\nonnegpsymbol      {\ident{nonnegp}}
\newcommand\wellsymbol         {\ident{well}}
\newcommand\welltailsymbol     {\ident{welltail}}
\newcommand\varsymbol          {\ident{var}}
\newcommand\aritysymbol        {\ident{arity}}

\newcommand\whilesymbol        {\ident{while}}

\newcommand\nullsymbol         {\ident{null}}
\newcommand\hdsymbol           {\ident{hd}}
\newcommand\tlsymbol           {\ident{tl}}
\newcommand\insymbol           {\ident{in}}
\newcommand\applysymbol        {\ident{app}}
\newcommand\termsymbol         {\ident{term}}
\newcommand\russellsymbol      {\ident{russell}}
\newcommand\sqrtindordsymbol[1]{\ident{sqrtio#1}}
\mathcommand\tightim{\longrightarrow}
\mathcommand\im{\ \tightim\ }
\mathcommand\rs{\:\rulesugar\:\:}
\mathcommand\rulesugar{\longleftarrow}

\mathcommand\doublepp[1]      {\doublesymbol   \beginargs{#1}\allargs}
\mathcommand\aritypp[1]      {\aritysymbol   \beginargs{#1}\allargs}
\mathcommand\lengthpp[1]      {\lengthsymbol   \beginargs{#1}\allargs}
\mathcommand\sizepp[1]      {\sizesymbol   \beginargs{#1}\allargs}
\mathcommand\wellpp[1]      {\wellsymbol   \beginargs{#1}\allargs}
\mathcommand\welltailpp[1]      {\welltailsymbol   \beginargs{#1}\allargs}
\mathcommand\varpp[1]      {\varsymbol   \beginargs{#1}\allargs}
\mathcommand\rempp[2]    {\remsymbol\beginargs{#1}\separgs{#2}\allargs}
\mathcommand\divpp[2]    {\divsymbol\beginargs{#1}\separgs{#2}\allargs}
\mathcommand\divrestpp[2]    {\divrestsymbol\beginargs{#1}\separgs{#2}\allargs}
\mathcommand\diveinspp[2]    {\diveinssymbol\beginargs{#1}\separgs{#2}\allargs}
\mathcommand\divzweipp[3]    {\divzweisymbol\beginargs{#1}\separgs{#2}
\separgs{#3}\allargs}
\mathcommand\diveinstailpp[4]    {\diveinstailsymbol\beginargs{#1}\separgs{#2}
\separgs{#3}\separgs{#4}\allargs}
\mathcommand\divzweitailpp[6]    {\divzweitailsymbol\beginargs{#1}\separgs{#2}
\separgs{#3}\separgs{#4}\separgs{#5}\separgs{#6}\allargs}
\mathcommand\mbppp[2]         {\mbpsymbol   \beginargs{#1}\separgs{#2}\allargs}
\mathcommand\revpp[1]     
{\revsymbol\beginargs{#1}\allargs}
\mathcommand\revppi[2]     
{\revsymbol^{#1}\beginargs{#2}\allargs}
\mathcommand\revtailpp[2]     
{\revtailsymbol\beginargs{#1}\separgs{#2}\allargs}
\mathcommand\revtailppi[3]
{\revtailsymbol^{#1}\beginargs{#2}\separgs{#3}\allargs}
\mathcommand\Permpp[2]     
{\Permsymbol\beginargs{#1}\separgs{#2}\allargs}
\mathcommand\Permppi[3]
{\Permsymbol^{#1}\beginargs{#2}\separgs{#3}\allargs}
\mathcommand\appendpp[2]      
{\appendsymbol \beginargs{#1}\separgs{#2}\allargs}
\mathcommand\appendppi[3]      
{\appendsymbol^{#1}\beginargs{#2}\separgs{#3}\allargs}
\mathcommand\Everypp[2]      
{\Everysymbol \beginargs{#1}\separgs{#2}\allargs}
\mathcommand\RExistspp[1]      
{\RExistssymbol \beginargs{#1}\allargs}
\mathcommand\appendlongpp[2]      
{\appendsymbol\left(\begin{array}{@{}l@{}}{#1}\separgs\\{#2}\end{array}\right)}
\mathcommand\cnspp[2]         {\cnssymbol   \beginargs{#1}\separgs{#2}\allargs}
\mathcommand\cnsppi[3]       {\cnssymbol^{#1}\beginargs{#2}\separgs{#3}\allargs}
\mathcommand\cnsindexpp[3]
{\cnsindexsymbol{#1}\beginargs{#2}\separgs{#3}\allargs}
\mathcommand\dlpp[2]          {\dlsymbol    \beginargs{#1}\separgs{#2}\allargs}
\mathcommand\dloncepp[2]      {\dloncesymbol\beginargs{#1}\separgs{#2}\allargs}
\mathcommand\dlonceppi[3]{\dloncesymbol^{#1}\beginargs{#2}\separgs{#3}\allargs}
\mathcommand\rcpp[2]          {\rcsymbol    \beginargs{#1}\separgs{#2}\allargs}
\mathcommand\brpp[2]          {\brsymbol    \beginargs{#1}\separgs{#2}\allargs}
\mathcommand\orpp[2]          {\orsymbol    \beginargs{#1}\separgs{#2}\allargs}
\mathcommand\andpp[2]         {\andsymbol   \beginargs{#1}\separgs{#2}\allargs}
\mathcommand\shortcnspp[2]    {\csymbol     \beginargs{#1}\separgs{#2}\allargs}
\mathcommand\tightshortcnspp[2]
{\csymbol\beginargs{#1}\tightsepargs{#2}\allargs}
\mathcommand\spp[1]           {\ssymbol     \beginargs{#1}\allargs}
\mathcommand\sppiterated[2]   {\ssymbol^{#1}\beginargs{#2}\allargs}
\mathcommand\sqrtindordpp[3]
                       {\sqrtindordsymbol{#1}\beginargs{#2}\separgs{#3}\allargs}
\mathcommand\ppp[1]           {\psymbol     \beginargs{#1}\allargs}
\mathcommand\pppiterated[2]   {\psymbol^{#1}\beginargs{#2}\allargs}
\mathcommand\zeropp           {\ident 0}
\mathcommand\Julietpp         {\ident{Juliet}}
\mathcommand\Romeopp          {\ident{Romeo}}
\mathcommand\Ipp              {\ident I}
\mathcommand\onepp            {\ident1}
\mathcommand\twopp            {\ident2}
\mathcommand\threepp          {\ident3}
\mathcommand\invertpp[1]      {\invertsymbol\beginargs{#1}\allargs}
\mathcommand\invpp[1]         {\invsymbol\beginargs{#1}\allargs}
\mathcommand\abspp[1]         {\abssymbol\beginargs{#1}\allargs}
\mathcommand\naturalspp[1]    {\naturalssymbol\beginargs{#1}\allargs}
\mathcommand\gensympp[1]      {\gensymsymbol\beginargs{#1}\allargs}
\mathcommand\nilpp            {\ident{nil}}
\mathcommand\falsepp          {\ident{false}}
\mathcommand\truepp           {\ident{true}}
\mathcommand\FALSEpp          {\ident{FALSE}}
\mathcommand\TRUEpp           {\ident{TRUE}}
\mathcommand\UNDEFpp          {\ident{UNDEF}}
\mathcommand\weirdppp         {\ident{weirdp}}
\mathcommand\ambigppp         {\ident{ambigp}}
\mathcommand\zeropredicatepp[1]{\zeropredicatesymbol\beginargs{#1}\allargs}
\mathcommand\cppeins       [1]{\csymbol     \beginargs{#1}\allargs}
\mathcommand\cppzwei       [2]{\csymbol\beginargs{#1}\separgs{#2}\allargs}
\mathcommand\eppeins       [1]{\esymbol     \beginargs{#1}\allargs}
\mathcommand\fppeins       [1]{\fsymbol     \beginargs{#1}\allargs}
\mathcommand\fppeinsindex  [2]{\fsymbol_{#1}\beginargs{#2}\allargs}
\mathcommand\fppeinsiterated[2]{\fsymbol^{#1}\beginargs{#2}\allargs}
\mathcommand\gppeins       [1]{\gsymbol     \beginargs{#1}\allargs}
\mathcommand\gppzwei       [2]{\gsymbol     \beginargs{#1}\separgs{#2}\allargs}
\mathcommand\hppeins       [1]{\hsymbol     \beginargs{#1}\allargs}
\mathcommand\kppeins       [1]{\ksymbol     \beginargs{#1}\allargs}
\mathcommand\appzero          {\ident a}
\mathcommand\bppzero          {\ident b}
\mathcommand\cppzero          {\ident c}
\mathcommand\dppzero          {\ident d}
\mathcommand\eppzero          {\ident e}
\mathcommand\eqindexpp[3]{\eqindexsymbol{#1}\beginargs{#2}\separgs{#3}\allargs}
\mathcommand\ifthenindexpp
[3]{\ifthenindexsymbol{#1}\beginargs{#2}\separgs{#3}\allargs}
\mathcommand\ifthenelseindexpp
[4]{\ifthenelseindexsymbol{#1}\beginargs{#2}\separgs{#3}\separgs{#4}\allargs}
\mathcommand\eqpp[2]{\eqsymbol\beginargs{#1}\separgs{#2}\allargs}
\mathcommand\leqpp[2]{\leqsymbol\beginargs{#1}\separgs{#2}\allargs}
\mathcommand\lespp[2]{\lessymbol\beginargs{#1}\separgs{#2}\allargs}
\mathcommand\lexlespp[2]{\lexlessymbol\beginargs{#1}\separgs{#2}\allargs}
\mathcommand\lexlimlespp[3]
               {\lexlimlessymbol\beginargs{#1}\separgs{#2}\separgs{#3}\allargs}
\mathcommand\lexpp[3]{\lexsymbol\beginargs{#1}\separgs{#2}\separgs{#3}\allargs}
\mathcommand\ackpp[2]{\acksymbol\beginargs{#1}\separgs{#2}\allargs}
\mathcommand\switchpp[1]{\switchsymbol\beginargs{#1}\allargs}
\mathcommand\swatchpp[1]{\swatchsymbol\beginargs{#1}\allargs}
\mathcommand\whilepp[2]{\whilesymbol\beginargs{#1}\separgs{#2}\allargs}
\mathcommand\nullpp[1]{\nullsymbol\beginargs{#1}\allargs}
\mathcommand\nullppiterated[2]{\nullsymbol^{#1}\beginargs{#2}\allargs}
\mathcommand\hdpp[1]{\hdsymbol\beginargs{#1}\allargs}
\mathcommand\hdppiterated[2]{\hdsymbol^{#1}\beginargs{#2}\allargs}
\mathcommand\carpp[1]{\carsymbol\beginargs{#1}\allargs}
\mathcommand\cdrpp[1]{\cdrsymbol\beginargs{#1}\allargs}
\mathcommand\tlpp[1]{\tlsymbol\beginargs{#1}\allargs}
\mathcommand\tlppiterated[2]{\tlsymbol^{#1}\beginargs{#2}\allargs}
\mathcommand\inpp[2]{\insymbol\beginargs{#1}\separgs{#2}\allargs}
\mathcommand\inppiterated[3]{\insymbol^{#1}\beginargs{#2}\separgs{#3}\allargs}
\mathcommand\applypp[2]{\applysymbol\beginargs{#1}\separgs{#2}\allargs}
\mathcommand\applyppiterated
[3]{\applysymbol^{#1}\beginargs{#2}\separgs{#3}\allargs}
\mathcommand\termpp[2]{\termsymbol\beginargs{#1}\separgs{#2}\allargs}
\mathcommand\setpp[1]{\set\beginargs{#1}\allargs}
\mathcommand\russellpp[1]{\russellsymbol\beginargs{#1}\allargs}

\mathcommand\Tpp[6]{\turingmachinesymbol\beginargs{#1}\separgs{#2}\separgs
{#3}\separgs{#4}\separgs{#5}\separgs{#6}\allargs}
\mathcommand\Tppseven[7]{\turingmachinesymbol\beginargs{#1}\separgs{#2}\separgs
{#3}\separgs{#4}\separgs{#5}\separgs{#6}\separgs{#7}\allargs}
\mathcommand\foreverppp[6]{\ident{foreverp}\beginargs{#1}\separgs{#2}\separgs
{#3}\separgs{#4}\separgs{#5}\separgs{#6}\allargs}
\mathcommand\terminatesppp[6]{\terminatespsymbol\beginargs{#1}\separgs
{#2}\separgs{#3}\separgs{#4}\separgs{#5}\separgs{#6}\allargs}
\mathcommand\terminatespppone[1]{\terminatespsymbol \beginargs{#1}\allargs}
\mathcommand\statepp
[3]{\statesymbol\beginargs{#1}\separgs{#2}\separgs{#3}\allargs}
\mathcommand\tightstatepp
[3]{\statesymbol\beginargs{#1}\tightsepargs{#2}\tightsepargs{#3}\allargs}
\mathcommand\cmdpp
[3]{\cmdsymbol  \beginargs{#1}\separgs{#2}\separgs{#3}\allargs}
\mathcommand\tightcmdpp
[3]{\cmdsymbol  \beginargs{#1}\tightsepargs{#2}\tightsepargs{#3}\allargs}
\mathcommand\stoppp           {\ident{stop}}
\mathcommand\leftpp           {\ident{left}}
\mathcommand\rightpp          {\ident{right}}
\mathcommand\nthpp         [2]{\nthsymbol  \beginargs{#1}\separgs{#2}\allargs}
\mathcommand\pppp          [1]{\ppsymbol\beginargs{#1}            \allargs}
\mathcommand\qppp          [2]{\qpsymbol\beginargs{#1}\separgs{#2}\allargs}
\mathcommand\Eppp          [1]{\Epsymbol\beginargs{#1}            \allargs}
\mathcommand\Epppzwei      [2]{\Epsymbol\beginargs{#1}\separgs{#2}\allargs}
\mathcommand\Pppp          [1]{\Ppsymbol\beginargs{#1}            \allargs}
\mathcommand\Ppppeinsindex [2]{\Ppsymbol_{#1}\beginargs{#2}\allargs}
\mathcommand\Ppppdrei      
[3]{\Ppsymbol\beginargs{#1}\separgs{#2}\separgs{#3}\allargs}
\mathcommand\Ppppvier
[4]{\Ppsymbol\beginargs{#1}\separgs{#2}\separgs{#3}\separgs{#4}\allargs}
\mathcommand\Qppp          [2]{\Qpsymbol\beginargs{#1}\separgs{#2}\allargs}
\mathcommand\Qpppeins      [1]{\Qpsymbol\beginargs{#1}\allargs}
\mathcommand\Qpppeinsindex [2]{\Qpsymbol_{#1}\beginargs{#2}\allargs}
\mathcommand\Qpppdrei      
[3]{\Qpsymbol\beginargs{#1}\separgs{#2}\separgs{#3}\allargs}
\mathcommand\Fatherpp      [2]{\Fathersymbol\beginargs{#1}\separgs{#2}\allargs}
\mathcommand\Marriespp     [2]{\Marriessymbol\beginargs{#1}\separgs{#2}\allargs}
\mathcommand\Lovespp       [2]{\Lovessymbol\beginargs{#1}\separgs{#2}\allargs}
\mathcommand\StolenBypp    [2]
{\StolenBysymbol\beginargs{#1}\separgs{#2}\allargs}
\mathcommand\Humanpp       [1]{\Humansymbol\beginargs{#1}\allargs}
\mathcommand\Evenpp        [1]{\Evensymbol\beginargs{#1}\allargs}
\mathcommand\Evenppi       [2]{\Evensymbol^{#1}\beginargs{#2}\allargs}
\mathcommand\Oddpp         [1]{\Oddsymbol\beginargs{#1}\allargs}
\mathcommand\Primepp       [1]{\Primesymbol\beginargs{#1}\allargs}
\mathcommand\EveryPairpp  [2]{\EveryPairsymbol\beginargs{#1}\separgs
{#2}\allargs}
\mathcommand\mindexppeins  [2]{\mindexsymbol{#1}\beginargs{#2}\allargs}
\mathcommand\Givepp        [3]{\Givesymbol
\beginargs{#1}\separgs{#2}\separgs{#3}\allargs}
\mathcommand\mindexppzwei  [3]{\mindexsymbol
{#1}\beginargs{#2}\separgs{#3}\allargs}
\mathcommand\mindexppdrei  [4]{\mindexsymbol
{#1}\beginargs{#2}\separgs{#3}\separgs{#4}\allargs}

\mathcommand\nonnegppp     [1]{\nonnegpsymbol\beginargs{#1}\allargs}

\mathcommand\anonymouscsymbol{c}
\mathcommand\anonymouscindexsymbol[1]{\anonymouscsymbol_{#1}}
\mathcommand\anonymousfsymbol{f}
\mathcommand\anonymouscpp
[2]{\anonymouscsymbol\beginargs{#1}\separgs\ldots\separgs{#2}\allargs}
\mathcommand\anonymouscindexpp
[3]{\anonymouscindexsymbol{#1}\beginargs{#2}\separgs\ldots\separgs{#3}\allargs}
\mathcommand\anonymousfpp
[2]{\anonymousfsymbol\beginargs{#1}\separgs\ldots\separgs{#2}\allargs}
\mathcommand\coerceindexpp[3]{[#3]_{#1}^{#2}}

\mathcommand\Elephantppp    [1]{\Elephantpsymbol\beginargs{#1}\allargs}
\mathcommand\Flowerppp      [1]{\Flowerpsymbol  \beginargs{#1}\allargs}
\mathcommand\Bicycleppp     [1]{\Bicyclepsymbol \beginargs{#1}\allargs}
\mathcommand\Germanppp      [1]{\Germanpsymbol  \beginargs{#1}\allargs}
\mathcommand\Hugeppp        [1]{\Hugepsymbol    \beginargs{#1}\allargs}
\mathcommand\Animalppp      [1]{\Animalpsymbol  \beginargs{#1}\allargs}
\mathcommand\Maleppp        [1]{\Malepsymbol    \beginargs{#1}\allargs}
\mathcommand\Boyppp         [1]{\Boypsymbol     \beginargs{#1}\allargs}
\mathcommand\Girlppp        [1]{\Girlpsymbol    \beginargs{#1}\allargs}
\mathcommand\Femaleppp      [1]{\Femalepsymbol  \beginargs{#1}\allargs}
\mathcommand\Roundppp       [1]{\Roundpsymbol   \beginargs{#1}\allargs}
\mathcommand\Bishoppp       [1]{\Bishopsymbol   \beginargs{#1}\allargs}
\mathcommand\Quadrangularppp[1]{\Quadrangularpsymbol  \beginargs{#1}\allargs}
\mathcommand\Kissedppp[2]{\Kissedpsymbol\beginargs{#1}\separgs{#2}\allargs}
\mathcommand\Metppp[2]   {\Metpsymbol   \beginargs{#1}\separgs{#2}\allargs}

\newcommand\bound     {{\rm bound}}
\newcommand\free      {{\rm free}}

\mathcommand\Vtripleindex[3]{\V\!_{{#1},\,{#2},\,{#3}}}
\mathcommand\Vdoubleindex[2]{\V\!_{{#1},\,{#2}}}
\mathcommand\Vsingleindex[1]{\V\!_{{#1}}}

\newcommand\exrelation{\mbox{\math\Gammaoffont-}relation}
\newcommand\unrelation{\mbox{\math\Deltaoffont-}relation}
\mathcommand\Erel[1]{\Gammaoffont\!_{#1}}
\mathcommand\Urel[1]{\Deltaoffont_{#1}}

\newcommand\exRsub  {\math R-sub\-sti\-tu\-tion  on \nolinebreak\Vsome}
\newcommand\aexRsub {an \math R-sub\-sti\-tu\-tion  on \nolinebreak\Vsome}
\newcommand\exRsubs {\math R-sub\-sti\-tu\-tions on \nolinebreak\Vsome}
\newcommand\ExRSubs {\math R-Sub\-sti\-tu\-tions on \nolinebreak\Vsome}
\newcommand\Rsub    {\math R-sub\-sti\-tu\-tion}
\newcommand\aRsub   {an \math R-sub\-sti\-tu\-tion}
\newcommand\RSub    {\math R-Sub\-sti\-tu\-tion}
\newcommand\Rsubs   {\math R-sub\-sti\-tu\-tions}
\newcommand\RSubs   {\math R-Sub\-sti\-tu\-tions}
\newcommand\aRprimesub{an \math{R'}-sub\-sti\-tu\-tion}
\newcommand\aexRprimesub{\aRprimesub\ on \nolinebreak\Vsome}
\newcommand\exemptysetsub
{\math\emptyset-sub\-sti\-tu\-tion  on \nolinebreak\Vsome}

\newcommand\quasiexRsubs
{\Rsubs\ on \nolinebreak\math{\Vsome\cup\Vsall}}
\newcommand\aquasiexRsub
{\aRsub\ on \nolinebreak\math{\Vsome\cup\Vsall}}
\newcommand\isquasiexistential{is a substitution on \nlbmath{\Vsome\cup\Vsall}}

\newcommand\aexRsuboptquasi
{\aRsub\ on \nolinebreak\math{\Vsome\,\opt{\cup\,\Vsall}}}

\newcommand\strongsigmaupdate{\math\sigma-update}

\newcommand\strongexvalof[2]{\pair{#1}{#2}-valu\-a\-tion}
\newcommand\astrongexvalof[2]{an\/ \strongexvalof{#1}{#2}}
\newcommand\strongexRval {\strongexvalof\salgebra R}
\newcommand\StrongexRVal {\pair\salgebra R-Valu\-a\-tion}
\newcommand\strongexRvals{\pair\salgebra R-valu\-a\-tions}
\newcommand\StrongexRVals{\pair\salgebra R-Valu\-a\-tions}
\newcommand\astrongexRval{an \strongexRval}
\newcommand\strongexRprimeval{\pair\salgebra{R'}-valu\-a\-tion}
\newcommand\astrongexRprimeval{an\/ \strongexRprimeval}

\newcommand\cc{choice-condition}
\newcommand\CC{Choice-Condition}
\newcommand\Cc{Choice-condition}
\newcommand\vc{vari\-able-con\-di\-tion}
\newcommand\VC{Vari\-able-Con\-di\-tion}
\newcommand\Vc{Vari\-able-con\-di\-tion}
\newcommand\semanticobject{\mbox{\math\Sigmaoffont-}struc\-ture}

\newcommand\validity[2]{\pair{#1}{#2}-validity}
\newcommand\Validity[2]{\pair{#1}{#2}-Validity}
\newcommand\valid[2]{\pair{#1}{#2}-valid}
\newcommand\validstrongtrip[3]{\trip{#1}{#2}{#3}-valid}
\newcommand\validitystrongtrip[3]{\trip{#1}{#2}{#3}-validity}

\newcommand\stronglyreduces[3]{\pair{#1}{#2}-reduces}
\mathcommand\theRprimefromstrongtoweak{
  \inparenthesesinlinetight{
     \domres\id{\Vwall\cup\Vsome\setminus\RAN\varsigma}
     \nottight{\nottight\uplus}
     \reverserelation\varsigma
  }
  \nottight{\circ}
  \ranres
    {\transclosureinline R}
    {\Vwall\cup\Vsome\setminus\RAN\varsigma}
  \nottight{\nottight{\nottight{\uplus}}}
  \Vsome\tighttimes\Vsall
}

\mathcommand\deltaminus{\delta^-}
\mathcommand\deltaplus{\delta^+}
\mathcommand\deltaplusplus{\delta^{+^+}}
\mathcommand\deltastar{\delta^*}
\mathcommand\deltastarstar{\delta^{*^*}}
\newcommand\varihelper[1]{free \discretionary
{\mbox{\math{#1}-vari-}}{\mbox{able}}{\mbox{\math{#1}-variable}}}
\newcommand\Varihelper[1]{Free \discretionary
{\mbox{\math{#1}-vari-}}{\mbox{able}}{\mbox{\math{#1}-variable}}}
\newcommand\fev {\varihelper\gamma}
\newcommand\fuv {\varihelper\delta}
\newcommand\wfuv{\varihelper{\delta^-}}
\newcommand\sfuv{\varihelper{\delta^+}}
\newcommand\Fev {\Varihelper\gamma}
\newcommand\Fuv {\Varihelper\delta}
\newcommand\Wfuv{\Varihelper{\delta^-}}
\newcommand\Sfuv{\Varihelper{\delta^+}}
\newcommand\FUV {Free \math{\delta  }-Variable}
\newcommand\SFUV {Free \math{\delta^+}-Variable}

\mathcommand\Vall     {\Vsingleindex\indexdelta         }
\mathcommand\Vwall    {\Vsingleindex\indexdeltaminu     }
\mathcommand\Vsall    {\Vsingleindex\indexdeltaplus     }
\mathcommand\Vgsome   {\Vsingleindex\indexgammaplus     }
\mathcommand\Vsome    {\Vsingleindex\indexgamma         }
\mathcommand\Vfree    {\Vsingleindex\indexfree          }
\mathcommand\Vbound   {\Vsingleindex\indexbound         }
\mathcommand\Vsomesall{\Vsingleindex\indexgammadeltaplus}

\mathapplycommand\VARall      {\VARsingleindex\indexdelta         }
\mathapplycommand\VARwall     {\VARsingleindex\indexdeltaminu     }
\mathapplycommand\VARsall     {\VARsingleindex\indexdeltaplus     }
\mathapplycommand\VARgsome    {\VARsingleindex\indexgammaplus     }
\mathapplycommand\VARsome     {\VARsingleindex\indexgamma         }
\mathapplycommand\VARfree     {\VARsingleindex\indexfree          }
\mathapplycommand\VARbound    {\VARsingleindex\indexbound         }
\mathapplycommand\VARsomesall {\VARsingleindex\indexgammadeltaplus}
\mathcommand\displayVARsall[1]{\VARsingleindex\indexdeltaplus
\!\!\!\:\left(\begin{array}{@{}c@{}}#1\end{array}\right)}

\mathcommand\rigidvari     [2]{#1_{#2}^\indexgammadeltaplus}
\mathcommand\existsvari    [2]{#1_{#2}^\indexgamma    }
\mathcommand\forallvari    [2]{#1_{#2}^\indexdelta    }
\mathcommand\freevari      [2]{#1_{#2}^\indexfree     }
\mathcommand\wforallvari   [2]{#1_{#2}^\indexdeltaminu}
\mathcommand\sforallvari   [2]{#1_{#2}^\indexdeltaplus}
\mathcommand\gexistsvari   [2]{#1_{#2}^\indexgammaplus}
\mathcommand\boundvari     [2]{#1_{#2}}
\mathcommand\vari          [2]{#1_{#2}}
\mathcommand\wforallvarilow[2]{#1_{#2}^
{\raisebox{-.82ex}{\math\indexdeltaminu}}}

\newcommand\indexhelper[1]{{\scriptscriptstyle#1\:\!\!}}
\newcommand\indexdeltaplus
{\indexhelper{\delta^{\raisebox{-.17ex}{\fvesf\hskip-0.14em +}}}}
\newcommand\indexdeltaminu
{\indexhelper{\delta^{\mbox{\fvesf\hskip-0.14em\rule[.2ex]{.7em}{.15ex}}}}}
\newcommand\indexgammaplus
{\indexhelper{\gamma^{\mbox{\fvesf\hskip-0.14em +}}}}
\newcommand\indexgammadeltaplus
{\indexhelper{\gamma\delta^{\raisebox{-.17ex}{\fvesf\hskip-0.14em +}}}}

\newcommand\indexdelta{\indexhelper\delta}
\newcommand\indexgamma{\indexhelper\gamma}
\newcommand\indexfree
{{\scriptscriptstyle\free}}
\newcommand\indexbound
{{\scriptscriptstyle\bound}}

\newcommand\Wellfsymb{\ident{Wellf}}
\mathapplycommand\Wellfpp{\Wellfsymb}

\mathcommand\beginargs{(}
\mathcommand\allargs  {)}
\mathcommand\separgs  {,\,}
\mathcommand\tightsepargs{,}

\mathcommand\minusppnoparentheses  [2]{{#1}\,\minussymbol\,{#2}}
\mathcommand\tightminusppnoparentheses  [2]{{#1}\minussymbol{#2}}
\mathcommand\divideppnoparentheses [2]{{#1}\,\dividesymbol\,{#2}}
\mathcommand\plusppnoparentheses   [2]{{#1}\,\plussymbol \,{#2}}
\mathcommand\plusppnoparenthesesi  [3]{{#2}\,\plussymbol^{#1}\,{#3}}
\mathcommand\tightplusppnoparentheses   [2]{{#1}\plussymbol{#2}}
\mathcommand\timesppnoparentheses  [2]{{#1}\,\timessymbol\,{#2}}
\mathcommand\undppnoparentheses    [2]{{#1}\und            {#2}}
\mathcommand\oderppnoparentheses   [2]{{#1}\oder           {#2}}
\mathcommand\impliesppnoparentheses[2]{{#1}\implies        {#2}}
\mathcommand\leqinfixppnoparentheses[2]{{#1}\,\tight\leq\,{#2}}
\mathcommand\geqinfixppnoparentheses[2]{{#1}\,\tight\geq\,{#2}}
\mathcommand\dividepp [2]{(\divideppnoparentheses {#1}{#2})}
\mathcommand\minuspp  [2]{(\minusppnoparentheses  {#1}{#2})}
\mathcommand\pluspp   [2]{(\plusppnoparentheses   {#1}{#2})}
\mathcommand\timespp  [2]{(\timesppnoparentheses  {#1}{#2})}
\mathcommand\undpp    [2]{(\undppnoparentheses    {#1}{#2})}
\mathcommand\oderpp   [2]{(\oderppnoparentheses   {#1}{#2})}
\mathcommand\impliespp[2]{(\impliesppnoparentheses{#1}{#2})}

\usepackage{amssymb}
\usepackage{named}
\bibliographystyle{named}
\def\citep{\cite}
\def\citet#1{\citeauthor{#1} \shortcite{#1}}
\newcommand\startcite{{\raise.2ex\hbox{[}}}
\newcommand\stopcite {\raise.2ex\hbox{]}}
\newcommand\citehelper[1]{\startcite #1\stopcite}
\newcommand\makeaciteoftwo[2]
{\citehelper{\citeauthor{#1}, \citeyear{#1}; \citeyear{#2}}}
\newcommand\makeacitetoftwo[2]
{\citeauthor{#1} \citehelper{\citeyear{#1}; \citeyear{#2}}}
\newcommand\makeaciteofthree[3]
{\citehelper{\citeauthor{#1}, \citeyear{#1}; \citeyear{#2}; \citeyear{#3}}}
\newcommand\makeaciteoffour[4]
{\citehelper{\citeauthor{#1}, \citeyear{#1}; \citeyear{#2}; \citeyear{#3};
\citeyear{#4}}}
\newcommand\makeacitetoffour[4]
{\citeauthor{#1} \citehelper{\citeyear{#1}; \citeyear{#2}; \citeyear{#3};
\citeyear{#4}}}
\newcommand\makeaciteoffive[5]
{\citehelper{\citeauthor{#1}, \citeyear{#1}; \citeyear{#2}; \citeyear{#3};
\citeyear{#4}; \citeyear{#5}}}
\newcommand\makeaciteofsix[6]
{\citehelper{\citeauthor{#1}, \citeyear{#1}; \citeyear{#2}; \citeyear{#3};
\citeyear{#4}; \citeyear{#5}; \citeyear{#6}}}
\newcommand\makeaciteofseven[7]
{\citehelper{\citeauthor{#1}, \citeyear{#1}; \citeyear{#2}; \citeyear{#3};
\citeyear{#4}; \citeyear{#5}; \citeyear{#6}; \citeyear{#7}}}
\newcommand\startacitewithnine[9]
{\startcite\citeauthor{#1}, \citeyear{#1}; 
\citeyear{#2}; \citeyear{#3}; \citeyear{#4}; \citeyear{#5};
\citeyear{#6}; \citeyear{#7}; \citeyear{#8}; \citeyear{#9};
}
\newcommand\stopacitewithone[1]{\citeyear{#1}\stopcite}
\newcommand\makeanextendedciteoftwo[4]
{\citehelper{\citeauthor{#2}, \citeyear{#2}, #1; \citeyear{#4}, #3}}
\newcommand\translationnotewithlongcite[3]{\cite[#1, #3]{#2}}

\newcommand\thispaper{this paper}

\mathcommand\tightdefeasibleantiimplies{\leftarrow}
\newcommand\defeasibleantiimplies
                              {\ \tightdefeasibleantiimplies  \penalty-2\relax\ }
}

\newcommand\SEKIREVIEWSTATUS{1}
\newcommand\SEKIREVIEWER
{\stolzenburgname
\\\Institute
\\E-mail: {\tt fstolzenburg@hs-harz.de}
\\WWW: \url{http://fstolzenburg.hs-harz.de}
}
\newcommand\SEKIDOCUMENTCLASS{0}
\newcommand\SEKIYEAR{2013}
\newcommand\SEKINUMBEROFYEAR{01}
\newcommand\SEKITYPE{0}
\newcommand\SEKIPUBLISHEDTYPE{0}
\title
{\poolename's Specificity Revised
}
\author
{\wirthname
\\\stolzenburgname
\\\Institute
\\\emailcp\\{\tt fstolzenburg@hs-harz.de}
}
\date
{\small\SEKIedition\\Preliminary Version, \Nov\,22, 2013}
\input seki-deckblatt-5
\newcommand\naught
{\rule{1.5cm}{0mm}}\newcommand\myborder
{\raisebox{-.9ex}{\rule{.5mm}{3ex}}}\newcommand\mybox
[2]{\makebox[#1][c]{\myborder\leftarrowfill#2\rightarrowfill\myborder}}%
{\mbox{} \mbox{}\rm}
\newif\ifaux
\IfFileExists{\jobname.aux}{\auxtrue}{\auxfalse}%
\usepackage{amssymb}

\usepackage[T1]{fontenc}

\mathchardef\Gammaoffont="7000
\mathchardef\Gamma="0100
\mathchardef\Deltaoffont="7001
\mathchardef\Delta="0101
\mathchardef\Thetaoffont="7002
\mathchardef\Theta="0102
\mathchardef\Lambdaoffont="7003
\mathchardef\Lambda="0103
\mathchardef\Xioffont="7004
\mathchardef\Xi="0104
\mathchardef\Pioffont="7005
\mathchardef\Pi="0105
\mathchardef\Sigmaoffont="7006
\mathchardef\Sigma="0106
\mathchardef\Upsilonoffont="7007
\mathchardef\Upsilon="0107
\mathchardef\Phioffont="7008
\mathchardef\Phi="0108
\mathchardef\Psioffont="7009
\mathchardef\Psi="0109
\mathchardef\Omegaoffont="700A
\mathchardef\Omega="010A
\mathchardef\itype="017B

\catcode`\@=11

\gdef\allowhyphens{\penalty\@M \hskip\z@skip}

\gdef\set@low@box#1{\setbox\tw@\hbox{,}\setbox\z@\hbox{#1}\dimen\z@\ht\z@
     \advance\dimen\z@ -\ht\tw@
     \setbox\z@\hbox{\lower\dimen\z@ \box\z@}\ht\z@\ht\tw@ \dp\z@\dp\tw@ }
\gdef\set@low@boxsingle#1{\setbox\tw@\hbox{\rm,}\setbox\z@\hbox{#1}\dimen\z@\ht\z@
     \advance\dimen\z@ -\ht\tw@
     \setbox\z@\hbox{\lower\dimen\z@ \box\z@}\ht\z@\ht\tw@ \dp\z@\dp\tw@ }

\gdef\@glqq{%
\ifhmode\edef\@SF{\spacefactor\the\spacefactor}%
\else\let\@SF\empty
\fi
\CheckFamily\font\fraknomath\ifSameFamily ``\relax
\else\CheckFamily\font\swab\ifSameFamily ``\relax
\else\leavevmode\set@low@box{''}\box\z@\kern-.04em\allowhyphens\@SF\relax
\fi\fi}
\gdef\glqq{\protect\@glqq\kern+.07em}
\gdef\@grqq{%
\ifhmode\edef\@SF{\spacefactor\the\spacefactor}%
\else\let\@SF\empty 
\fi 
\CheckFamily\font\fraknomath\ifSameFamily ''\relax
\else\CheckFamily\font\swab\ifSameFamily ''\relax
\else\kern+.07em``\kern.07em\@SF\relax
\fi\fi}
\gdef\grqq{\protect\@grqq}
\gdef\@glq{{\ifhmode \edef\@SF{\spacefactor\the\spacefactor}\else
     \let\@SF\empty \fi \leavevmode
     \set@low@boxsingle{'\/}\box\z@\kern-.04em\allowhyphens\@SF\relax}}
\gdef\glq{\protect\@glq\kern+.07em}
\gdef\@grq{\ifhmode \edef\@SF{\spacefactor\the\spacefactor}\else
     \let\@SF\empty \fi \kern-.0125em`\kern.07em\@SF\relax}
\gdef\grq{\protect\@grq}

\catcode`\@=12

\newcommand\closequotecommanospace{''\nolinebreak\hskip-0.23em,}
\newcommand\closequotecomma      {\closequotecommanospace\         \,}
\newcommand\closequotecommasmallextraspace{\closequotecommanospace\ \,\,}

\newcommand\closequotefullstopextraspace   
                                 {\closequotefullstopnospace\    \ \,}

\newcommand\closequotefullstopnospace
                                 {''\nolinebreak\hskip-0.20em\@.}

\newcommand\commanospace          {,\nolinebreak\hskip-0.23em}

\makeatletter
\if \@ptsize 0
   \newfont{\scriptscriptscriptgoth}{ygoth scaled 760}
   \newfont{\scriptscriptgoth}{ygoth scaled 833}
   \newfont{\scriptgoth}{ygoth scaled 912}
   \newfont{\gothnomath}{ygoth}
   \newfont{\Goth}{ygoth scaled \magstephalf}
   \newfont{\GOth}{ygoth scaled \magstep1}
   \newfont{\GOTh}{ygoth scaled \magstep2}
   \newfont{\GOTH}{ygoth scaled \magstep3}

   \newfont{\scriptscriptscriptswab}{yswab scaled 760}
   \newfont{\scriptscriptswab}{yswab scaled 833}
   \newfont{\scriptswab}{yswab scaled 912}
   \newfont{\swab}{yswab}
   \newfont{\Swab}{yswab scaled \magstephalf}
   \newfont{\SWab}{yswab scaled \magstep1}
   \newfont{\SWAb}{yswab scaled \magstep2}
   \newfont{\SWAB}{yswab scaled \magstep3}

   \newfont{\scriptscriptscriptfrak}{yfrak scaled 760}
   \newfont{\scriptscriptfrak}{yfrak scaled 833}
   \newfont{\scriptfrak}{yfrak scaled 912}
   \newfont{\fraknomath}{yfrak}
   \newfont{\Frak}{yfrak scaled \magstephalf}
   \newfont{\FRak}{yfrak scaled \magstep1}
   \newfont{\FRAk}{yfrak scaled \magstep2}
   \newfont{\FRAK}{yfrak scaled \magstep3}

   \newfont{\init}{yinit}
   \newfont{\Init}{yinit scaled \magstephalf}
   \newfont{\INit}{yinit scaled \magstep1}
   \newfont{\INIt}{yinit scaled \magstep2}
   \newfont{\INIT}{yinit scaled \magstep3}
\fi
\if \@ptsize 1
   \newfont{\scriptscriptscriptgoth}{ygoth scaled 833}
   \newfont{\scriptscriptgoth}{ygoth scaled 912}
   \newfont{\scriptgoth}{ygoth}
   \newfont{\gothnomath}{ygoth scaled \magstephalf}
   \newfont{\Goth}{ygoth scaled \magstep1}
   \newfont{\GOth}{ygoth scaled \magstep2}
   \newfont{\GOTh}{ygoth scaled \magstep3}
   \newfont{\GOTH}{ygoth scaled \magstep4}

   \newfont{\scriptscriptscriptswab}{yswab scaled 833}
   \newfont{\scriptscriptswab}{yswab scaled 912}
   \newfont{\scriptswab}{yswab}
   \newfont{\swab}{yswab scaled \magstephalf}
   \newfont{\Swab}{yswab scaled \magstep1}
   \newfont{\SWab}{yswab scaled \magstep2}
   \newfont{\SWAb}{yswab scaled \magstep3}
   \newfont{\SWAB}{yswab scaled \magstep4}

   \newfont{\scriptscriptscriptfrak}{yfrak scaled 833}
   \newfont{\scriptscriptfrak}{yfrak scaled 912}
   \newfont{\scriptfrak}{yfrak}
   \newfont{\fraknomath}{yfrak scaled \magstephalf}
   \newfont{\Frak}{yfrak scaled \magstep1}
   \newfont{\FRak}{yfrak scaled \magstep2}
   \newfont{\FRAk}{yfrak scaled \magstep3}
   \newfont{\FRAK}{yfrak scaled \magstep4}

   \newfont{\init}{yinit scaled \magstephalf}
   \newfont{\Init}{yinit scaled \magstep1}
   \newfont{\INit}{yinit scaled \magstep2}
   \newfont{\INIt}{yinit scaled \magstep3}
   \newfont{\INIT}{yinit scaled \magstep4}
\fi
\if \@ptsize 2
   \newfont{\scriptscriptscriptgoth}{ygoth scaled 912}
   \newfont{\scriptscriptgoth}{ygoth}
   \newfont{\scriptgoth}{ygoth scaled \magstephalf}
   \newfont{\gothnomath}{ygoth scaled \magstep1}
   \newfont{\Goth}{ygoth scaled \magstep2}
   \newfont{\GOth}{ygoth scaled \magstep3}
   \newfont{\GOTh}{ygoth scaled \magstep4}
   \newfont{\GOTH}{ygoth scaled \magstep5}

   \newfont{\scriptscriptscriptswab}{yswab scaled 912}
   \newfont{\scriptscriptswab}{yswab}
   \newfont{\scriptswab}{yswab scaled \magstephalf}
   \newfont{\swab}{yswab scaled \magstep1}
   \newfont{\Swab}{yswab scaled \magstep2}
   \newfont{\SWab}{yswab scaled \magstep3}
   \newfont{\SWAb}{yswab scaled \magstep4}
   \newfont{\SWAB}{yswab scaled \magstep5}

   \newfont{\scriptscriptscriptfrak}{yfrak scaled 833}
   \newfont{\scriptscriptfrak}{yfrak}
   \newfont{\scriptfrak}{yfrak scaled \magstephalf}
   \newfont{\fraknomath}{yfrak scaled \magstep1}
   \newfont{\Frak}{yfrak scaled \magstep2}
   \newfont{\FRak}{yfrak scaled \magstep3}
   \newfont{\FRAk}{yfrak scaled \magstep4}
   \newfont{\FRAK}{yfrak scaled \magstep5}

   \newfont{\init}{yinit scaled \magstep1}
   \newfont{\Init}{yinit scaled \magstep2}
   \newfont{\INit}{yinit scaled \magstep3}
   \newfont{\INIt}{yinit scaled \magstep4}
   \newfont{\INIT}{yinit scaled \magstep5}
\fi

\newcommand{\mscriptscriptfrak}      [1]{\mbox{\scriptscriptscriptfrak#1}}
\newcommand{\mscriptfrak}            [1]{\mbox{\scriptscriptscriptfrak#1}}

\newcommand{\mfrak}[1]{\mbox{\fraknomath#1}}

\makeatother

\newif\ifSameFamily
\def\CheckFamily#1#2{\GetFamilyName{#1}\ArgOne
        \GetFamilyName{#2}\ArgTwo
        \ifx\ArgOne\ArgTwo\SameFamilytrue\else\SameFamilyfalse\fi}
\def\GetFamilyName#1{\edef\Tempa{#1}\def\Tempb{#1}\ifx\Tempa\Tempb
        \edef\Tempa{\fontname#1}\fi
        \edef\Tempa{\Tempa\space}%
        \expandafter\iGetFamilyName\Tempa\\}
\def\iGetFamilyName#1 #2\\#3{\def#3{#1}}
\def\DefFontName#1#2{{\escapechar-1\expandafter\expandafter\expandafter
        \iDefFontName\expandafter{\csname#2\endcsname}%
        \xdef#1{\expandafter\string\Tempa}}}
\def\iDefFontName{\def\Tempa}

\newcommand\unprotectedae
{\CheckFamily\font\fraknomath\ifSameFamily *a\else
 \CheckFamily\font\swab\ifSameFamily\char'212\else\"a\fi\fi}
\newcommand\unprotectedoe
{\CheckFamily\font\fraknomath\ifSameFamily 
*o\else\CheckFamily\font\swab\ifSameFamily\char'232\else\"o\fi\fi}

\DefFontName\eccclarge{eccc1200}
\DefFontName\eccc{eccc1000}
\DefFontName\ecccsmall{eccc0900}
\DefFontName\ecccfootnotesize{eccc0800}

\newcommand\unprotectedes
{\CheckFamily\font\fraknomath\ifSameFamily\char'215\else
\CheckFamily\font\swab\ifSameFamily\char'215\else  
s\fi\fi}

\newcommand\unprotectedmyparagraphsymbol
{\CheckFamily\font\fraknomath\ifSameFamily 
\char'244\else\CheckFamily\font\swab\ifSameFamily
\char'244\else\S\fi\fi}

\renewcommand\ae{\protect\unprotectedae}
\renewcommand\oe{\protect\unprotectedoe}

\newcommand\es  {\protect\unprotectedes}

\newcommand\myparagraphsymbol{\protect\unprotectedmyparagraphsymbol}

\def\mathfrak#1{%
\mathchoice
{{\mfrak{#1}}}
{{\mfrak{#1}}}
{{\mscriptfrak{#1}}}
{{\mscriptscriptfrak{#1}}}
}

\newcommand\namefont{}

\usepackage{url}

\newcommand\majorfootroom{\raisebox{-1.9ex}{\rule{0ex}{.5ex}}}

\newcommand\footroom{\raisebox{-1.5ex}{\rule{0ex}{.5ex}}}

\newcommand\smallfootroom{\raisebox{-1.0ex}{\rule{0ex}{3ex}}}

\newcommand\majorheadroom{\rule{0ex}{3.2ex}}

\newcommand\claus   {Clau\es}
\newcommand\david   {{\namefont David}}

\newcommand\peter   {Peter}

\newcommand\fermatbirthyear 
{160?}

\newcommand\secondincompletenesstheorem
{second \incompletenesstheorem}
\newcommand\secondIncompletenessTheorem
{Second \IncompletenessTheorem}

\newcommand\firstincompletenesstheorem
{first \incompletenesstheorem}
\newcommand\firstIncompletenessTheorem
{First \IncompletenessTheorem}
\newcommand\incompletenesstheorem{incompleteness theorem}
\newcommand\IncompletenessTheorem{Incompleteness Theorem}

\newcommand\kleene          {{\namefont Kleene}}

\newcommand\lambert         {{\namefont Lambert}}

\newcommand\lambertdiagram  {\lambert\ diagram}

\newcommand\lukasiewicz     {{\namefont \L uka\-sie\-wicz}}

\newcommand\poole           {{\namefont Poole}}
\newcommand\poolename       {{\namefont\david\ \poole}}

\newcommand\stolzenburg     {{\namefont Stol\-zen\-burg}}
\newcommand\stolzenburgname {{\namefont Frie\-der \stolzenburg}}

\newcommand\wirthindex                 {\index{Wirth, Claus-Peter (*1963)}}

\newcommand\wirth           {{\namefont Wirth}}
\newcommand\wirthnamenoindex{{\namefont\claus-\peter\ \wirth}}
\newcommand\wirthname       {\wirthindex\wirthnamenoindex}

\newcommand\afortiori{a fortiori}

\newcommand\FB   {FB}

\newcommand\FBautinfveryshort{\FB\ AI}

\newcommand\f    {\mbox{}{f.}}   
\newcommand\ff   {\mbox{}{ff.}}  

\newcommand\defi {Definition} 

\newcommand\Nov  {Nov.}
\newcommand\notonly{not \nolinebreak only}
\newcommand\onlyif{only \nolinebreak if}

\newcommand\qedhelp[1]{Q.e.d.~({#1})}
\newcommand\getittotheright[1]  
{\hfill\mbox{}\penalty 100\mbox{\ \,}\nolinebreak
\hfill\hfill\hfill\nolinebreak\mbox{#1}\ignorespaces}

\newcommand\Qedbf    [1]{\mbox{\bf\qedhelp{#1}}}

\newcommand\QEDbf    [1]{\getittotheright{\Qedbf    {#1}}}

\newcommand\role{r\^ole}

\newcommand\theo {Theorem}

\newcommand\WWW  {WWW}

\newcommand\aswell{as \nolinebreak well}

\newcommand\Cf   {Cf.}
\newcommand\cf   {cf.}
\def\nlbcite{\nolinebreak\cite}
\newcommand\Cfnlb{\Cf\nolinebreak}
\newcommand\cfnlb{\cf\nolinebreak}

\newcommand\CS   {Computer \Sci}

\newcommand\eg   {e.g.}

\newcommand\reductioadabsurdum{{\it argumentum ad absurdum}}

\newcommand\ie   {i.e.}

\newcommand\udiff{\ if\ }

\newcommand\p    {p.}
\newcommand\pp   {pp.}
\newcommand\PP[2]{\pp\,\ignorespaces#1--\ignorespaces#2}

\newcommand\sect {\myparagraphsymbol} 
\newcommand\sects{\myparagraphsymbol\myparagraphsymbol}

\newcommand\Sci  {Sci.}

\newcommand\wellfounded{well-founded}

\newcommand\wellfoundedness{well-founded\-ness}

\newcommand\wrt  {w.r.t.}

\newcommand\thewordand{and}
\newcommand\litspageref[1]{Page\,#1}

\newcommand\littheoref[1]{\theo\,#1}
\newcommand\litsectref[1]{\sect\,#1} 

\newcommand\litsectfromtoref[2]{\sects\,#1--#2}

\newcommand\litdefiref[1]{\defi\,#1}
\newcommand\Examplename{Ex\-am\-ple}
\newcommand\litexamref[1]{\Examplename\,#1}

\newcommand\litlemmref[1]{Lem\-ma\,#1}

\newcommand\litcororef[1]{Co\-rollary\,#1}

\newcommand\litexamrefs[2]{Examples~#1~\thewordand~#2}

\newcommand\litcororefs[2]
{Co\-rol\-laries \nolinebreak #1 \thewordand\ \nolinebreak #2}

\newcommand\litsectrefs[2]
{\sects\         \nolinebreak #1 \thewordand\ \nolinebreak #2}

\newcommand\litdefirefs[2]
{Definitions     \nolinebreak #1 \thewordand\ \nolinebreak #2}

\newcommand\litexamrefsss
[4]{Examples \nolinebreak #1, #2, #3, \thewordand\ \nolinebreak #4}
\newcommand\litexamrefssss
[5]{Examples \nolinebreak #1, #2, #3, #4 \thewordand\ \nolinebreak #5}

\newcommand\cororef[1]{\litcororef{\ref{#1}}}
\newcommand\lemmref[1]{\litlemmref{\ref{#1}}}

\newcommand\examref[1]{\litexamref{\ref{#1}}}

\newcommand\defiref[1]{\litdefiref{\ref{#1}}}
\newcommand\theoref[1]{\littheoref{\ref{#1}}}

\newcommand\sectref[1]{\litsectref{\ref{#1}}}
\newcommand\nlbsectref[1]{\nolinebreak\sectref{#1}}

\newcommand\defirefs[2]{\litdefirefs{\ref{#1}}{\ref{#2}}}

\newcommand\examrefs[2]{\litexamrefs{\ref{#1}}{\ref{#2}}}

\newcommand\cororefs[2]{\litcororefs{\ref{#1}}{\ref{#2}}}

\newcommand\sectrefs[2]{\litsectrefs{\ref{#1}}{\ref{#2}}}

\newcommand\examrefsss[4]
             {\litexamrefsss{\ref{#1}}{\ref{#2}}{\ref{#3}}{\ref{#4}}}
\newcommand\examrefssss[5]
  {\litexamrefssss{\ref{#1}}{\ref{#2}}{\ref{#3}}{\ref{#4}}{\ref{#5}}}

\newcommand\nthpositioner[2]
{#1\raisebox{0.52ex}{\scriptsize\hspace{0.07em}#2}}
\newcommand\nth[1]{\nthtinypositioner{#1}{\nthstring{#1}}}
\newcommand\nthtinypositioner[2]{#1\raisebox{0.52ex}{\tiny\hspace{0.07em}#2}}

\newcommand\mthpositioner[2]
{\math{#1}\raisebox{0.52ex}{\scriptsize\hspace{0.07em}#2}}
\newcommand\modulointocountzero[2]
{\count1=#1
\count2=#2
\count0=\count1
\divide  \count0 by \count2
\multiply\count0 by-\count2
\advance \count0 by \count1}
\newcommand\absolutevalueintocountzero[1]
{\count0=#1
\ifnum\count0<0\multiply\count0 by -1\fi}
\newcommand\nthstring[1]
{\def\myargone{#1}\ifcat a\myargone th\else\nthstringnochar{#1}\fi}
\newcommand\nthstringnochar[1]
{\absolutevalueintocountzero{#1}%
\modulointocountzero{\count0}{100} 
\ifnum\count0>9\ifnum\count0<20 th\else\stupidnthstring\fi
                                  \else\stupidnthstring\fi}
\newcommand\stupidnthstring
{\modulointocountzero{\count0}{10}
\ifnum\count0=1 \hskip-0.2em st\else
\ifnum\count0=2 nd\else
\ifnum\count0=3 rd\else 
                th\fi\fi\fi}

\newcommand\writeascents
[1]{\count4=#1
\ifnum\count4<0 
-\multiply\count4 by -1\fi
\modulointocountzero{\count4}{10}
\divide\count4 by 10
\count3=\the\count0
\modulointocountzero{\count4}{10}
\divide\count4 by 10
\the\count4
.\the\count0
\the\count3
}

\newcommand\frenchnthstring[1]
{\def\myargone{#1}\ifcat a\myargone th\else\frenchnthstringnochar{#1}\fi}
\newcommand\frenchnthstringnochar[1]
{\absolutevalueintocountzero{#1}%
\modulointocountzero{\count0}{100} 
\ifnum\count0>9\ifnum\count0<20 th\else\frenchstupidnthstring\fi
                                  \else\frenchstupidnthstring\fi}
\newcommand\frenchstupidnthstring
{\modulointocountzero{\count0}{10}
\ifnum\count0=1 \hskip-0.2em re\else
\ifnum\count0=2 me\else
\ifnum\count0=3 rd\else 
                th\fi\fi\fi}

\newcommand\CLAM      {{\rm CL\kern-.36em\raise.39ex\hbox{\sc a}\kern-.15emM}}

\newcommand\TEXMACS   {{\sc T\kern-.1667em\lower.5ex\hbox{E}\kern-.125emX\kern-.1em\lower.5ex\hbox{\textsc{m\kern-.05ema\kern-.125emc\kern-.05ems}}}}

\newcommand\Wernigerode    {Werni\-gerode}

\newcommand\plzwernigerode{\mbox{D--38855}}

\def       \emailcp      {{\tt wirth@logic.at}}

\newcommand\Institutedept
{\FBautinfveryshort}
\newcommand\Instituteinst
{\mbox{Hochschule Harz}}
\newcommand\Instituteplac
{\plzwernigerode\,\Wernigerode}
\newcommand\Institutecoun{Germany}
\newcommand\Institute
{\Institutedept, \Instituteinst, \Instituteplac, \Institutecoun}

\newcommand\academicpress{Academic Press (\elsevier)}

\newcommand\elsevier{Elsevier}

\newcommand\newspaperreference[5]
{\def\nameofjournalpress{#2}#1, #4 #5, #3\if?\nameofjournalpress
\else, #2\fi}

\newcommand\dateinjournal[1]{}

\newcommand\journalreference[6]
{\def\nameofjournalpress{#2}#1\nolinebreak\hskip.2em%
\dateinjournal{(#3) }{\mbox{\bf #4}}, \PP{#5}{#6}\if?\nameofjournalpress
\else, #2\fi}

\newcommand\journalreferenceprintyear[6]
{\def\nameofjournalpress{#2}#1 
#4:#5--#6, #3%
\if?\nameofjournalpress
\else, #2\fi}

\newcommand\journalreferenceprintyearaspartofnumber[6]
{\def\nameofjournalpress{#2}#1 
{#4/#3}, \PP{#5}{#6}\if?\nameofjournalpress
\else, #2\fi}

\newcommand\jscname
{J. Symbolic Computation}

\newcommand\jscprintyear
{\journalreferenceprintyear{\jscname}\academicpress}

\newcommand\tcsname{Theoretical \CS}
\newcommand\tcsjournal
{\journalreference\tcsname\elsevier}
\newcommand\tcsjournalprintyear
{\journalreferenceprintyear\tcsname\elsevier}

\urldef\wirthkuhnurl\url
{http://wirth.bplaced.net/SEKI/welcome.html#SWP-2007-01}

\urldef\urlsrninetythreedashzerofive\url
{http://wirth.bplaced.net/SEKI/welcome.html#SR-93-05}

\urldef\urlsreightyeighttwelve\url
{http://wirth.bplaced.net/SEKI/welcome.html#SR-88-12}

\renewcommand\namefont{\sc}
\begin{document}
\makecover
\maketitle
\begin{abstract}
In the middle of the 1980s,
\poolename\ introduced a semantical, model-theoretic notion of specificity to the 
artificial-intelligence community.
Since then it has found further applications 
in non-monotonic reasoning, in particular
in defeasible reasoning. 
\poole\ tried to approximate the intuitive human concept of specificity,
which seems to be
essential for reasoning in everyday life with its partial and inconsistent
information.
His notion, however, turns out to be intricate and problematic,
which 
\mbox{---~as we show~---} can be overcome to some extent by 
a closer approximation of the intuitive human concept of 
specificity. 
Besides the
intuitive advantages of our novel
specificity ordering over \poole's 
specificity relation in the classical examples of the literature, 
we also report some hard mathematical facts:
Contrary to what was claimed before,
we show that \poole's relation is not transitive.
The present means to decide our novel specificity relation,
however,
show only a slight improvement over the known ones for \poole's relation,
and further work is needed in this aspect.
\end{abstract}
\yestop\tableofcontents\vfill\pagebreak
\section{Introduction}
A possible explanation 
of
how humans manage to interact with reality
---~in spite of the fact that their 
information on 
the world is partial
and inconsistent~--- 
mainly consists of the following two points:\begin{enumerate}\noitem\item
Humans use a certain amount of 
rules for default reasoning and  
are aware that 
some arguments relying on these rules 
may be defeasible.
\noitem\item
In case of the frequent conflicting or even contradictory results 
of their reasoning,
they prefer more specific arguments to less specific ones.%
\noitem\end{enumerate}
The intuitive concept of 
specificity plays an essential \role\ in this explanation,
which is most interesting because it seems to be highly successful in practice, 
even if it were just an epiphenomenon
providing an {\it ex eventu}\/ explanation of human behavior.

On the 
long way approaching this proven intuitive human concept of specificity,
the first milestone marks the development of a semantical, 
model-theoretic notion of specificity having passed 
first tests of its usefulness and empirical validity.
\label{section where the requirement on being model-theoretic is given}%
Indeed,
at least as the first step,
a semantical, model-theoretic notion will probably 
offer a broader and better basis for applications
in 
systems for
common sense reasoning than notions of specificity that depend on 
peculiarities of special calculi or even on extra-logical procedures.
This holds in particular if the results 
of these systems
are to be accepted by human users or even 
by the human society.

\poolename\ has sketched such a notion 
as a binary relation on arguments
and evaluated its
intuitive validity with some examples 
in \cite{Poole-Preferring-Most-Specific-1985}. \hskip.4em
\poole's notion of specificity was given a more appropriate formalization
in \cite{Simari-Loui-Defeasible-Reasoning-1992}. \hskip.4em
The properties of this formalization
were examined in detail in \cite{Stolzenburg-etal-Computing-Specificity-2003}.

In this paper,
before we give a specification of the formal requirements on any 
reasonably conceivable relation of specificity
in \nlbsectref
{section Requirements Specification of Specificity in Logic Programming}, 
\hskip.2em
we present a detailed analysis of the intentional motivation 
of our {\em\mbox{intuition}
that \poole's specificity is a first step on the right way}
(\sectref{section Toward an Intuitive Notion of Specificity}). 
We \nolinebreak expect 
that the results of this analysis will carry us even beyond this 
paper to future improved concepts of specificity,
especially \wrt\ efficiency,
but also \wrt\ intuitive adequacy.
We hope that the closer we get to human intuition,
the more efficiently our concepts can be implemented,
simply because they seem to run so well on the human hardware,
which \mbox{---~by all} that we know today~---
seems to be pretty slow.

Moreover,
in \sectref{section Formalizations of Specificity}, \hskip.1em
we clearly disambiguate \poole's specificity from minorly improved versions
such as the one in \cite{Simari-Loui-Defeasible-Reasoning-1992}, \hskip.2em
and introduce a novel specificity relation
(\math{\lesssim_{\rm CP}}), \hskip.2em
which presents a major correction of \poole's specificity
because it removes a crucial shortcoming of \poole's original relation
(\math{\lesssim_{\rm P1}})
and its minor improvements 
(\maths{\lesssim_{\rm P2}}, \math{\lesssim_{\rm P3}}), \hskip.2em
namely their lack of transitivity. \hskip.2em

Furthermore,
in \sectref{section Putting Specificity to Test}, \hskip.1em
we present several examples that will convince every
carefully contemplating reader of the superiority of 
our novel specificity relation \nlbmath{\lesssim_{\rm CP}}
\wrt\ human intuition. \hskip.2em

Finally, 
we briefly discuss efficiency issues 
in \nlbsectref{section Efficiency Considerations}, \hskip.2em
and conclude with \nlbsectref{section Conclusion}.
\vfill\pagebreak
\section{Basic Notions and Notation}
\noindent
For the remainder of \thispaper, let us narrow the general logical 
setting of specificity down to the concrete framework 
of {\em defeasible logic with the restrictions of logic programming}, \hskip.2em
as \nolinebreak found \eg\ in  
\cite{Stolzenburg-etal-Computing-Specificity-2003} and
\cite{Chesnevar-etal-Relating-Deafeasible-2003}.
\begin{definition}[Literal, Rule]
\\\noindent A {\em literal}\/ is an atom, possibly prefixed with the 
symbol 
\nolinebreak ``\math\neg\nolinebreak\hskip.1em\nolinebreak''
for 
negation.\footnote{%
 To distinguish this kind of negation here from default negation, 
 the symbol ``\math\sim'' is sometimes used in the literature  
 in place of our standard symbol ``\math\neg\closequotefullstopnospace}
A {\em rule}\/ is a literal, 
but
possibly suffixed with a reverse implication symbol 
``\tightantiimplies\nolinebreak\hskip.1em\nolinebreak''
that is followed by a conjunction of literals,
consisting of one literal at least.
\end{definition}\begin{definition}[Theory, Derivation, Contradictory]
\\\noindent 
Let \math\Pi\ be a set of rules.
The {\em theory of \nlbmath\Pi}\/ is the set \nlbmath{\mathfrak T_\Pi}
inductively defined to contain\footroom\begin{itemize}\notop\item 
all instances of literals from \nlbmath\Pi\ 
and \noitem\item
all literals \math{L} 
for which
there is a conjunction \nlbmath{C} of literals from \nlbmath{\mathfrak T_\Pi}
such that \\\maths{L\antiimplies C}{} 
is an instance of a rule in \nlbmaths\Pi.\notop\footroom\end{itemize}
For \math{\mathfrak L\subseteq\mathfrak T_\Pi}, 
we also say that {\em\math\Pi\,derives\/ \nlbmath{\mathfrak L}}, \
and write \maths{\Pi\yields\mathfrak L}.
\\\noindent
\math\Pi\ is called
{\em contradictory}
\udiff\
there is an atom \math A such that \bigmaths{\Pi\yields\{A,\neg A\}};
otherwise \math\Pi\ is {\em non-contradictory}.
\end{definition}\begin{example}\sloppy
\bigmaths{\{\ident A,\ \neg\ident A\,\tightantiimplies\,\ident A\}}{}
is contradictory, \ 
but \bigmaths{\{\ident A\,\tightantiimplies\,\neg\ident A,\ 
\neg\ident A\,\tightantiimplies\,\ident A\}}{}
is non-contradictory, \hskip.1em
although we can infer both \math{\ident A} and \nlbmath{\neg\ident A}
in classic (\ie\ two-valued) \hskip.1em logic. \hskip.5em
Our notions of consequence and consistency are equivalent
both ---~because of the restrictiveness of our poor formal language~---
to intuitionistic logic and to the three-valued logic 
where \math\neg~and~\tightund\ are given as usual\commanospace\footnote{%
 The standard interpretation is that \TRUEpp\ is \nlbmaths 1, \ 
 \UNDEFpp\ is \nlbmaths{{1\over 2}}, \
 \FALSEpp\ is \nlbmaths 0, \
 \math{\neg A} is \nlbmaths{1\tight-A}, \ 
 and 
 \math{A\tightund B} is \nlbmaths{\min\{A,B\}}. \ \ In other words: \
 \mbox{\maths{\neg\TRUEpp=\FALSEpp},} \ \
 \maths{\neg\UNDEFpp=\UNDEFpp}, \ \
 \maths{\neg\FALSEpp=\TRUEpp}; \ \
 \maths{\TRUEpp\tightund A=A}, \ \
 \maths{\UNDEFpp\tightund\TRUEpp=\UNDEFpp}, \ \
 \maths{\UNDEFpp\tightund\UNDEFpp=\UNDEFpp}, \ \
 \maths{\UNDEFpp\tightund\FALSEpp=\FALSEpp}, \ \
 \maths{\FALSEpp\tightund A=\FALSEpp}.%
} \hskip.2em
but \hskip.1em
(following neither \kleene\ nor \lukasiewicz) \hskip.1em
implication has to be defined via
\\[+.5ex]\noindent
\mbox{}\hfill\maths{\inpit{A\tightantiimplies\TRUEpp}=A}, \ \hfill
\maths{\inpit{A\tightantiimplies\FALSEpp}=\TRUEpp}, \ \hfill
\maths{\inpit{A\tightantiimplies\UNDEFpp}=\TRUEpp}. \hfill
\mbox{}\end{example}
\subsection{Global Parameters for the Given Specification}\label
{subsection Global Parameters for a Given Specification}
Throughout \thispaper,
we will assume a set of literals and
two sets of rules to be given:\begin{itemize}\noitem\item
A \nolinebreak set \math{\Pioffont^{\rm F}} of literals 
meant to describe the {\em\underline facts}\/ 
of the concrete situation under consideration,\noitem\item
a set \math{\Pioffont^{\rm G}} 
of {\em\underline general rules}\/
meant to hold in all possible worlds,
and\noitem\item 
a set \math\Deltaoffont\ of {\em\underline defeasible}\/ (or default) rules
meant to hold in most situations.\footnote{%
 In the approach of
 \cite{Stolzenburg-etal-Computing-Specificity-2003},
 the set 
 \nlbmath{\Pioffont^{\rm G}} must not contain mere literals 
 (without suffixed condition). 
 To obtain a more general setting, 
 we omit this additional restriction in the context of \thispaper,
 simply because it is neither intuitive nor required for our
 framework here.%
}%
\noitem\end{itemize}\noindent The set $\Pioffont := \Pioffont^{\rm F}\cup\Pioffont^{\rm G}$ is the set of {\em strict}\/ rules that 
---~contrary to the defeasible rules~---
are considered to be safe and are not doubted in any concrete situation.%

\subsection{Formalization of Arguments}
There is no difference in derivation between 
the 
strict rules from \nlbmath\Pioffont\
and the defeasible rules from \nlbmaths\Deltaoffont. 
\hskip.2em
If a contradiction occurs, 
however,
we will narrow the defeasible rules from 
\nlbmath\Deltaoffont\ down to a subset \nlbmath{\mathcal A} 
of its {\em ground}\/ instances; \hskip.2em
\ie\ instances without free variables, 
such that no further instantiation can occur. \hskip.2em
Such a subset, together with the literal whose derivation is in focus, 
is called an {\em argument}. \hskip.2em
With implicit reference to the fixed sets of rules \math\Pioffont\ and 
\nlbmath\Deltaoffont, 
the formal definition is as simple as follows.\footnotemark\notop\halftop
\begin{definition}[Argument]\label{definition  argument}
\\\pair{\mathcal A}L is an {\em argument}
\udiff\ \math{\mathcal A} is a 
set of ground instances of rules 
from \nlbmaths\Deltaoffont\ \
and \bigmaths{\mathcal A\cup\Pioffont\ \yields\ \{L\}}.
\end{definition}
\subsection{Notation of Concrete Examples}
For ease of distinction, 
we will use the special symbol ``\tightdefeasibleantiimplies''
as a syntactical sugar
in concrete examples of defeasible rules from \nlbmaths\Deltaoffont, \hskip.2em
instead of the symbol ``\tightantiimplies\closequotecomma
which 
---~in our concrete examples~--- 
will be used only in strict rules.\notop\halftop
\begin{example}[\litexamref 1 of \cite{Poole-Preferring-Most-Specific-1985}]
\label{example poole 1 precisely}
\\[-3ex]\math{\begin{array}[t]{@{}l l l@{}}
\mbox{}
\\[+.3ex]
  \Pioffont^{\rm F}_{\ref{example poole 1 precisely}}
 &:=
 &\left\{\begin{array}{l}\ident{bird}(\ident{tweety}),
   \\\ident{emu}(\ident{edna})
   \\\end{array}\right\},
\\\Pioffont^{\rm G}_{\ref{example poole 1 precisely}}
 &:=
 &\left\{\begin{array}{l}\ident{bird}(x)\antiimplies\ident{emu}(x),
   \\\neg\ident{flies}(x)\antiimplies\ident{emu}(x)
   \\\end{array}\right\},
\\\Deltaoffont_{\ref{example poole 1 precisely}}
 &:=
 &\left\{\begin{array}{l}
     \ident{flies}(x)\defeasibleantiimplies\ident{bird}(x)
   \\\end{array}\right\},
\\\mathcal A_2
 &:=
 &\left\{\begin{array}{l}
     \ident{flies}(\ident{edna})\defeasibleantiimplies\ident{bird}(\ident{edna})
   \\\end{array}\right\}.
\\\end{array}}\hfill
\begin{tikzpicture}[baseline=(current bounding box.north),
>=stealth,->,looseness=.5,auto]
\matrix [matrix of math nodes,
column sep={1.4cm,between origins},
row sep={1.3cm,between origins}]{
   |(notfliesedna)| \math{\neg\ident{flies}(\ident{edna})} 
 & 
 & |(fliesedna)| \math{\ident{flies}(\ident{edna})} 
 &
 & |(fliestweety)| \math{\ident{flies}(\ident{tweety})} 
\\ 
 &
 & |(birdedna)| \math{\ident{bird}(\ident{edna})}
 &  
 & |(birdtweety)| \math{\ident{bird}(\ident{tweety})}  
\\
 & |(emuedna)| \math{\ident{emu}(\ident{edna})}
 &
 & |(TRUE)| \math{\TRUEpp}    
\\
};
\begin{scope}[every node/.style={font=\small\itshape}]
\draw [] (birdedna) -- node [right] 
      {$\scriptscriptstyle\!\mathcal A_2$}  
      (fliesedna);
\draw [] (birdtweety) -- (fliestweety);
\draw [double] (emuedna) -- (birdedna);
\draw [double] (emuedna) -- (notfliesedna);
\draw [double] (TRUE) -- (emuedna);
\draw [double] (TRUE) -- (birdtweety);
\end{scope}
\end{tikzpicture}
\\[-1.3ex]\noindent We have
\bigmaths{\begin{array}[t]{l l l}
  \mathfrak T_{\Pioffont_{\ref{example poole 1 precisely}}}
 &=
 &\{\ident{bird}(\ident{tweety}),
  \ident{emu}(\ident{edna}),
  \ident{bird}(\ident{edna}),
  \neg\ident{flies}(\ident{edna})\},
\\\mathfrak T_{\Pioffont_{\ref{example poole 1 precisely}}
        \cup\Deltaoffont_{\ref{example poole 1 precisely}}}
 &=
 &\{\ident{flies}(\ident{edna}),\ident{flies}(\ident{tweety})\}\cup
  \mathfrak T_{\Pioffont_{\ref{example poole 1 precisely}}}.
\\\end{array}}{}
\\\noindent 
It is intuitively clear here that we prefer the
argument 
\pair\emptyset{\neg\ident{flies}(\ident{edna})}
to the argument
\pair{\mathcal A_2}{\ident{flies}(\ident{edna})},
simply because the former is more specific.
We will further discuss this in \examref{example poole 1 precisely discussion}.%
\end{example}
\subsection{Quasi-Orderings}\label
{subsection Quasi-Orderings}%
We will use several binary relations comparing arguments 
according to their specificity. \hskip.3em
For \nolinebreak any 
relation 
written as \nlbmath{\lesssim_N}
(``being more or equivalently specific \wrt\ \nlbmath N''), \
we set%
\\[+.5ex]\noindent\LINEmaths{\begin{array}[t]{@{}l l l@{}}\tight{\gtrsim_N}
 &:= 
 &\setwith{\pair X Y}{Y\lesssim_N X}
\hfill\mbox{(``less or equivalently specific'')},
\\\tight{\approx_N}
 &:=
 &\tight{\lesssim_N}\cap\tight{\gtrsim_N}
\hfill\mbox{(``equivalently specific'')},
\\\tight{<_N}
 &:=
 &\tight{\lesssim_N}\setminus\tight{\gtrsim_N}
\hfill\mbox{(``properly more specific'')},
\\\tight{\leq_N}
 &:=
 &\tight{<_N}\cup\setwith{\pair X X}{X\mbox{ is an argument}}
\hfill\mbox{(``more specific or equal'')},
\\\tight{\vartriangle_N}
 &:=
 &\displaysetwith{\pair X Y}{\begin{array}{@{}l@{}}X,Y\mbox{ are arguments with }
     \\X\not\lesssim_N Y\mbox{ and }X\not\gtrsim_N Y
     \\\end{array}}
\hfill\mbox{~~(``incomparable \wrt\ specificity'')}.
\\\end{array}}{}
\\\noindent
A {\em quasi-ordering}\/ is a reflexive transitive relation. \hskip.2em
An {\em (irreflexive) ordering}\/ 
is an irreflexive transitive relation. \hskip.2em
A {\em reflexive ordering}\/ (also called: ``partial ordering'') \hskip.1em 
is an anti-symmetric quasi-ordering. \hskip.2em
An {\em equivalence}\/ is a symmetric quasi-ordering.%
\notop\halftop
\begin{corollary}\label{corollary quasi-orderings}\ \sloppy
If\/ \nlbmath{\lesssim_N} is a quasi-ordering, \hskip.2em
then\/ 
\math{\approx_N} is an equivalence, \hskip.1em
\math{<_N} is an ordering, \hskip.1em
and\/
\math{\leq_N} is a reflexive ordering.%
\end{corollary}%
\pagebreak\footnotetext{%
 Some authors (\cf\ \eg\ \cite{Stolzenburg-etal-Computing-Specificity-2003},
 \cite{Chesnevar-etal-Relating-Deafeasible-2003}) \hskip.2em
 require arguments to be non-contradictory \wrt\ \nlbmath\Pioffont,
 and the \nth 1 element of an argument to be
 \tight\subset-minimal \wrt\ the derivability of the \nth 2 element.
 To obtain a more general setting, 
 we omit these additional restrictions in the context of \thispaper.
 For the omission of the minimality requirement see also 
 \cororefs{corollary sub-argument P}{corollary sub-argument CP}.%
}%
\section{%
Toward an Intuitive Notion of Specificity
}\label{section Toward an Intuitive Notion of Specificity}
\halftop\noindent
It is part of general knowledge that
a criterion is \opt{properly} more specific than another one if 
the \mbox{``class\,of\,candidates\,that\,satisfy\,it''}
is a \opt{proper} subclass of that 
of the other one.

Analogously ---~taking logical formulas as the criteria~---
a \nolinebreak formula \math A is \opt{properly} more specific 
than a formula \nlbmaths B, \hskip.2em
if the model class of \nlbmath A is a \opt{proper} subclass of 
the model class of \nlbmaths B, \hskip.3em
\ie\nolinebreak\ if \hskip.2em\maths{A\models B}{} \
\mbox{\opt{and \math{B\,\notmodels\,A}}}.

\mbox{If we consider} a formula as a predicate 
on model-theoretic structures,
its model class becomes the extension of this predicate.
From this viewpoint,
we can state \bigmaths{A\models B}{}
also as the syllogism 
\mbox{``every \math A is \math B\hskip.06em\closequotecommanospace} \,
and also as the \lambertdiagram\footnote{%
 \majorheadroom
 \Cfnlb\ \cite[Dianoiologie, \litsectfromtoref{173}{194}]{lambert-1764}.%
}
\par\halftop\noindent\LINEmath
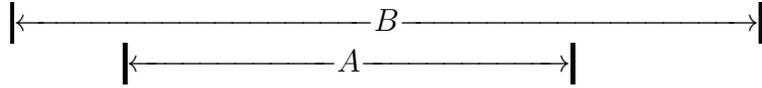
{
\begin{tabular}
{@{}l@{}l@{}l@{}l@{}l@{}l@{}l@{}l@{}}%
\multicolumn{8}{@{}l@{}}{\mybox{10cm}{\math B}}
\\\naught&\multicolumn{4}{@{}l@{}}{\mybox{6cm}{\math A}}
\\\end{tabular}}\par
\halftop\halftop\halftop
\subsection{Arguments as an Intuitive Abstraction}\label
{subsection Arguments}
To enable a closer investigation 
of the critical parts of a defeasible derivation,
we have to isolate the defeasible parts in a derivation.
Abstracting from the concrete derivation of a literal \nlbmaths L, \hskip.1em
let us take the set \nlbmath{\mathcal A}
of the ground instances of the defeasible rules
that are actually applied in the derivation,
and form the pair \pair{\mathcal A}L,
which we already called an {\em argument}\/ in \defiref{definition  argument}.

\subsection{Activation Sets}\label
{subsection Activation Sets}
If we want to classify a derivation with defeasible rules
according to its specificity,
then we \nolinebreak have to isolate the defeasible part of the 
derivation and look at its input.
In \nolinebreak our setting, 
the input consists of 
the set of those literals on 
which the defeasible part of the derivation is based,
called the {\em activation set}\/ 
for the defeasible part of the derivation.
In our framework of defeasible logic programming, 
the only relevant 
property of an activation set can be the conjunction of its
literals which is immediately represented by the set itself.

Because all literals of an activation set have been derived 
from the given specification,
it \nolinebreak does not make sense
to compare activation sets \wrt\ the models of the entire
specification. \hskip.3em
Indeed,
only a comparison
\wrt\ 
the models of 
a sub-specification can show any differences between them. \hskip.3em
In our case, 
we have to exclude \nlbmath{\Pioffont^{\rm F}}
from the specification. \hskip.3em
This exclusion makes sense because the defeasible rules are 
typically default rules 
not written in \nolinebreak\mbox{particular} for the given 
concrete situation 
that is formalized 
by \nlbmaths{\Pioffont^{\rm F}\!}. \hskip.3em
Moreover,
as we \nolinebreak want to compare the defeasible parts of derivations,
we have to exclude the defeasible rules from \nlbmath{\Deltaoffont} \aswell.
\hskip.2em
Thus, 
on the one hand, 
all we can take into account from our specification
is a subset 
of the general rules \nlbmaths{\Pioffont^{\rm G}\!}. \hskip.3em
On the other hand,
it is clear that
we want to have the {\em entire}\/ set \nlbmath{\Pioffont^{\rm G}}
for our comparison of activation sets,
simply because we want to base our specificity classification
on our specification, namely on its general and strict part. 
Moreover,
as will be explained in 
\sectref{subsubsection Growth of the Defeasible Parts toward the Leaves},
\hskip.2em
it is hardly meaningful to exclude any proper (\ie\ non-literal)
rule from \nlbmaths{\Pioffont^{\rm G}\!}. \hskip.3em
All in all, 
we conclude that \math{\Pioffont^{\rm G}} is that part of our specification
according to which activation sets are to be compared.

Very roughly speaking, if we have fewer activation sets, 
then we have fewer models,
which again means to have a higher specificity.
Accordingly,
the first straightforward
sketch of a notion of specificity could be given as
follows:\begin{quote}
An argument \nlbmath{\pair{\mathcal A_1}{L_1}}
is \opt{properly} {\em more specific
than}\/ an argument \nlbmath{\pair{\mathcal A_2}{L_2}} \hskip.3em
if, \hskip.2em
for each activation set \nlbmath{H_1} of 
\nlbmaths{\pair{\mathcal A_1}{L_1}}, \hskip.3em
there is an activation set 
\math{H_2\subseteq\mathfrak T_{H_1\cup\,\Pioffont^{\rm G}}}
of \nlbmath{\pair{\mathcal A_2}{L_2}}
\mbox{\opt{but not vice versa}.}\end{quote}
Note that this notion of specificity is preliminary,
and that the notion
of an activation set has not been properly defined yet.

\subsection{Isolation of the Defeasible Parts of a Derivation}

On the one hand,
the argument \pair{\mathcal A}L 
(described in \sectref{subsection Arguments}) \hskip.1em
is a nice abstraction from the derivation of \nlbmaths L, \hskip.1em
because it perfectly suits our model-theoretic intentions
described in 
\sectref{section where the requirement on being model-theoretic is given}.
By this abstraction, on the other hand,
we lose the possibility to isolate the defeasible
parts of the derivation more precisely.

\subsubsection{Precise Isolation in And-Trees}%
\label{subsubsection Precise Isolation in And-Trees} 
Let us compare this set \nlbmath{\mathcal A}
with an {\em and-tree of the derivation}. \hskip.4em
Every node in such a tree is labeled with 
the conclusion of an instance of a rule, 
such that its children are labeled exactly with the 
elements of the conjunction in the condition of this instance.

\halftop\noindent
An isolation of the defeasible parts of an and-tree of the derivation may proceed
as follows:\begin{itemize}\noitem\item
Starting from the root of the tree, 
we iteratively erase all applications of strict rules.
This results in a set of trees, each of which has the 
application of a defeasible rule at the root.\noitem\item
Starting now from the leaves of these trees,
we again erase all applications of strict rules.
This results in a set of trees where all nodes {\em all}\/ 
of whose children are leaves
result from an application of 
\nolinebreak a \nolinebreak defeasible \nolinebreak rule.\end{itemize}
\subsubsection{A first approximation of Activation Sets}%
In a first approximation,
we may now take all leaves of all resulting trees as 
the activation set for the original derivation.%

\pagebreak

\subsubsection{Growth of the Defeasible Parts toward the Leaves}\label
{subsubsection Growth of the Defeasible Parts toward the Leaves}%
Note that in the set of trees resulting from the procedure described in 
\sectref{subsubsection Precise Isolation in And-Trees}, \hskip.2em
there may well have remained instances of rules from \nlbmath{\Pioffont^{\rm G}}
connecting a defeasible root application with the defeasible applications
right at the leaves. 
Thus
\mbox{---~to cover} the whole defeasible part of the derivation
in our abstraction~---
we \nolinebreak have to consider the set
\nlbmath{\mathcal A\cup\Pioffont^{\rm G}} \mbox{instead of just the set
\nlbmaths{\mathcal A}.}

 More precisely, 
 we have to include all proper rules 
 (\ie\ those with non-empty conditions) \hskip.1em
 from 
 \nlbmath{\Pioffont^{\rm G}\!}, 
 and may also include the literals in 
 \nlbmath{\Pioffont^{\rm G}} because they cannot do any harm.
 Note that
 the need to \nolinebreak include all proper rules and to exclude 
 the literals from \nlbmath{\Pioffont^{\rm F}}
 provides a motivation for simply defining 
 \math{\Pioffont^{\rm G}} \nolinebreak to \nolinebreak contain 
 exactly the proper rules 
 \nolinebreak of \nlbmaths\Pioffont, \hskip.1em
 such as found in \cite{Stolzenburg-etal-Computing-Specificity-2003}.

As a consequence, \hskip.1em
in the modeling via our abstraction \nlbmath{\mathcal A}, \hskip.2em
we cannot prevent the 
precisely isolated defeasible sub-trees
resulting from the procedure described in 
\sectref{subsubsection Precise Isolation in And-Trees} \hskip.1em
from using the rules from 
\nlbmath{\Pioffont^{\rm G}} to grow
toward the root and toward the leaves again.\footnote{%
 Of course, our abstraction admits even different 
 defeasible parts of a different derivation tree that derives the 
 same literal in focus from the same set \nlbmath{\mathcal A}
 of instances of defeasible rules, 
 \ie\ different derivations of \math L from \math{\mathcal A\cup\Pioffont}
 for the identical argument \pair{\mathcal A}L\@. \ 
 The admission of these multiple derivations is intended
 in our model-theoretic treatment.
 The only effect on our current discussion, however, is that we 
 would have to treat several trees disjunctively, which actually makes no
 difference for the ideas we are currently trying to express.%
}
It is clear, 
however,
that 
only the growth toward the leaves 
can affect our activation \nolinebreak sets and our 
notion of specificity.

Let us have a closer look at the effects of such a growth
toward the leaves
in the most simple case. \hskip.3em
In addition to a given
activation set \nlbmath{\{\Qpppeins\appzero\}}, \hskip.2em
in the presence of a general rule
\par\noindent\LINEmaths{\Qpppeins x
\antiimplies\Ppppeinsindex 0 x\tightund\cdots\tightund\Ppppeinsindex{n-1}x}{}
\par\noindent
from \nlbmaths{\Pioffont^{\rm G}\!}, \hskip.25em
we will also have to consider 
the activation set
\nlbmaths{\setwith{\Ppppeinsindex i\appzero}{i\tightin\{0,\ldots,n\tight-1\}}}. 
\hskip.3em
This has two effects, which we will discuss in 
what follows.

The first effect is that we immediately realize  
that every model of \math{\Pioffont^{\rm G}}
that is \mbox{represented} by the activation set
\nlbmath{\setwith{\Ppppeinsindex i\appzero}{i\tightin\{0,\ldots,n\tight-1\}}} 
\hskip.1em
is also 
represented by the activation set
\nlbmath{\{\Qpppeins\appzero\}}; \hskip.3em
simply because 
\nlbmath{\setwith{\Ppppeinsindex i\appzero}{i\tightin\{0,\ldots,n\tight-1\}}}
is added to the activation sets
by a growth toward the leaves,
preventing that we fail to realize that an argumentation
based on \nlbmath{\{\Qpppeins\appzero\}}
is less \mbox{(or equivalently)} 
specific than any argumentation that gets along with
\mbox
{\maths{\setwith{\Ppppeinsindex i\appzero}{i\tightin\{0,\ldots,n\tight-1\}}}.}%
\notop\halftop
\subsubsection{Preference of  the ``More Concise''}\label
{subsubsection Preference of  the ``More Concise''}
The second effect, however, is that
an argumentation that gets along with \nlbmath{\{\Qpppeins\appzero\}}
becomes even {\em properly}\/ less specific than one that actually requires
\nlbmath{\setwith{\Ppppeinsindex i\appzero}{i\tightin\{0,\ldots,n\tight-1\}}}
\hskip.2em
and does not get along with \nlbmaths{\{\Qpppeins\appzero\}},\footnote{%
 \majorheadroom
 This can happen \onlyif\ we have 
 \bigmaths{\setwith{\Ppppeinsindex i\appzero}{i\tightin\{0,\ldots,n\tight-1\}}
 \nsubseteq\{\Qpppeins\appzero\}},
 \ie\ \onlyif\ \bigmaths{n\tightnotequal 0}.%
} \hskip.25em
simply because the former argumentation has  
the additional activation set \nlbmath{\{\Qpppeins\appzero\}}. \hskip.3em

For instance, in our \examref{example poole 1 precisely}, \hskip.1em
an argumentation that gets along with \nlbmath{\ident{bird}(\ident{edna})}
\hskip.1em is properly less specific than one that actually requires
\nlbmaths{\ident{emu}(\ident{edna})}.
This effect is 
called\footnote{%
\majorheadroom
 \Cfnlb\ \eg\ \cite[\p\,94]{Stolzenburg-etal-Computing-Specificity-2003}, 
 \cite[\p 108]{Garcia-Simari-Defeasible-2004}.%
}
{\em preference of the 
\mbox{``more concise''}}.

\pagebreak
\noindent
The problem now is that the statement 
\par\noindent\LINEmaths{\Qpppeins\appzero\ \notmodels\
\Ppppeinsindex 0\appzero\tightund\cdots\tightund\Ppppeinsindex{n-1}\appzero},
\par\noindent which is required to justify the 
the appropriateness of this effect,
is not explicitly given by the specification 
via \nlbmaths{\trip{\Pioffont^{\rm F}}{\Pioffont^{\rm G}}\Deltaoffont}. \,

\indent
Nevertheless
---~if we do not just want to see it as a matter-of-fact property
of notions of specificity in the style of \poole~---
the preference of the ``more concise''
can be justified by the habits of human specifiers as follows:

\indent
If human specifiers write an implication
in form of a rule
\bigmaths{\Qpppeins x
\antiimplies\Ppppeinsindex 0 x\tightund\cdots\tightund\Ppppeinsindex{n-1}x}{}
into a specification \nlbmath\Pioffont\
of strict (\ie\ non-defeasible) knowledge,
then they typically intend that 
the implication is proper in the sense that its converse does not hold
in general; \hskip.2em
otherwise they would have used an equivalence or equality symbol 
instead of the implication symbol, \hskip.1em
or replaced
each occurrence of each \Qpppeins t with
\bigmaths{\Ppppeinsindex 0 t\tightund\cdots\tightund\Ppppeinsindex{n-1}t},
respectively. In particular, in our setting of logic programming 
---~where 
disjunctive properties of the definition of 
a predicate are spread over several rules~--- 
the implications definitely tend to be proper.

\indent
Therefore, if \nolinebreak seasoned specifiers write down such a rule,
then they do not want to exclude models where \Qpsymbol\ holds for
some object \appzero, but not all of the \math{\Ppsymbol_i} do.
This means that if we \nolinebreak find such a rule
in the strict and general part \nlbmath{\Pioffont^{\rm G}} of a specification, 
then
it is reasonable to assume that the implication is proper
\wrt\ the intuition captured in the defeasible rules in \nlbmaths\Deltaoffont.

\indent
As a consequence, 
it makes sense to consider a defeasible argument
based on \setwith{\Ppppeinsindex i\appzero}
{i\tightin\{0,\ldots,n\tight-1\}} \hskip.3em
to be properly more specific than an
argument that can get along with \nlbmaths{\Qpppeins\appzero}.
\par\halftop\halftop\noindent
\LINEmath
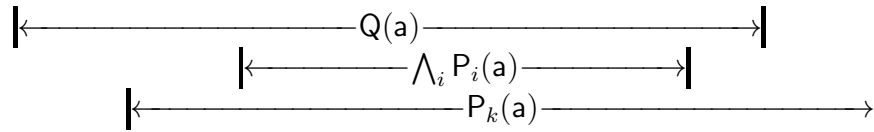
{
\begin{tabular}
{@{}l@{}l@{}l@{}l@{}l@{}l@{}l@{}l@{}}%
\multicolumn{8}{@{}l@{}}{\mybox{10cm}{\Qpppeins\appzero}}
\\\naught&\naught
 &\multicolumn{4}{@{}l@{}}{\mybox{6cm}
  {\math{\bigwedge_i\Ppppeinsindex i\appzero}}}
\\\naught&\multicolumn{6}{@{}l@{}}{\mybox{10cm}
  {\math{\Ppppeinsindex k\appzero}}}
\\\end{tabular}}%
\par\halftop\halftop\halftop\halftop\indent
The standard example for the preference of the ``more concise''
is \examref{example poole 2}. \hskip.3em
It is a variation of our former \examref{example poole 1 precisely}.
\par\halftop\indent
Finally, 
let us remark that our justification for the preference of the
``more concise'' does not apply if \hskip.1em
\bigmaths{\Qpppeins x
\antiimplies\Ppppeinsindex 0 x\tightund\cdots\tightund\Ppppeinsindex{n-1}x}{}
is a {\em defeasible}\/ rule instead of a strict one, \hskip.1em
because we then have the following three problems:\begin{itemize}\noitem\item
the inclusion given by the rule is not generally intended 
(otherwise the rule should be a strict one),\noitem\item
we cannot easily describe 
the actual instances to which the default rule is meant to apply 
(otherwise this more concrete description of the defeasible rule should
be stated as a strict rule), \hskip.3em
and\noitem\item
the direct treatment of a defeasible equivalence neither has to be 
appropriate as a default rule in the given situation,
nor do we have any means to express a defeasible equivalence in the 
current setting.\noitem\end{itemize}
Accordingly,
there is, 
for instance, 
no clear reason to prefer the first argument of 
\examref{example poole 3} to the second one.

\subsubsection{Preference of the ``More Precise''}%
By an analogous argumentation on the intentions of human specifiers,
we can say that an argument that essentially requires an activation
set \setwith{\Ppppeinsindex i\appzero}
{i\tightin\{0,\ldots,n\}} \hskip.1em
is {\em properly}\/ more specific than an argument that gets along with 
a proper subset \setwith{\Ppppeinsindex i\appzero}{i\tightin I} \hskip.1em
for some index set \nlbmaths{I\subset\{0,\ldots,n\}}. \hskip.3em
The effect of the assumption of this
intention is sometimes\footnote{%
 \Cfnlb\ \eg\ \cite[\p\,94]{Stolzenburg-etal-Computing-Specificity-2003}, 
 \cite[\p 108]{Garcia-Simari-Defeasible-2004}.%
}
called {\em preference of the 
\mbox{``more precise\closequotefullstopnospace}}

\halftop\indent
There is, however, an exception to be observed where this analogy 
does not apply,
namely
the case that we actually can derive 
the set from its subset with the help 
of \nlbmaths{\Pioffont^{\rm G}\!}. \hskip.4em
In this case, 
the above-mentioned growth
toward the leaves with rules from \nlbmath{\Pioffont^{\rm G}}
again implements
the approximation of the subclass relation 
among model classes via the one among activation sets,
as demonstrated in \examref{example stolzenburg 1}.

\halftop\indent
Apart from this exception,
there is again a problem, 
namely that it is not the case that 
\par\noindent\LINEmaths{
\bigwedge_{i\in I}\Ppppeinsindex i\appzero\ \notmodels\
\bigwedge_{i\in\{0,\ldots,n\}}\Ppppeinsindex i\appzero}{}
\par\noindent would be explicitly given by the specification
via \nlbmaths{\trip{\Pioffont^{\rm F}}{\Pioffont^{\rm G}}\Deltaoffont}. \,
Nevertheless
---~if we do not just want to see it as a matter-of-fact property
of notions of specificity in the style of \poole~---
we \nolinebreak can again justify that it is unlikely that a seasoned 
specifier would not have intended this non-consequence statement,
\hskip.2em
namely by an argumentation analogous to the one we gave for 
the preference of the ``more concise\closequotefullstopextraspace
Indeed, a seasoned specifier who wants to exclude the above
non-consequence would just specify a rule like
\par\noindent\LINEmaths{\Ppppeinsindex j x\antiimplies
\bigwedge_{i\in I}\Ppppeinsindex i x},
\par\noindent for each \math{j\tightin\{0,\ldots,n\}\tightsetminus I}.
\par\halftop\halftop\halftop\noindent
\LINEmath{
\begin{tabular}
{@{}l@{}l@{}l@{}l@{}l@{}l@{}l@{}l@{}}%
  \naught
 &\naught
 &\multicolumn
    {6}
    {@{}l@{}}
    {\mybox{8cm}{\math{\bigwedge_{i\in I}\Ppppeinsindex i\appzero}}}
\\\naught
 &\naught
 &\naught
 &\multicolumn
    {4}
    {@{}l@{}}
    {\mybox{6cm}{\math{\bigwedge_{i\in\{1,\ldots,n\}} \Ppppeinsindex i\appzero}}}
\\\naught&\multicolumn{6}{@{}l@{}}{\mybox{10cm}{\math{\Ppppeinsindex k\appzero}}}
\\\end{tabular}}{}
\par\halftop\halftop\halftop
An elementary example for the preference of the ``more precise''
is \examref{example lovely one}.

\yestop\yestop\yestop\subsubsection{Conclusion on the Preferences}
\halftop\noindent
After all,
even if you do not buy our justification of 
the preference of the ``more concise'' and the 
``more precise\closequotecommasmallextraspace
you can still follow our investigations into the properties
of these preferences \wrt\ \poole's model-theoretic notion 
of specificity and our correction \nolinebreak of \nolinebreak it 
in \nolinebreak the \nolinebreak following sections.

\vfill\pagebreak
\section{Requirements Specification of Specificity in Logic Programming}\label
{section Requirements Specification of Specificity in Logic Programming}
\halftop\noindent
With implicit reference to specification 
via \nlbmath{\trip{\Pioffont^{\rm F}}{\Pioffont^{\rm G}}\Deltaoffont}
(\cfnlb\ \sectref{subsection Global Parameters for a Given Specification}),
\hskip.3em
let \nolinebreak us designate \poole's relation of being more 
(or equivalently) specific
by \hskip.2em``\math{\lesssim_{\rm P1}}\closequotefullstopextraspace
Here, ``P1'' stands for 
``\poole's original version\closequotefullstopnospace

The standard usage of the symbol 
``\math\lesssim\nolinebreak\hskip.1em\nolinebreak''
is to denote a {\em quasi-ordering} 
(\cfnlb\ \sectref{subsection Quasi-Orderings}). \hskip.3em
Instead of the symbol 
\nolinebreak\hskip.2em\nolinebreak``\math\lesssim\nolinebreak\hskip.1em
\nolinebreak\closequotecommasmallextraspace
however, \hskip.1em
\citet{Poole-Preferring-Most-Specific-1985} 
uses the symbol \nolinebreak
``\math\leq\nolinebreak\hskip.1em\nolinebreak\closequotefullstopextraspace 
The standard usage of the 
symbol \nolinebreak``\math\leq\nolinebreak\hskip.1em\nolinebreak'' is to denote 
a {\em reflexive ordering} 
(\cfnlb\ \sectref{subsection Quasi-Orderings}). \hskip.3em
We \nolinebreak cannot conclude from this, \hskip.1em
however, \hskip.1em
that \poole\
intended the additional property of anti-symmetry; \hskip.3em
indeed, 
we find 
a concrete example specification
in \nlbcite{Poole-Preferring-Most-Specific-1985} \hskip.1em
where the lack of anti-symmetry of 
\nolinebreak\hskip.2em\nlbmath{\lesssim_{\rm P1}} \hskip.1em
is made explicit.\footnote{%
 Here we refer to the last three sentences of
 \litsectref{3.2} on \litspageref{145} of 
 \cite{Poole-Preferring-Most-Specific-1985}.%
}

The possible lack of anti-symmetry of quasi-orderings
---~\ie\ that different \mbox{arguments} may have an equivalent specificity~---
cannot be a problem because any quasi-ordering \nlbmath{\lesssim_N} \hskip.1em
immediately provides us with its equivalence \nlbmaths{\approx_N}, \hskip.2em
its ordering \nlbmaths{<_N}, \hskip.2em
and its reflexive ordering \nlbmath{\leq_N} 
(\cfnlb\ \cororef{corollary quasi-orderings}).

By contrast to the non-intended anti-symmetry,
{\em transitivity}\/ is obviously a {\it conditio sine qua non}\/
for any useful notion of specificity. \hskip.2em
Indeed,
if we already have 
an argument \nlbmath{\pair{\mathcal A_2}{\ident{wine}}}
that is more specific than another argument 
\nlbmaths{\pair{\mathcal A_3}{\ident{vodka}}}, \hskip.2em
and if we come up with yet another argument 
\nlbmath{\pair{\mathcal A_1}{\ident{beer}}} 
that is even more specific than \nlbmaths{\pair{\mathcal A_2}{\ident{wine}}},
\hskip.2em
then, \hskip.1em
by all means, \hskip.1em
\math{\pair{\mathcal A_1}{\ident{beer}}} \nolinebreak should 
be more specific than 
the argument \nlbmath{\pair{\mathcal A_3}{\ident{vodka}}} \aswell. \hskip.2em
It is obvious that a notion of specificity without transitivity 
could hardly be helpful in practice.

A further {\it conditio sine qua non}\/ for any useful notion of 
specificity is that the conjunctive combination of 
respectively more specific arguments results in a more specific argument.
Indeed, if a square is more specific than a rectangle and a circle is
more specific than an ellipse, then a square inscribed into a circle
should be more specific than a rectangle inscribed into an ellipse. \hskip.2em
Already in \cite{Poole-Preferring-Most-Specific-1985}, \hskip.2em
we find an example\footnote{%
 \majorheadroom
 Here we refer to 
 \litexamref 6 of \cite[\litsectref{3.5}, \p 146]
 {Poole-Preferring-Most-Specific-1985}, \hskip.2em
 which we present here as our \examref{example poole 6}.%
} 
where \math{\lesssim_{\rm P1}} \nolinebreak
violates this monotonicity property of the conjunction,
which is described there as 
``seemingly un\-intuitive\closequotefullstopnospace\footnote{%
 \majorheadroom
 See our \examref{example poole 6} and the references there.%
} 

Further 
intricacies of computing \poole's specificity
in concrete
examples are described in 
\cite{Stolzenburg-etal-Computing-Specificity-2003}\commanospace\footnote{%
 \majorheadroom
  Here we refer to \litsectref{3.2\ff} of 
 \cite{Stolzenburg-etal-Computing-Specificity-2003}, \hskip.1em
 where it is demonstrated that,
 for deciding \poole's specificity relation 
 (actually \nlbmath{\lesssim_{\rm P2}} instead of \nlbmaths{\lesssim_{\rm P1}}, 
  but this does not make any difference here) 
 for two input arguments, 
 we sometimes have to consider even those defeasible rules 
 which are not part of any of these arguments.%
} \hskip.2em
which will make it hard to implement
\math{\lesssim_{\rm P1}} \nolinebreak
or its minor corrections
as efficiently as required in the practice of logic programming.
\vfill\pagebreak
\section{Formalizations of Specificity}\label
{section Formalizations of Specificity}%
\yestop\subsection{Activation Sets}
\yestop\noindent A generative, bottom-up 
(\ie\ from the leaves to the root) \hskip.1em
derivation with defeasible rules can now be
split into three phases of derivation of 
literals from literals. \hskip.1em
This splitting follows the discussion in 
\sectref{subsubsection Precise Isolation in And-Trees}
on how to isolate the defeasible part of a derivation (phase\,2) \hskip.1em
from strict parts that may occur toward the root~(phase\,3) \hskip.1em
and toward the leaves (phase\,1):\label
{section where the phases are}\begin{description}\item[(phase\,1) ]
First we derive the literals 
that provide the basis for specificity considerations.

In our approach we derive the set \nlbmath{\mathfrak T_\Pioffont} here. 
\hskip.2em
\poole\ takes the set 
\nlbmath{\mathfrak T_{\Pioffont\cup\Deltaoffont}} instead.\item[(phase\,2) ]
On the basis of \begin{itemize}\notop\item
a subset \math H of the literals derived in phase\,1,\noitem\item
the first item \nlbmath{\mathcal A} of a given argument 
\pair{\mathcal A}L, \hskip.2em and\noitem\item
the general rules \nlbmath{\Pioffont^{\rm G}\!},\notop\end{itemize}
we derive a further set of literals \nlbmath{\mathfrak L}: \ \bigmaths
{H\cup\mathcal A\cup\Pioffont^{\rm G}\ \yields\ \mathfrak L}.\item[(phase\,3) ]
Finally, \hskip.1em
on the basis of \nolinebreak\hskip.2em\maths{\mathfrak L}, \hskip.2em 
the literal of the argument is derived:
\bigmaths{\mathfrak L\cup\Pioffont\ \yields\ \{L\}}.

In \poole's approach, 
phase\,3 is empty and we simply have \bigmaths{\mathfrak L\tightequal\{L\}}. \
In our approach, \hskip.1em
however, \hskip.1em
it is admitted to use the facts from \nlbmath{\Pioffont^{\rm F}} 
in phase\,3, \hskip.2em
in addition to the general rules from \nlbmath{\Pioffont^{\rm G}\!}, \hskip.2em
which were already admitted in phase\,2\@.
\end{description}

\yestop\noindent
With implicit reference to our sets 
\math{\Pioffont=\Pioffont^{\rm F}\cup\Pioffont^{\rm G}} \hskip.1em
and \nlbmaths\Deltaoffont, \hskip.3em
the phases~2 and~3 can be more easily expressed with the help of the 
following notions.

\label{section definition activation set}%
\begin{definition}
[\opt{Minimal} \opt{Simplified} Activation Set]\label{definition activation set}
\\Let \math{\mathcal A} be a set of ground instances of rules from 
\nlbmaths\Deltaoffont,
and let \math L be a literal.
\\\math H is a {\em simplified activation set for}\/
\nlbmath{\pair{\mathcal A}L} 
\udiff\ \maths{L\in\mathfrak
T_{H\,\tightcup\,\mathcal A\,\tightcup\,\Pioffont^{\rm G}}\,}.
\\\math H is an {\em activation set for}\/
\nlbmath{\pair{\mathcal A}L} 
\udiff\ \math{
L\in\mathfrak T_{\,\mathfrak L\,\cup\Pioffont}} 
for some \maths{\mathfrak L\subseteq
\mathfrak T_{H\,\tightcup\,\mathcal A\,\tightcup\,\Pioffont^{\rm G}}\,}.
\\\math H is a {\em minimal}\/ \opt{{\em simplified}\/} 
{\em activation set for}\/
\nlbmath{\pair{\mathcal A}L} 
\udiff\ \math H is an \opt{simplified} activation set for 
\nlbmaths{\pair{\mathcal A}L}, 
\hskip.2em
but no proper subset of \nlbmath H 
is an \opt{simplified} activation set for \nlbmaths{\pair{\mathcal A}L}.%
\end{definition}

\yestop\noindent
Roughly speaking, 
an argument is now more (or equivalently) specific
than another one 
if, for each of its activation sets \nlbmaths{H_1},
the same set \nlbmath{H_1}
is also an activation set for the
other argument. \hskip.3em
Note that we have replaced here the option of some 
\math{H_2\subseteq\mathfrak T_{H_1\cup\,\Pioffont^{\rm G}}}
of the first straightforward sketch for a notion of specificity displayed in 
\sectref{subsection Activation Sets} \hskip.15em
with the more restrictive \maths{H_2\tightequal H_1}. \hskip.4em
Indeed, 
this simplification applies here
because all we consider from any activation set \nlbmath H
in \defiref{definition activation set} 
(such as \nlbmath{H_2} in this case) \hskip.1em
is just the 
closure \nolinebreak\hskip.15em\nlbmaths{
\mathfrak T_{H\,\tightcup\,\mathcal A\,\tightcup\,\Pioffont^{\rm G}}
=
\mathfrak T_{\mathfrak T_{H\cup\Pioffont^{\rm G}}
\,\tightcup\,\mathcal A\,\tightcup\,\Pioffont^{\rm G}}}.

\yestop\yestop
\noindent
Activation sets that are not simplified differ from simplified ones 
by the admission of facts from \nlbmath{\Pioffont^{\rm F}}
(in addition to the general rules \nlbmath{\Pioffont^{\rm G}}) \hskip.1em
after the defeasible part of the argumentation is completed
(as can be seen in \examref{example stolzenburg 2}). \hskip.2em
\pagebreak
Our introduction of activation sets that are not simplified 
is a conceptually important correction of \poole's approach:
It must be admitted to use the facts besides the general rules
in a purely strict argumentation 
that is based on literals resulting from completed defeasible arguments, 
\hskip.1em
simply because 
the defeasible parts of an argumentation \mbox{(as isolated} in 
\nlbsectref{subsubsection Precise Isolation in And-Trees}) \hskip.1em
should not get more specific by the later
use of additional facts that do not provide input
to the defeasible parts.\footnote{%
 We do not further discuss this obviously appropriate correction here
 and leave the construction of examples that make the conceptual necessity 
 of this correction intuitively clear as an exercise; \hskip.2em 
 \eg\nolinebreak\ by presenting two different sets of strict rules
 with equal derivability, where only one needs the facts in phase\,3
 and where the additional specificity gained by these facts 
 violates the intuition.%
} \hskip.3em
Note that the difference between simplified and non-simplified 
activation sets typically occurs in real applications, \hskip.1em
but 
---~except \examref{example stolzenburg 2}~---
not in our toy examples of \sectref{section Putting Specificity to Test}
designed to discuss the results of the differences in phase\,1.

\subsection{\poole's Specificity Relation P1,\\and its Minor Corrections P2 and P3}\label
{subsection Specificity Relation P1}%

In this section
we \nolinebreak will define the binary relations 
\math{\lesssim_{\rm P1}}, 
\math{\lesssim_{\rm P2}},
\math{\lesssim_{\rm P3}} 
of \hskip.2em``being more or equivalently specific according to 
\poolename\nolinebreak\hskip.05em\nolinebreak''
with implicit reference to our sets of facts and of general and defeasible rules
\mbox{(\ie\ to \maths{\Pioffont^{\rm F}\!},
\maths{\Pioffont^{\rm G}\!},
and \nlbmath\Deltaoffont, respectively).}

The relation \nlbmath{\lesssim_{\rm P1}} of the 
following definition is precisely \poole's
original relation \nlbmath\geq\ as defined at the bottom of the left column
on \litspageref{145} of 
\cite{Poole-Preferring-Most-Specific-1985}. \hskip.3em
See 
\sectref{section Requirements Specification of Specificity in Logic Programming}
\hskip.1em
for our reasons to write ``\math\gtrsim''  
instead of \hskip.2em``\math\geq'' as a first change.
\hskip.3em
Moreover, as a second change required by mathematical standards, \hskip.1em 
we have replaced the symbol 
\nolinebreak``\nlbmath\gtrsim\nolinebreak\hskip.1em\nolinebreak'' with the symbol
\nolinebreak``\nlbmath\lesssim\nolinebreak\hskip.1em\nolinebreak''
(such that the smaller argument becomes the more specific one), \hskip.2em
so that
the relevant \wellfoundedness\ becomes the one of its ordering \nlbmath< 
instead of the reverse \tight>.

\begin{definition}%
[\poolename's Original Specificity \math{\lesssim_{\rm P1}}]
\label{DefSpec1}
\\\maths{\pair{\mathcal A_1}{L_1}
\lesssim_{\rm P1}
\pair{\mathcal A_2}{L_2}}{}
\udiff\
\pair{\mathcal A_1}{L_1} and
\pair{\mathcal A_2}{L_2} 
are arguments, \hskip.1em
and if, \hskip.2em
for every \math{H\subseteq\mathfrak T_{\Pioffont\cup\Deltaoffont}} 
that is a simplified activation set for 
\nlbmath{\pair{\mathcal A_1}{L_1}} 
but not a simplified activation set for \nlbmath{\pair{\mathcal A_2}{L_1}}, \ 
\math H is also 
a simplified activation set for \nlbmaths{\pair{\mathcal A_2}{L_2}}.
\end{definition}

\noindent
The relation \nlbmath{\lesssim_{\rm P2}} of the following 
definition is the relation \nlbmath\succeq\ of \litdefiref{10}
on \litspageref{94} 
of \cite{Stolzenburg-etal-Computing-Specificity-2003}
(attributed to \cite{Poole-Preferring-Most-Specific-1985}). \hskip.3em
Moreover, the relation \math{>_{\rm spec}} of \litdefiref{2.12} on 
\litspageref{132} of 
\cite{Simari-Loui-Defeasible-Reasoning-1992}
(attributed to \cite{Poole-Preferring-Most-Specific-1985} \aswell) \hskip.1em
is \nolinebreak the 
relation \nlbmath{\tight{<_{\rm P2}}\nottight{\nottight{:=}}
\tight{\lesssim_{\rm P2}}\setminus
\tight{\gtrsim_{\rm P2}}}.

\begin{definition}%
[Standard Version of \poolename's Specificity \math{\lesssim_{\rm P2}}]
\label{DefSpec2}
\\\maths{\pair{\mathcal A_1}{L_1}
\lesssim_{\rm P2}
\pair{\mathcal A_2}{L_2}}{}
\udiff\
\pair{\mathcal A_1}{L_1} and
\pair{\mathcal A_2}{L_2} 
are arguments, \hskip.1em
and if, \hskip.2em
for every \math{H\subseteq\mathfrak T_{\Pioffont\cup\Deltaoffont}} 
that is a simplified activation set for 
\nlbmath{\pair{\mathcal A_1}{L_1}} 
but not a simplified activation set for \nlbmath{\pair\emptyset{L_1}}, \ 
\math H is also 
a simplified activation set for \nlbmaths{\pair{\mathcal A_2}{L_2}}.
\end{definition}
The only change in \defiref{DefSpec2} as compared to 
\defiref{DefSpec1} is that ``\pair{\mathcal A_2}{L_1}'' is replaced with
``\pair\emptyset{L_1}\closequotefullstopextraspace
We did not
encounter any example yet where this most appropriate correction of the 
counter-intuitive variant ``\pair{\mathcal A_2}{L_1}''
of \defiref{DefSpec1}
makes any difference to today's standard ``\pair\emptyset{L_1}'' in 
\defiref{DefSpec2}, \hskip.1em
and leave it as an exercise to construct one.

\pagebreak

The relations \nlbmath{\lesssim_{\rm P1}} and \nlbmath{\lesssim_{\rm P2}}
were not meant to compare arguments for literals that do not need 
any defeasible rules 
---~or at least they do not show an intuitive behavior
on such arguments, as shown in 
\examref{example not well on non-defeasible arguments}.
\begin{example}
[Minor Flaw of \nlbmath{\lesssim_{\rm P1}} and \nlbmath{\lesssim_{\rm P2}}]\label
{example not well on non-defeasible arguments}%
\\[-3ex]\math{\begin{array}[t]{@{}l l l@{}}
\mbox{}
\\[+.3ex]
  \Pioffont^{\rm F}_{\ref{example not well on non-defeasible arguments}}
 &:=
 &\left\{\begin{array}{l}\ident{thirst}
   \\\end{array}\right\},
\\\Pioffont^{\rm G}_{\ref{example not well on non-defeasible arguments}}
 &:=
 &\left\{\begin{array}{l}\ident{drink}\antiimplies\ident{thirst}
   \\\end{array}\right\},
\\\Deltaoffont_{\ref{example not well on non-defeasible arguments}}
 &:=
 &\left\{\begin{array}{l}
     \ident{beer}\defeasibleantiimplies\ident{thirst}
   \\\end{array}\right\},
\\\mathcal A_1
 &:=
 &\Deltaoffont_{\ref{example not well on non-defeasible arguments}}.
\\\end{array}}\hfill
\begin{tikzpicture}[baseline=(current bounding box.north),
>=stealth,->,looseness=.5,auto]
\matrix [matrix of math nodes,
column sep={1.4cm,between origins},
row sep={1.3cm,between origins}]{
   |(drink)| \math{\ident{drink}} 
 &
 & |(beer)| \math{\ident{beer}}
\\
 & |(thirst)| \math{\ident{thirst}}
 &
 & |(TRUE)| \math{\TRUEpp}    
\\
};
\begin{scope}[every node/.style={font=\small\itshape}]
\draw [] (thirst) -- node [right] 
      {$\scriptscriptstyle\!\mathcal A_1$}  
      (beer);
\draw [double] (thirst) -- (drink);
\draw [double] (TRUE) -- (thirst);
\end{scope}
\end{tikzpicture}
\\\noindent Let us compare the specificity of the arguments 
\pair{\mathcal A_1}{\ident{beer}}
and 
\pair\emptyset{\ident{drink}}.
\\\noindent We have
\LINEmaths{\begin{array}[t]{l l l}
  \mathfrak T_{\Pioffont_{\ref{example not well on non-defeasible arguments}}}
 &=
 &\{\ident{thirst},\ident{drink}\},
 \end{array}\hfill\begin{array}[t]{l l l}
 \mathfrak T_{\Pioffont_{\ref{example not well on non-defeasible arguments}}
        \cup\Deltaoffont_{\ref{example not well on non-defeasible arguments}}}
 &=
 &\{\ident{beer}\}\cup
  \mathfrak T_{\Pioffont_{\ref{example not well on non-defeasible arguments}}}.
\\\end{array}}{}
\\\noindent We have 
\bigmaths{\pair{\mathcal A_1}{\ident{beer}}\lesssim_{\rm P2}
\pair\emptyset{\ident{drink}}}{}
because for every \math{H\subseteq\mathfrak T
_{\Pioffont_{\ref{example not well on non-defeasible arguments}}
\cup\Deltaoffont_{\ref{example not well on non-defeasible arguments}}}} 
that is a simplified activation set for 
\nlbmaths{\pair{\mathcal A_1}{\ident{beer}}}, \hskip.3em
but not a simplified activation set for \nlbmath{\pair\emptyset{\ident{beer}}}, 
\hskip.3em
we \nolinebreak have \bigmathnlb{H\tightequal\{\ident{thirst}\}},
which is a 
simplified activation set also for \nlbmaths{\pair\emptyset{\ident{drink}}}.
\\We have 
\bigmaths{
\pair\emptyset{\ident{drink}}
\lesssim_{\rm P2}
\pair{\mathcal A_1}{\ident{beer}}
}{}
because there 
cannot be 
a simplified activation set for
\nlbmath{\pair\emptyset{\ident{drink}}} \hskip.1em
that is not  
a simplified activation set 
for \nlbmaths{\pair\emptyset{\ident{drink}}}. 
\\All in all, we get 
\bigmaths{\pair{\mathcal A_1}{\ident{beer}}\approx_{\rm P2}
\pair\emptyset{\ident{drink}}},
although \bigmaths{\pair\emptyset{\ident{drink}}
<_{\rm P3}\pair{\mathcal A_1}{\ident{beer}}
}{} must be given according to intuition,
because, if \ident{beer} produces a conflict with our drinking habits,
there is no reason to prefer it to another \ident{drink}.
\\Finally note that by \cororef{corollary ordering the P}, \hskip.2em 
we will get \bigmaths{\pair{\mathcal A_1}{\ident{beer}}\approx_{\rm P1}
\pair\emptyset{\ident{drink}}}{} \aswell.
\end{example}

\yestop
\noindent To overcome this minor flaw,
we finally add an implication as an additional requirement
in \defiref{DefSpec}.
\hskip.25em
This implication guarantees that no argument that requires
defeasible rules can be more specific
than an argument that does not require any defeasible rules at all.
\begin{definition}%
[Rather Unflawed Version of \poolename's Specificity \math{\lesssim_{\rm P3}}]
\label{DefSpec}
\\\maths{\pair{\mathcal A_1}{L_1}
\lesssim_{\rm P3}
\pair{\mathcal A_2}{L_2}}{}
\udiff\
\pair{\mathcal A_1}{L_1} and
\pair{\mathcal A_2}{L_2} 
are arguments,
\bigmaths{L_2\tightin\mathfrak T_\Pioffont}{} implies
\bigmaths{L_1\tightin\mathfrak T_\Pioffont}, 
and if, 
for every \math{H\subseteq\mathfrak T_{\Pioffont\cup\Deltaoffont}} 
that is a \opt{minimal}\footnote{%
 Note that the omission of the optional restriction to {\em minimal}\/
 simplified activation sets for \nlbmath{\pair{\mathcal A_1}{L_1}} 
 in \defiref{DefSpec} 
 has no effect on the extension of the defined notion, \hskip.1em
 simply because the additional 
 non-minimal simplified activation sets \nlbmath{\pair{\mathcal A_1}{L_1}} 
 will then be simplified
 activation sets for \nlbmath{\pair{\mathcal A_2}{L_2}} {\it\afortiori}.%
} 
simplified activation set for 
\nlbmath{\pair{\mathcal A_1}{L_1}} 
but not a simplified activation set for \nlbmath{\pair\emptyset{L_1}}, \ 
\math H is also 
a simplified activation set for \nlbmaths{\pair{\mathcal A_2}{L_2}}.
\end{definition}

\yestop\noindent
As every simplified activation set that passes the 
condition of \defiref{DefSpec1} \hskip.2em
also passes the one of \defiref{DefSpec2},
we get the following corollary of our three definitions. 
\begin{corollary}\label{corollary ordering the P} \
\bigmaths{\tight{\lesssim_{\rm P3}}
\nottight{\nottight\subseteq}\tight{\lesssim_{\rm P2}}
\nottight{\nottight\subseteq}\tight{\lesssim_{\rm P1}}
}.%
\end{corollary}\begin{corollary}\label{corollary sub-argument P} \
If\/ \pair{\mathcal A_1}{L_1},
\pair{\mathcal A_2}{L_2} 
are arguments and we have
\bigmaths{\mathcal A_1\,\tightsubseteq\,\mathcal A_2}{} and
\bigmathnlb{L_1\tightequal L_2},
then we have 
\bigmaths{\pair{\mathcal A_1}{L_1}
\lesssim_{\rm P3}
\pair{\mathcal A_2}{L_2}}{}.\end{corollary}

\yestop\noindent
By \cororefs{corollary ordering the P}{corollary sub-argument P}, \hskip.5em
\maths{\lesssim_{\rm P1}},  \hskip.2em
\maths{\lesssim_{\rm P2}},  \hskip.2em
\maths{\lesssim_{\rm P3}}{} \hskip.1em
are reflexive relations on arguments, \hskip.2em
but 
\ \mbox{---~~as we will show in \examref{example P not transitive}}
\hskip.2em and \mbox{state in \theoref{theorem P not transitive}~~---} \ 
not quasi-orderings in general. \hskip.5em

\yestop
\begin{example}[Counterexample to the Transitivities]
\label{example P not transitive}\sloppy
\\[-2.2ex]\math{\begin{array}[t]{@{}l l@{}}\multicolumn{2}{@{}l@{}}{%
\begin{array}[t]{@{}l@{\,}l@{\,}l@{}}
    \\[-0.4ex]\Pioffont^{\rm F}_{\ref{example P not transitive}}
     &:=
     &\left\{\begin{array}{l}\ident{alcohol},
        \\\ident{blessing},
        \\\ident{thirst}
        \\\end{array}\right\},
    \\\Pioffont^{\rm G}_{\ref{example P not transitive}}
     &:=
     &\left\{\begin{array}{l}\ident{wine}\antiimplies\ident e
        \\\end{array}\right\},
    \\\Deltaoffont_{\ref{example P not transitive}}
     &:=
     &\mathcal A_1\cup\mathcal A_2\cup\mathcal A_3,
    \\\mathcal A_1
     &:=
     &\left\{\begin{array}{l}\ident e
          \defeasibleantiimplies\ident{alcohol}
                       \tightund\ident{blessing}\tightund\ident{thirst},
        \\\ident{beer}\defeasibleantiimplies\ident e
        \\\end{array}\right\},
    \\\mathcal A_2
     &:=
     &\left\{\begin{array}{l}\ident{wine}
          \defeasibleantiimplies\ident{alcohol}\tightund\ident{blessing}
        \\\end{array}\right\},
    \\\mathcal A_3
     &:=
     &\left\{\begin{array}{l}\ident{vodka}
          \defeasibleantiimplies\ident{alcohol}
        \\\end{array}\right\}.
    \\\end{array}}
\\
\\[-1.3ex]
  \multicolumn{2}{@{}l@{}}{\mbox{Compare the specificity of the arguments}}
\\\multicolumn{2}{@{}l@{}}{\pair{\mathcal A_1}{\ident{beer}}, \
\pair{\mathcal A_2}{\ident{wine}}, \
\pair{\mathcal A_3}{\ident{vodka}}\,!}
\\\end{array}}\hspace*{-1.8em}%
\begin{tikzpicture}[baseline=(current bounding box.north),
>=stealth,->,looseness=.5,auto]
\matrix [matrix of math nodes,
column sep={1.4cm,between origins},
row sep={1.3cm,between origins}]{
   |(vodka)|    \math{\ident{vodka}} 
&& |(wineeins)| \math{\ident{wine}} 
&& |(winezwei)| \math{\ident{wine}} 
&& |(beer)|     \math{\ident{beer}}
\\ 
&& 
&& |(e)|        \math{\ident e}
\\ 
\\ 
 & |(alcohol)|  \math{\ident{alcohol}}
&& |(blessing)| \math{\ident{blessing}}
&& |(c)|        \math{\ident{thirst}}
\\
&& 
 & |(TRUE)|     \math{\TRUEpp}  
\\
};
\begin{scope}[every node/.style={midway,auto}]
\draw [] (alcohol) -- node [near start, right] 
      {$\scriptscriptstyle\mathcal A_1$} 
      (e);
\draw [] (blessing) -- node [right] 
      {$\scriptscriptstyle\!\!\mathcal A_1$} 
      (e);
\draw [] (c) -- node [right] 
      {$\scriptscriptstyle\mathcal A_1$} 
      (e);
\draw [] (e) -- node [near start, right] 
      {$\scriptscriptstyle\,\,\mathcal A_1$} 
      (beer);
\draw [] (alcohol) -- node [right] 
      {$\scriptscriptstyle\!\!\mathcal A_2$} 
      (wineeins);
\draw [] (blessing) -- node [near end, right] 
      {$\scriptscriptstyle\!\mathcal A_2$} 
      (wineeins);
\draw [] (alcohol) -- node [right] 
      {$\scriptscriptstyle\!\!\mathcal A_3$} 
      (vodka);
\draw [double, -] (wineeins) -- (winezwei);
\draw [double]    (e)        -- (winezwei);
\draw [double]    (TRUE)     -- (alcohol);
\draw [double]    (TRUE)     -- (blessing);
\draw [double]    (TRUE)     -- (c);
\end{scope}\end{tikzpicture}\end{example}
\yestop\begin{lemma}\label{lemma P not transitive}
\\[+.1ex]\noindent There are\/\begin{itemize}\notop\item
a specification\/
\trip{\Pioffont^{\rm F}_{\ref{example P not transitive}}}
{\Pioffont^{\rm G}_{\ref{example P not transitive}}}
{\Deltaoffont_{\ref{example P not transitive}}} \hskip.2em
without any negative literals
\\(\ie, {\it\afortiori}, \hskip.2em
 \nlbmath{\Pioffont^{\rm F}_{\ref{example P not transitive}}\cup
 \Pioffont^{\rm G}_{\ref{example P not transitive}}\cup
 \Deltaoffont_{\ref{example P not transitive}}}
 \mbox{is non-contradictory),} \hskip.3em
and\noitem\item
arguments\/ \hskip.1em
\pair{\mathcal A_1}{L_1}, \hskip.2em
\pair{\mathcal A_2}{L_2}, \hskip.2em
\pair{\mathcal A_3}{L_3} \hskip.1em
with respectively minimal sets\/
\maths{\mathcal A_1},
\nlbmaths{\mathcal A_2},
\nlbmath {\mathcal A_3}
\\(\ie, \hskip.1em\pair{\mathcal A'_i}{L_i} is not an argument for any
 proper subset \math{\mathcal A'_i\subset\mathcal A_i}),\noitem\end{itemize}
 
\noindent\maths{
\begin{array}{@{}r@{\mbox{~~~}} l c l c l c l@{}}\mbox{such that}
 &\pair{\mathcal A_1}{L_1}
 &\lesssim_{\rm P3}
 &\pair{\mathcal A_2}{L_2}
 &\lesssim_{\rm P3}
 &\pair{\mathcal A_3}{L_3}
 &\not\gtrsim_{\rm P1}
 &\pair{\mathcal A_1}{L_1}
\\\mbox{and}
 &\pair{\mathcal A_1}{L_1}
 &\not\gtrsim_{\rm P1}
 &\pair{\mathcal A_2}{L_2}
 &\not\gtrsim_{\rm P1}
 &\pair{\mathcal A_3}{L_3}.
\\\end{array}}{}\end{lemma}
\yestop\yestop\begin{proofparsepqed}{\lemmref{lemma P not transitive}}
Looking at \examref{example P not transitive}, \hskip.2em
we see that only the quasi-ordering properties 
in the last two lines of \lemmref{lemma P not transitive}
are non-trivial.
\hskip.2em
We have
\par\noindent\LINEmaths{
  \mathfrak T_{\Pioffont_{\ref{example P not transitive}}}
 =
 \{\ident{alcohol},\ident{blessing},\ident{thirst}\},
 \hfill
 \mathfrak T_{\Pioffont_{\ref{example P not transitive}}
        \cup\Deltaoffont_{\ref{example P not transitive}}}
 =
 \{\ident e,\ident{beer},\ident{wine},\ident{vodka}\}\cup
  \mathfrak T_{\Pioffont_{\ref{example P not transitive}}}}.
\par\noindent
Thus, regarding the arguments \pair{\mathcal A_1}{\ident{beer}},
\pair{\mathcal A_2}{\ident{wine}},
\pair{\mathcal A_3}{\ident{vodka}}, \hskip.2em
the additional implication condition of \defiref{DefSpec}
as compared to \defirefs{DefSpec1}{DefSpec2}
is always satisfied, simply because 
its condition is always false.

\initial{\underline{\maths{\pair{\mathcal A_3}{\ident{vodka}}
\not\gtrsim_{\rm P1}\pair{\mathcal A_1}{\ident{beer}}
\lesssim_{\rm P3}\pair{\mathcal A_2}{\ident{wine}}}:}}
The minimal simplified activation sets for \pair{\mathcal A_1}{\ident{beer}}
that are subsets of \nlbmath{\mathfrak 
T_{\Pioffont_{\ref{example P not transitive}}
\cup\Deltaoffont_{\ref{example P not transitive}}}}
and no simplified activation sets for \pair\emptyset{\ident{beer}}
(or, without any difference, for \pair{\mathcal A_3}{\ident{beer}})
are \nlbmath{\{\ident{alcohol},\ident{blessing},\ident{thirst}\}} 
and \maths{\{\ident e\}}, \hskip.2em
which are simplified activation sets for \pair{\mathcal A_2}{\ident{wine}}
\ \mbox{---~~but} \math{\{\ident e\}} is no simplified activation set 
for \pair{\mathcal A_3}{\ident{vodka}}.

\initial{\underline{\maths{\pair{\mathcal A_1}{\ident{beer}}
\not\gtrsim_{\rm P1}\pair{\mathcal A_2}{\ident{wine}}
\lesssim_{\rm P3}\pair{\mathcal A_3}{\ident{vodka}}}:}}
The only minimal simplified activation set for \pair{\mathcal A_2}{\ident{wine}}
that is a subset of \nlbmath{\mathfrak 
T_{\Pioffont_{\ref{example P not transitive}}
\cup\Deltaoffont_{\ref{example P not transitive}}}}
and no simplified activation set for \pair\emptyset{\ident{wine}}
(such \nolinebreak as \nlbmath{\{\ident e\}}) \hskip.1em
(or, without any difference, for \pair{\mathcal A_1}{\ident{wine}})
is \nlbmath{\{\ident{alcohol},\ident{blessing}\}}, \hskip.2em
which is a simplified activation set for \pair{\mathcal A_3}{\ident{vodka}},
but not for \pair{\mathcal A_2}{\ident{beer}}.

\initial{\underline{\maths{\pair{\mathcal A_2}{\ident{wine}}
\not\gtrsim_{\rm P1}\pair{\mathcal A_3}{\ident{vodka}}}:}}
The only minimal simplified activation set for \pair{\mathcal A_3}{\ident{vodka}}
that is a subset of \nlbmath{\mathfrak 
T_{\Pioffont_{\ref{example P not transitive}}
\cup\Deltaoffont_{\ref{example P not transitive}}}}
and no simplified activation set for \pair{\mathcal A_2}{\ident{vodka}}
is \math{\{\ident{alcohol}\}}, \hskip.2em
which is not a simplified activation 
set for \pair{\mathcal A_2}{\ident{wine}}.
\end{proofparsepqed}\begin{sloppypar}%
\pagebreak\indent
The relations stated in \lemmref{lemma P not transitive} \hskip.1em
hold \notonly\ for the given indices, \hskip.1em
but 
\mbox{---~by} \mbox{\cororef{corollary ordering the P}~---}
actually for all of P1, P2, P3; \hskip.3em
and so we immediately get:
\end{sloppypar}\halftop\begin{theorem}\label{theorem P not transitive}
\\There is a specification\/ 
\trip{\Pioffont^{\rm F}_{\ref{example P not transitive}}}
{\Pioffont^{\rm G}_{\ref{example P not transitive}}}
{\Deltaoffont_{\ref{example P not transitive}}}, \hskip.2em
such that\/
\nlbmath{\Pioffont^{\rm F}_{\ref{example P not transitive}}\cup
 \Pioffont^{\rm G}_{\ref{example P not transitive}}\cup
 \Deltaoffont_{\ref{example P not transitive}}}
 \mbox{is non-contradictory,} \hskip.3em
but none of\/        \hskip.1em
\maths{\lesssim_{\rm P1}}, \hskip.2em 
\maths{\lesssim_{\rm P2}}, \hskip.2em 
\maths{\lesssim_{\rm P3}}, \hskip.2em 
\maths{<_{\rm P1}},        \hskip.2em 
\maths{<_{\rm P2}},        \hskip.2em 
\math{<_{\rm P3}} 
\mbox{is transitive.} \hskip.3em
Moreover, all counterexamples to transitivity can be restricted to
arguments with minimal sets of ground instances of defeasible rules.%
\end{theorem}

\yestop\noindent
As a consequence of \theoref{theorem P not transitive}, \hskip.2em
the respective relations in 
\cite
{Stolzenburg-etal-Computing-Specificity-2003} \hskip.1em and 
\cite
{Simari-Loui-Defeasible-Reasoning-1992} 
\hskip.1em are not transitive. 
This means
that these relations are not quasi-orderings, \hskip.2em
let alone reflexive orderings.

\begin{sloppypar}
 This consequence is immediate for the 
 relation \nlbmath\succeq\ 
 \cite[\litdefiref{10}, \p\,94]{Stolzenburg-etal-Computing-Specificity-2003} 
 \hskip.2em
 and for the relation
 \nlbmath{>_{\rm spec}} 
 \cite[\litdefiref{2.12}, \p 132]{Simari-Loui-Defeasible-Reasoning-1992}, 
 \hskip.2em
 simply because we can replace \nlbmath\succeq\ and  \nlbmath{>_{\rm spec}}
 with \nlbmath{\lesssim_{\rm P2}} and \nlbmath{<_{\rm P2}}
 in the context of \examref{example P not transitive}, respectively.

Although transitivity of these relations
is strongly suggested by the special choice of their symbols 
and seems to be taken for granted in general,
 we found an actual statement of such a transitivity only 
 for the relation 
 \nlbmath{\sqsupseteq} of \litdefiref{2.22} on 
 \litspageref{134} of 
 \hskip.1em\cite{Simari-Loui-Defeasible-Reasoning-1992}, \hskip.1em
 namely in \mbox{``\litlemmref{2.23}''} 
 \cite[\p 134]{Simari-Loui-Defeasible-Reasoning-1992}. \hskip.3em
According to the rules of good scientific and historiographic practice,
we pinpoint the violation
of this ``Lemma'' now as follows.
Non-transitivity of \nlbmath{\sqsupseteq}
follows here immediately from the non-transitivity
of the relation \nlbmath{\geq_{\rm spec}} of \litdefiref{2.15}, \hskip.2em
which, \hskip.1em
however, \hskip.1em
is not identical to the above-mentioned relation \nlbmaths\succeq, \hskip.2em
but actually a subset of \nlbmaths\succeq, \hskip.2em
because it is defined via a peculiar additional equivalence 
\nlbmath{\approx_{\rm spec}} introduced in 
\litdefiref{2.14} \cite[\p 132]{Simari-Loui-Defeasible-Reasoning-1992},
\hskip.3em
namely via 
\bigmathnlb{
\tight{\geq_{\rm spec}}
\nottight{\nottight{:=}}
\tight{>_{\rm spec}}\cup
\tight{\approx_{\rm spec}}}{}
\cite[\litdefiref{2.15}, \p 132\f]{Simari-Loui-Defeasible-Reasoning-1992}.
\hskip.3em
Directly from \litdefiref{2.14} of 
\cite{Simari-Loui-Defeasible-Reasoning-1992}, \hskip.2em
we get \bigmathnlb{\tight{\approx_{\rm spec}}\subseteq
\tight{\approx_{\rm P2}}}.
Thus, \hskip.1em
by \cororef{corollary ordering the P}, \hskip.2em
we get \bigmaths{
\tight{\geq_{\rm spec}}
\nottight{\nottight{\subseteq}}\tight{\lesssim_{\rm P2}}
\nottight{\nottight{\subseteq}}\tight{\lesssim_{\rm P1}}};
and so 
(recollecting 
\nlbmath{\tight{<_{\rm P2}}\subseteq\tight{>_{\rm spec}}
\subseteq\tight{\geq_{\rm spec}}}) 
the result
\par\noindent\LINEmaths{
\pair{\mathcal A_1}{L_1}
<_{\rm P2}
\pair{\mathcal A_2}{L_2}
<_{\rm P2}
\pair{\mathcal A_3}{L_3}
\not\gtrsim_{\rm P1}
\pair{\mathcal A_1}{L_1}}{}
\par\noindent of \lemmref{lemma P not transitive} 
and \cororef{corollary ordering the P} \hskip.1em
gives us the following counterexample to transitivity:
\par\noindent\LINEmaths{\pair{\mathcal A_1}{L_1}
\geq_{\rm spec}
\pair{\mathcal A_2}{L_2}
\geq_{\rm spec}
\pair{\mathcal A_3}{L_3}
\not\leq_{\rm spec}
\pair{\mathcal A_1}{L_1}}.
\end{sloppypar}
\yestop\yestop
\vfill\pagebreak
\subsection{Our Novel Specificity Ordering CP}\label
{subsection Our Novel Specificity Ordering CP}%

\yestop\yestop\noindent
In the previous section, 
we have seen that {\em minor corrections}\/ of \poole's original relation 
\nolinebreak P1 \hskip.1em
\mbox{(such as P2, P3)} \hskip.1em
do not cure the 
(up to our finding of \examref{example P not transitive}) \hskip.1em
hidden and even denied formal deficiency of these relations,
namely their lack of transitivity.

Please keep in mind,
however,
that our true motivation for overcoming this deficiency
by a {\em major correction}\/ of P3
was not this formal deficiency,
but actually an informal one, namely that it failed to 
get sufficiently close to human intuition,
which will become even more 
clear in \sectref{section Putting Specificity to Test}
than in \sectref{section Toward an Intuitive Notion of Specificity}.

Therefore,
in this section, 
we \nolinebreak now define our major \underline correction of 
\underline{\namefont P}{\namefont oole}'s specificity
---~the binary relation \math{\lesssim_{\rm CP}}~---
with implicit reference to our sets of facts and of general and defeasible rules
\mbox{(\ie\ to \maths{\Pioffont^{\rm F}\!},
\maths{\Pioffont^{\rm G}\!},
and \nlbmath\Deltaoffont, respectively)} \hskip.2em
as follows.

\yestop\begin{definition}[Our version of Specificity \math{\lesssim_{\rm CP}}]
\label{definition CP specificity}%
\\[+.5ex]\noindent\math{\pair{\mathcal A_1}{L_1}\lesssim_{\rm CP}
\pair{\mathcal A_2}{L_2}} \udiff\
\pair{\mathcal A_1}{L_1} and
\pair{\mathcal A_2}{L_2} 
are arguments, \hskip.2em
and we have\begin{enumerate}\noitem\item
\maths{L_1\tightin\mathfrak T_\Pioffont}{} \ or\noitem\item
\math{L_2\tightnotin\mathfrak T_\Pioffont} and 
every \math{H\subseteq\mathfrak T_{\Pioffont}} 
that is an 
\opt{minimal}\footnote{%
 Note that the omission of the optional restriction to {\em minimal}\/
 activation sets for \nlbmath{\pair{\mathcal A_1}{L_1}} 
 in \defiref{definition CP specificity} 
 has no effect on the extension of the defined notion, \hskip.1em
 simply because the additional 
 non-minimal activation sets \nlbmath{\pair{\mathcal A_1}{L_1}} 
 will then be 
 activation sets for \nlbmath{\pair{\mathcal A_2}{L_2}} {\it\afortiori}.%
} 
activation set for 
\nlbmath{\pair{\mathcal A_1}{L_1}} 
is also an activation set for 
\nlbmaths{\pair{\mathcal A_2}{L_2}}.\end{enumerate}\end{definition}

\yestop\yestop\noindent
The crucial change in \defiref{definition CP specificity} as compared 
to \defiref{DefSpec} is not the merely technical emphasis it puts on the case
``\math{\!L_1\tightin\mathfrak T_\Pioffont}\closequotecomma
which has no effect on the extension of the relation as compared to 
\nlbmath{\lesssim_{\rm P3}}. \
The crucial changes actually are\begin{enumerate}\noitem\item[(A)] 
the replacement of 
``\math{\!H\tightsubseteq\mathfrak T_{\Pioffont\cup\Deltaoffont}\!}'' 
with 
``\math{\!H\tightsubseteq\mathfrak T_\Pioffont}''
(as explained already in phase\,1 of 
 \nlbsectref{section where the phases are}), \hskip.3em
and the thereby enabled\noitem\item[(B)]
omission of the previously technically required,\footnote{%
 \majorheadroom
 See the discussion in \examref{example lovely one} \hskip.1em
 on why this condition is technically required for P1, P2, and P3.%
} \hskip.2em
but unintuitive negative 
condition on derivability
(of the form 
``but not a simplified activation set for \pair\emptyset{L_1}'').\noitem
\end{enumerate}
An additional minor change,
which we have already discussed in 
\sectref{section definition activation set}, \hskip.2em
is the one from simplified to (non-simplified) activation sets.

\yestop\begin{corollary}\label{corollary sub-argument CP} \
If\/ \pair{\mathcal A_1}{L_1},
\pair{\mathcal A_2}{L_2} 
are arguments and we have
\bigmaths{\mathcal A_1\,\tightsubseteq\,\mathcal A_2}{} and
\bigmathnlb{L_1\tightequal L_2},
then we have 
\bigmaths{\pair{\mathcal A_1}{L_1}
\lesssim_{\rm CP}
\pair{\mathcal A_2}{L_2}}{}.\end{corollary}

\vfill\pagebreak

\halftop\begin{theorem}\label{theorem CP quasi-ordering} \ 
\math{\lesssim_{\rm CP}} is a quasi-ordering on arguments.
\end{theorem}

\yestop\begin{proofparsepqed}{\theoref{theorem CP quasi-ordering}}
\math{\lesssim_{\rm CP}} is a reflexive relation on arguments
because of \cororef{corollary sub-argument CP}. \par
To show transitivity, 
let \nolinebreak us assume 
\par\noindent\LINEmaths{
\pair{\mathcal A_1}{L_1}
\lesssim_{\rm CP}
\pair{\mathcal A_2}{L_2}
\lesssim_{\rm CP}
\pair{\mathcal A_3}{L_3}}.
\par\noindent According to \defiref{definition CP specificity}, \hskip.2em
because of \bigmaths{
\pair{\mathcal A_1}{L_1}
\lesssim_{\rm CP}
\pair{\mathcal A_2}{L_2}
}, we have 
\maths{L_1\tightin\mathfrak T_\Pioffont}{}
\ ---~~and then immediately the desired 
\bigmaths{
\pair{\mathcal A_1}{L_1}
\lesssim_{\rm CP}
\pair{\mathcal A_3}{L_3}
}{}~--- \
or we have 
\math{L_2\tightnotin\mathfrak T_\Pioffont} \ and \
every \math{H\subseteq\mathfrak T_{\Pioffont}} 
that is an 
activation set for 
\nlbmath{\pair{\mathcal A_1}{L_1}} 
is also an activation set for 
\nlbmaths{\pair{\mathcal A_2}{L_2}}. \hskip.4em
The latter case excludes the first option in 
\defiref{definition CP specificity} 
as a justification for 
\bigmaths{
\pair{\mathcal A_2}{L_2}
\lesssim_{\rm CP}
\pair{\mathcal A_3}{L_3}
}, and thus 
we have 
\math{L_3\tightnotin\mathfrak T_\Pioffont} \ and \
every \math{H\subseteq\mathfrak T_{\Pioffont}} 
that is an 
activation set for 
\nlbmath{\pair{\mathcal A_2}{L_2}} 
is also an activation set for 
\nlbmaths{\pair{\mathcal A_3}{L_3}}. \
All in all,
we get that
every \math{H\subseteq\mathfrak T_{\Pioffont}} 
that is an 
activation set for 
\nlbmath{\pair{\mathcal A_1}{L_1}} 
is also an activation set for 
\nlbmaths{\pair{\mathcal A_3}{L_3}}. \
Thus, we get the desired 
\bigmaths{
\pair{\mathcal A_1}{L_1}
\lesssim_{\rm CP}
\pair{\mathcal A_3}{L_3}
}{} also in this case.\end{proofparsepqed}

\yestop\yestop\yestop\yestop
\noindent Obviously, 
an argument is ranked by \nlbmath{\lesssim_{\rm CP}} firstly 
on whether its literal is in \nlbmaths{\mathfrak T_\Pioffont},
and, if not, secondly on the set of its activation sets,
which is an element of the power set of the power set of 
\nlbmaths{\mathfrak T_\Pioffont}. \ So we get:
\begin{corollary}\label{corollary wellfounded} \
If\/ \math{\mathfrak T_\Pioffont} is finite, \hskip.2em
then\/ \math{<_{\rm CP}} 
is \wellfounded.\end{corollary}
\vfill\pagebreak
\yestop\subsection{Relation between the Specificity Relations P3 and CP}
\yestop\halftop\begin{theorem}\label{theorem super quasi-ordering}
\\[+.9ex]%
Set\/ \math{\Pioffont^{<2}} to be 
the set of rules from\/ \nlbmath\Pioffont\
that are unconditional or 
have exactly one literal in the conjunction of their condition.
\par\noindent
Set\/ \math{\Pioffont^{\geq 2}} to be the set of rules from\/ \nlbmath\Pioffont\
with more than one literal in their condition.
\par\noindent\math{\tight{\lesssim_{\rm P3}}\subseteq\tight{\lesssim_{\rm CP}}}
holds if one (or more) 
of the following conditions hold:\begin{enumerate}\noitem\item
For every \math{H\subseteq\mathfrak T_\Pioffont} 
and for every set \nlbmath{\mathcal A} of ground instances of rules from 
\nlbmaths\Deltaoffont, \hskip.2em
and for \\\math{\mbox{}\,\mathfrak L:=
\mathfrak T_{H\cup\mathcal A\cup\Pioffont^{\rm G}}}, \
we have \bigmaths
{\mathfrak T_{\,\mathfrak L\,\cup\Pioffont}\subseteq
\mathfrak L\cup\mathfrak T_\Pioffont}.\noitem\item
For each
instance\/ \math{L\antiimplies L_0'\tightund\ldots\tightund L_{n+1}'}
of each rule in\/ \nlbmath{\Pioffont^{\geq 2}}
with \maths{L\tightnotin\mathfrak T_{\Pioffont^{<2}}}, \hskip.3em 
\\we have\/ \math{L_j'\notin\mathfrak T_{\Pioffont^{<2}}}
for all \nlbmath{j\in\{0,\ldots,n\tight+1\}}.\noitem\item
For each
instance\/ \math{L\antiimplies L_0'\tightund\ldots\tightund L_{n+1}'}
of each rule in\/ \nlbmaths{\Pioffont^{\geq 2}\!}, \hskip.3em
\\we have\/ \math{L_j'\notin\mathfrak T_\Pioffont}
for all \nlbmath{j\in\{0,\ldots,n\tight+1\}}.\noitem\item
We have \bigmaths{\Pioffont^{\geq 2}=\emptyset}.\end{enumerate}\end{theorem}

\halftop\noindent
Note that 
if we had improved \nlbmath{\lesssim_{\rm P3}} only \wrt\ 
phase\,1 of \sectref{section where the phases are}, \hskip.2em
but not \wrt\ phase\,3 in addition, then 
the condition of \theoref{theorem super quasi-ordering}
would not have been required at all.
This means that a condition becomes necessary by our correction of
simplified activation sets to non-simplified ones,
but not because of the major changes (A) and (B) of 
\nlbsectref{subsection Our Novel Specificity Ordering CP}.

\halftop\halftop\halftop\halftop
\begin{proofparsepqed}{\theoref{theorem super quasi-ordering}}
First let us show that condition\,2 implies condition\,1. \hskip.3em
To this end, \hskip.1em
let \maths{H\subseteq\mathfrak T_\Pioffont}, \hskip.1em
let \math{\mathcal A} \nolinebreak be a set of ground instances of rules from 
\nlbmaths\Deltaoffont, \hskip.1em
and set \math{\mathfrak L:=
\mathfrak T_{H\cup\mathcal A\cup\Pioffont^{\rm G}}}. \hskip.3em
For an \mbox\reductioadabsurdum, \hskip.3em
let us assume \bigmathnlb{\mathfrak T_{\,\mathfrak L\,\cup\Pioffont}
\nsubseteq\mathfrak L\cup
\mathfrak T_\Pioffont}. \hskip.3em
Because of \bigmaths{\Pioffont^{\rm F}\tightsubseteq\,
\mathfrak T_{\Pioffont^{<2}}},
we have 
\bigmathnlb{\mathfrak L\,\tightcup\,\Pioffont\nottight{\nottight=}
\mathfrak L\,\tightcup\,\Pioffont^{\rm F}\,\tightcup\,\Pioffont^{\rm G}\!
\nottight{\nottight\subseteq}
\mathfrak L\,\tightcup\,\mathfrak T_{\Pioffont^{<2}}
\,\tightcup\,\Pioffont^{\rm G}\!},
and thus
\bigmaths{\mathfrak T_{\,\mathfrak L\,\cup\Pioffont}\subseteq
\mathfrak 
T_{\,\mathfrak L\,\cup\,\mathfrak T_{\Pioffont^{<2}}\,\cup\,\Pioffont^{\rm G}}},
and thus
\bigmathnlb{\mathfrak 
T_{\,\mathfrak L\,\cup\,\mathfrak T_{\Pioffont^{<2}}\,\cup\,\Pioffont^{\rm G}}
\nsubseteq\mathfrak L\cup\mathfrak T_{\Pioffont^{<2}}}{}
(because otherwise \maths{\mathfrak T_{\,\mathfrak L\,\cup\Pioffont}\subseteq
T_{\,\mathfrak L\,\cup\,\mathfrak T_{\Pioffont^{<2}}\,\cup\,\Pioffont^{\rm G}}
\subseteq\mathfrak L\cup\mathfrak T_{\Pioffont^{<2}}
\subseteq\mathfrak L\cup\mathfrak T_\Pioffont
}). \
Now \math{\mathfrak L} is closed under \nlbmath{\Pioffont^{\rm G}} 
by definition.
Moreover, 
\math{\mathfrak T_{\Pioffont^{<2}}} 
is closed under \nlbmath{\Pioffont^{<2}} by definition
and under \nlbmath{\Pioffont^{\geq 2}} by  condition\,2. \hskip.3em
Because
both of the sets of literals \math{\mathfrak L} 
and \nlbmath{\mathfrak T_{\Pioffont^{<2}}} 
are closed under \nlbmath{\Pioffont^{\rm G}} 
---~but nevertheless their union is not closed under \nlbmath{\Pioffont^{\rm G}} 
according to
\mbox{\math{\mathfrak 
T_{\,\mathfrak L\,\cup\,\mathfrak T_{\Pioffont^{<2}}\,\cup\,\Pioffont^{\rm G}}
\nsubseteq\mathfrak L\cup\mathfrak T_{\Pioffont^{<2}}}~---}
there must be an inference step {\em essentially based on both sets in parallel}.
\hskip.3em
More precisely, \hskip.1em
this means that there must be an instance 
\bigmathnlb{L\antiimplies L_1'\tightund\ldots\tightund L_n'}{}
of a rule from \nlbmath{\Pioffont^{\rm G}} \hskip.1em
with
\bigmaths{L\notin\mathfrak L\cup\mathfrak T_{\Pioffont^{<2}}}, 
and some \math{i,j\in\{1,\ldots,n\}}
with \bigmaths{L_i'\in\mathfrak L\tightsetminus\mathfrak T_{\Pioffont^{<2}}}{}
and \bigmaths{L_j'\in\mathfrak T_{\Pioffont^{<2}}\tightsetminus\mathfrak L}.
\hskip.3em
Then \bigmathnlb{L\antiimplies L_1'\tightund\ldots\tightund L_n'}{}
must actually be an instance of a rule from \nlbmaths{\Pioffont^{\geq 2}\!},
\hskip.3em  and \bigmaths{L\tightnotin\mathfrak T_{\Pioffont^{<2}}}, 
but \bigmathnlb{L_j'\tightin\mathfrak T_{\Pioffont^{<2}}}{}
in contradiction to condition\,2.

As condition\,2 implies condition\,1, condition\,3 trivially implies
condition\,2, and condition\,4 trivially implies condition\,3, \hskip.1em
it now suffices to show the claim
\bigmaths{\pair{\mathcal A_1}{L_1}\lesssim_{\rm CP}
\pair{\mathcal A_2}{L_2}}{}
under condition\,1 and 
the initial assumption of \hskip.2em\maths{\pair{\mathcal A_1}{L_1}
\lesssim_{\rm P3}\pair{\mathcal A_2}{L_2}}. \hskip.5em
By this assumption, \pair{\mathcal A_1}{L_1} and
\pair{\mathcal A_2}{L_2} 
are arguments and 
\bigmathnlb{L_2\tightin\mathfrak T_\Pioffont}{} implies
\bigmathnlb{L_1\tightin\mathfrak T_\Pioffont}. 

\pagebreak

If \bigmaths{L_1\tightin\mathfrak T_\Pioffont}{} holds, \
then our claim holds \aswell. 

Otherwise, 
we have \bigmaths{L_1,L_2\tightnotin\mathfrak T_\Pioffont},
and it \nolinebreak suffices to show 
the sub-claim that \math H is an activation set for
\nlbmath{\pair{\mathcal A_2}{L_2}}
under the additional sub-assumption that
\math{H\subseteq\mathfrak T_{\Pioffont}} 
is an activation set for 
\nlbmaths{\pair{\mathcal A_1}{L_1}}. \
Under the sub-assumption we \nolinebreak also have \bigmaths{H\,\tightsubseteq\,
\mathfrak T_{\Pioffont\cup\Deltaoffont}}{}
because of \bigmaths
{\mathfrak T_\Pioffont\,\tightsubseteq\,\mathfrak T_{\Pioffont\cup\Deltaoffont}},
and, \hskip.2em
for \maths{\mathfrak L:=
\mathfrak T_{H\cup\mathcal A_1\cup\Pioffont^{\rm G}}}, \
we have \bigmaths{L_1\tightin\mathfrak T_{\,\mathfrak L\,\cup\,\Pioffont}},
and then, \hskip.2em
by condition\,1,
\bigmaths{L_1\in\mathfrak L\cup\mathfrak T_\Pioffont},
and then, \hskip.2em
by our current case of 
\maths{L_1,L_2\tightnotin\mathfrak T_\Pioffont}, \hskip.2em
we have \maths{L_1\tightin\mathfrak L}. \hskip.4em
Thus, \math H is a {\em simplified}\/ activation set for 
\pair{\mathcal A_1}{L_1}. 

Let us now provide an \reductioadabsurdum\/ for 
the assumption that \math H is a simplified activation set also 
for \pair\emptyset{L_1}: \
Then we \nolinebreak would have 
\bigmaths{L_1\tightin\mathfrak T_{H\cup\Pioffont^{\rm G}}},
and because of 
\bigmath{H\,\tightsubseteq\,\mathfrak T_{\Pioffont}} 
and
\bigmath{\Pioffont^{\rm G}\,\tightsubseteq\,\Pioffont} 
we get
\bigmaths{L_1\tightin\mathfrak T_{\mathfrak T_\Pioffont\cup\Pioffont}
=\mathfrak T_\Pioffont
}{}
\mbox{---~~a~contradiction} to our current case of 
\maths{L_1,L_2\tightnotin\mathfrak T_\Pioffont}. \ \par
All in all,
by our initial assumption,
\math H must now be a simplified activation set for 
\pair{\mathcal A_2}{L_2} and,
{\it\afortiori}, an activation set for 
\nlbmaths{\pair{\mathcal A_2}{L_2}}, \hskip.2em
as \nolinebreak was to be shown for 
our only remaining sub-claim.\end{proofparsepqed}

\yestop\yestop\yestop\yestop\yestop\noindent
Finally, 
with the help of \theoref{theorem super quasi-ordering}, \hskip.2em
we can now 
analyze \examref{example poole 1 precisely}, \hskip.2em
and also
check how our relation CP behaves in case of our
counterexample for transitivity:

\halftop
\begin{example}[\litexamref 1 of \cite{Poole-Preferring-Most-Specific-1985}]
\hfill{\em(continuing \examref{example poole 1 precisely})}
\label{example poole 1 precisely discussion}
\\\noindent We have 
\bigmaths{\pair{\mathcal A_2}{\ident{flies}(\ident{edna})}\not\lesssim_{\rm CP}
\pair\emptyset{\neg\ident{flies}(\ident{edna})}}{}
\\\getittotheright{because \bigmaths{\ident{flies}(\ident{edna})\tightnotin
\mathfrak T_{\Pioffont_{\ref{example poole 1 precisely}}}}{}
and \ \maths
{\neg\ident{flies}(\ident{edna})\tightin\mathfrak T_{\Pioffont_{\ref{example poole 1 precisely}}}}.}%
\\We have \bigmaths{\pair\emptyset{\neg\ident{flies}(\ident{edna})}
\lesssim_{\rm P3}\pair{\mathcal A_2}{\ident{flies}(\ident{edna})}}, 
\\because \ \maths{\neg\ident{flies}(\ident{edna})\tightin
\mathfrak T_{\Pioffont_{\ref{example poole 1 precisely}}}} \ and because 
the premise of the last condition in \defiref{DefSpec}
is contradictory for 
\maths{\mathcal A_1:=\emptyset}, \hskip.2em 
and cannot be satisfied by any set 
\nlbmaths{H\subseteq\mathfrak T_{\Pioffont_{\ref{example poole 1 precisely}}
\cup\Deltaoffont_{\ref{example poole 1 precisely}}}}.
\\All in all, by \theoref{theorem super quasi-ordering}
(where condition\,4 is satisfied),
\\\math{\begin{array}[t]{@{}r r@{}c@{}l}\mbox{we get}
 &\pair\emptyset{\neg\ident{flies}(\ident{edna})}
 &<_{\rm CP}
 &\pair{\mathcal A_2}{\ident{flies}(\ident{edna})}
\\\mbox{and}
 &\pair\emptyset{\neg\ident{flies}(\ident{edna})}
 &<_{\rm P3}
 &\pair{\mathcal A_2}{\ident{flies}(\ident{edna})}.
\\\end{array}}\halftop\end{example}
\begin{example}\hfill{\em(continuing \examref{example P not transitive})}
\label{example P not transitive discussion}\sloppy
\\The following holds
for our specification of \examref{example P not transitive} 
by \lemmref{lemma P not transitive}
and \cororef{corollary ordering the P}:
\par\noindent\LINEmaths{
\pair{\mathcal A_1}{\ident{beer}}
<_{\rm P3}
\pair{\mathcal A_2}{\ident{wine}}
<_{\rm P3}
\pair{\mathcal A_3}{\ident{vodka}}
\not\gtrsim_{\rm P3}
\pair{\mathcal A_1}{\ident{beer}}}.
\\\noindent We have now:
\\\noindent\LINEmaths{
\pair{\mathcal A_1}{\ident{beer}}
<_{\rm CP}
\pair{\mathcal A_2}{\ident{wine}}
<_{\rm CP}
\pair{\mathcal A_3}{\ident{vodka}}
>_{\rm CP}
\pair{\mathcal A_1}{\ident{beer}}
}{} 
\par\noindent simply because the trouble-making set 
\math{\{\ident e\}} is not to be considered here: \hskip.1em
it is not 
a subset of \nolinebreak\hskip.2em
\nlbmaths{\mathfrak T_{\Pioffont_{\ref{example P not transitive}}}}{}
at all! \hskip.4em
The checking of the details is left to the reader. \hskip.3em
Note that, because of 
\lemmref{lemma P not transitive}, \hskip.2em
\theoref{theorem super quasi-ordering} 
(where condition\,4 is satisfied), \hskip.2em
\theoref{theorem CP quasi-ordering}, \hskip.2em
and \cororef{corollary quasi-orderings}, \hskip.2em
all that is actually left to show is
\par\noindent\LINEmaths{
\pair{\mathcal A_1}{\ident{beer}}
\not\gtrsim_{\rm CP}
\pair{\mathcal A_2}{\ident{wine}}
\not\gtrsim_{\rm CP}
\pair{\mathcal A_3}{\ident{vodka}}.}{~~~~~~~~~~~~~~~~~~\mbox{}}
\end{example}
\vfill\pagebreak
\section{Putting Specificity to Test \wrt\ Human Intuition}\label
{section Putting Specificity to Test}

\yestop\noindent
Before we will go on with further conceptual material 
and efficiency considerations 
in \sectref{section Efficiency Considerations}, \hskip.2em
let us put the two notions of specificity
\ ---~~as formalized in the two binary relations 
\math{\lesssim_{\rm P3}} 
and
\math{\lesssim_{\rm CP}}~~--- \
to test \wrt\ our changed phase\,1 of \sectref{section where the phases are}
in a series of classical examples. \hskip.3em

Note that we can freely apply \theoref{theorem super quasi-ordering}
because at least one\footnote{%
 Condition\,4 of \theoref{theorem super quasi-ordering}
 is satisfied for \examrefsss
 {example poole 2}
 {example poole 3}
 {example lovely one}
 {example stolzenburg 1}.
 Condition\,3 (but not condition\,4) is satisfied for \examrefssss
 {example poole 6}
 {example variation 1 poole 6}
 {variation 2 example poole 6}
 {example stolzenburg 2}
 {example variation 3 poole 6}.
} 
of its conditions is satisfied
in all the following examples except \examref{example stolzenburg 3},
which is the only example where we have an activation set that is
actually not a simplified one.
 
Besides freely applying \theoref{theorem super quasi-ordering}
---~to enable the reader to make his own selection of interesting examples
without problems of understanding~---
we are pretty explicit in all of the following examples.
\yestop\subsection{Implementation of the Preference of the ``More Concise''}
\yestop
\noindent First let us see what happens to \examref{example poole 2}
\hskip.1em if we are not so certain anymore whether no emu can fly.%

\begin{example}%
[\litexamref 2 of \cite{Poole-Preferring-Most-Specific-1985}]
\label{example poole 2}%
\\[-3ex]\math{\begin{array}[t]{@{}l l l@{}}
\\\Pioffont^{\rm F}_{\ref{example poole 2}}
 &:=
 &\Pioffont^{\rm F}_{\ref{example poole 1 precisely}}
\\\Pioffont^{\rm G}_{\ref{example poole 2}}
 &:=
 &\left\{\begin{array}{l}\ident{bird}(x)\antiimplies\ident{emu}(x)
   \\\end{array}\right\}, 
\\\Deltaoffont_{\ref{example poole 2}}
 &:=
 &\left\{\begin{array}{l}
     \neg\ident{flies}(x)\defeasibleantiimplies\ident{emu}(x),
   \\\ident{flies}(x)\defeasibleantiimplies\ident{bird}(x)    
   \\\end{array}\right\},
\\\mathcal A_1
 &:=
 &\left\{\begin{array}{l}
      \neg\ident{flies}(\ident{edna})
      \defeasibleantiimplies\ident{emu}(\ident{edna})
   \\\end{array}\right\},
\\\mathcal A_2
 &:=
 &\left\{\begin{array}{l}
      \ident{flies}(\ident{edna})
      \defeasibleantiimplies\ident{bird}(\ident{edna})
    \\\end{array}\right\}.
\\\end{array}}\hfill
\begin{tikzpicture}[baseline=(current bounding box.north),
>=stealth,->,looseness=.5,auto]
\matrix [matrix of math nodes,
column sep={1.4cm,between origins},
row sep={1.3cm,between origins}]{
   |(notfliesedna)| \math{\neg\ident{flies}(\ident{edna})} 
 &
 & |(fliesedna)|    \math{\ident{flies}(\ident{edna})} 
 &
 & |(fliestweety)|  \math{\ident{flies}(\ident{tweety})} 
\\ 
 &
 & |(birdedna)|     \math{\ident{bird}(\ident{edna})}
 &  
 & |(birdtweety)|   \math{\ident{bird}(\ident{tweety})}  
\\
 & |(emuedna)|      \math{\ident{emu}(\ident{edna})}  
 &
 & |(TRUE)|         \math{\TRUEpp}  
\\
};
\begin{scope}[every node/.style={midway,auto}]
\draw [] (birdedna) -- node [right] 
      {$\scriptscriptstyle\!\!\mathcal A_2$} 
      (fliesedna);
\draw [] (birdtweety) -- (fliestweety);
\draw [double] (emuedna) -- (birdedna);
\draw [] (emuedna) -- node [right] 
      {$\scriptscriptstyle\!\mathcal A_1$} 
      (notfliesedna);
\draw [double] (TRUE) -- (emuedna);
\draw [double] (TRUE) -- (birdtweety);
\end{scope}
\end{tikzpicture}
\\\noindent Let us compare the specificity of the arguments
\pair{\mathcal A_1}{\neg\ident{flies}(\ident{edna})}
and 
\pair{\mathcal A_2}{\ident{flies}(\ident{edna})}.
\\\noindent We have
\bigmaths{\begin{array}[t]{l l l}
  \mathfrak T_{\Pioffont_{\ref{example poole 2}}}
 &=
 &\{\ident{bird}(\ident{tweety}),
  \ident{emu}(\ident{edna}),
  \ident{bird}(\ident{edna})
\},
\\\mathfrak T_{\Pioffont_{\ref{example poole 2}}
        \cup\Deltaoffont_{\ref{example poole 2}}}
 &=
 &\{\neg\ident{flies}(\ident{edna}),
    \ident{flies}(\ident{edna}),\ident{flies}(\ident{tweety})\}\cup
  \mathfrak T_{\Pioffont_{\ref{example poole 2}}}.
\\\end{array}}{}
\\\noindent We have 
\bigmaths{\pair{\mathcal A_2}{\ident{flies}(\ident{edna})}
\not\lesssim_{\rm CP}
\pair{\mathcal A_1}{\neg\ident{flies}(\ident{edna})}}{}
because \bigmaths{\ident{flies}(\ident{edna})\tightnotin
 \mathfrak T_{\Pioffont_{\ref{example poole 2}}}}{}
and because
\math{\{\ident{bird}(\ident{edna})\}\subseteq
 \mathfrak T_{\Pioffont_{\ref{example poole 2}}}}
is an activation set for \pair{\mathcal A_2}{\ident{flies}(\ident{edna})},
but not for \pair{\mathcal A_1}{\neg\ident{flies}(\ident{edna})}.
\\We have \bigmaths{\pair{\mathcal A_1}{\neg\ident{flies}(\ident{edna})}
\lesssim_{\rm P3}\pair{\mathcal A_2}{\ident{flies}(\ident{edna})}}, 
because \bigmaths{\ident{flies}(\ident{edna})\tightnotin
 \mathfrak T_{\Pioffont_{\ref{example poole 2}}}}{}
and because, \hskip.2em
if \hskip.1em\maths{H\subseteq\mathfrak T_{{\Pioffont_{\ref{example poole 2}}}
\cup\Deltaoffont_{\ref{example poole 2}}}}{} 
is a simplified activation set for 
\pair{\mathcal A_1}{\neg\ident{flies}(\ident{edna})},
but not for \pair\emptyset{\neg\ident{flies}(\ident{edna})}, \hskip.2em
then we have \,\maths{\ident{emu}(\ident{edna})\tightin H},\,
and thus \math H is a simplified activation set also for 
\pair{\mathcal A_2}{\ident{flies}(\ident{edna})}.
\\All in all, by \theoref{theorem super quasi-ordering},
\math{\begin{array}[t]{@{}r r@{}c@{}l}\mbox{we get}
 &\pair{\mathcal A_1}{\neg\ident{flies}(\ident{edna})}
 &<_{\rm CP}
 &\pair{\mathcal A_2}{\ident{flies}(\ident{edna})}
\\\mbox{and}
 &\pair{\mathcal A_1}{\neg\ident{flies}(\ident{edna})}
 &<_{\rm P3}
 &\pair{\mathcal A_2}{\ident{flies}(\ident{edna})}.
\\\end{array}}\end{example}
\pagebreak
\begin{example}[Renamed Subsystem of \litexamref 3 of \cite
{Poole-Preferring-Most-Specific-1985}]
\label{example poole 3}
\\[-2ex]\math{\begin{array}[t]{@{}l l l}
\\[-.8ex]\Pioffont^{\rm F}_{\ref{example poole 3}}
 &:=
 &\left\{\begin{array}{l}
      \ident{emu}(\ident{edna})
    \\\end{array}\right\},
\quad\quad\Pioffont^{\rm G}_{\ref{example poole 3}}:=\emptyset,
\\\Deltaoffont_{\ref{example poole 3}}
 &:=
 &\left\{\begin{array}{l}
      \neg\ident{flies}(x)\defeasibleantiimplies\ident{emu}(x),
    \\\ident{flies}(x)\defeasibleantiimplies\ident{bird}(x),
    \\\ident{bird}(x)\defeasibleantiimplies\ident{emu}(x)
    \\\end{array}\right\},
\\\mathcal A_1
 &:=
 &\left\{\begin{array}{l}
      \neg\ident{flies}(\ident{edna})
      \defeasibleantiimplies\ident{emu}(\ident{edna})
    \\\end{array}\right\},
\\\mathcal A_2
 &:=
 &\left\{\begin{array}{l}
      \ident{flies}(\ident{edna})
      \defeasibleantiimplies\ident{bird}(\ident{edna}),
    \\\ident{bird}(\ident{edna})
      \defeasibleantiimplies\ident{emu}(\ident{edna})
  \\\end{array}\right\}.
\\\end{array}}\hfill
\begin{tikzpicture}[baseline=(current bounding box.north),
>=stealth,->,looseness=.5,auto]
\matrix [matrix of math nodes,
column sep={1.4cm,between origins},
row sep={1.3cm,between origins}]{
   |(notfliesedna)| \math{\neg\ident{flies}(\ident{edna})} 
 &
 & |(fliesedna)| \math{\ident{flies}(\ident{edna})} 
\\ 
 &
 & |(birdedna)| \math{\ident{bird}(\ident{edna})}
\\
 & |(emuedna)| \math{\ident{emu}(\ident{edna})}  
 &
 & |(TRUE)| \math{\TRUEpp}  
\\
};
\begin{scope}[every node/.style={midway,auto}]
\draw [] (birdedna) -- node [right] 
      {$\scriptscriptstyle\!\!\mathcal A_2$} 
      (fliesedna);
\draw [] (emuedna) -- node [right] 
      {$\scriptscriptstyle\!\!\mathcal A_2$} 
      (birdedna);
\draw [] (emuedna) -- node [right] 
      {$\scriptscriptstyle\!\mathcal A_1$} 
      (notfliesedna);
\draw [double] (TRUE) -- (emuedna);
\end{scope}
\end{tikzpicture}
\\\noindent Let us compare the specificity of the arguments
\pair{\mathcal A_1}{\neg\ident{flies}(\ident{edna})}
and \pair{\mathcal A_2}{\ident{flies}(\ident{edna})}.
\par\noindent\LINEmaths{\mathfrak T_{\Pioffont_{\ref{example poole 3}}}=
\{\ident{emu}(\ident{edna})\},
\hfill
\mathfrak T_{{\Pioffont_{\ref{example poole 3}}}\cup{\Deltaoffont_{\ref{example poole 3}}}}=
\{\ident{bird}(\ident{edna}),\ident{flies}(\ident{edna}),
\neg\ident{flies}(\ident{edna})\}\cup\mathfrak T_{\Pioffont_{\ref{example poole 3}}}}.
\par\noindent Now, however, we have 
\bigmaths{\pair{\mathcal A_2}{\ident{flies}(\ident{edna})}
\lesssim_{\rm CP}
\pair{\mathcal A_1}{\neg\ident{flies}(\ident{edna})}}{}
because \bigmaths{\neg\ident{flies}(\ident{edna})\tightnotin
\mathfrak T_{\Pioffont_{\ref{example poole 3}}}}{}
and, \
for every activation set 
\math{H\subseteq\mathfrak T_{\Pioffont_{\ref{example poole 3}}}} 
for \pair{\mathcal A_2}{\ident{flies}(\ident{edna})}, \ 
we get \bigmaths{\ident{emu}(\ident{edna})\tightin H},
and so \math H is an activation set also for 
\pair{\mathcal A_1}{\neg\ident{flies}(\ident{edna})}. \
Nevertheless, 
we still have \bigmaths{\pair{\mathcal A_2}{\ident{flies}(\ident{edna})}
\not\lesssim_{\rm P3}\pair{\mathcal A_1}{\neg\ident{flies}(\ident{edna})}},
because \math{\{\ident{bird}(\ident{edna})\}\subseteq\mathfrak T_{{
\Pioffont_{\ref{example poole 3}}}\cup{\Deltaoffont_{\ref{example poole 3}}}}} 
is a simplified activation set for 
\pair{\mathcal A_2}{\ident{flies}(\ident{edna})}, \ 
but neither for \pair\emptyset{\ident{flies}(\ident{edna})}, \
nor for \pair{\mathcal A_1}{\neg\ident{flies}(\ident{edna})}. \ 
\\We have \bigmaths{\pair{\mathcal A_1}{\neg\ident{flies}(\ident{edna})}
\lesssim_{\rm P3}
\pair{\mathcal A_2}{\ident{flies}(\ident{edna})}}, 
because of \bigmaths{\ident{flies}(\ident{edna})\tightnotin
\mathfrak T_{\Pioffont_{\ref{example poole 3}}}}{}
and because, \hskip.2em
if \hskip.1em\nlbmaths{H\subseteq\mathfrak T_{{\Pioffont_{\ref{example poole 3}}}
\cup\Deltaoffont_{\ref{example poole 3}}}}{} 
is a simplified activation set for 
\pair{\mathcal A_1}{\neg\ident{flies}(\ident{edna})}, \hskip.1em
but not for \pair\emptyset{\neg\ident{flies}(\ident{edna})}, \hskip.2em
then we have \ \maths{\ident{emu}(\ident{edna})\tightin H} \
and thus \math H is a simplified activation set also for 
\pair{\mathcal A_2}{\ident{flies}(\ident{edna})}.
\\All in all, by \theoref{theorem super quasi-ordering},
\math{\begin{array}[t]{@{}r r@{}c@{}l}\mbox{this time we get}
 &\pair{\mathcal A_1}{\neg\ident{flies}(\ident{edna})}
 &\approx_{\rm CP}
 &\pair{\mathcal A_2}{\ident{flies}(\ident{edna})}
\\\mbox{and}
 &\pair{\mathcal A_1}{\neg\ident{flies}(\ident{edna})}
 &<_{\rm P3}
 &\pair{\mathcal A_2}{\ident{flies}(\ident{edna})}.
\\\end{array}}
\\From a conceptual point of view, we have to ask ourselves,
  whether we would like a {\em defeasible}\/ rule instance
  such as \bigmaths{\ident{bird}(\ident{edna})
  \defeasibleantiimplies\ident{emu}(\ident{edna})}{}
  to reduce the specificity of \math{\mathcal A_2} as compared to 
  a system that seems equivalent for the given argument for 
  \nlbmaths{\ident{flies}(\ident{edna})}, \ 
  namely the argument 
  \pair{\{\ident{flies}(\ident{edna})
        \defeasibleantiimplies\ident{emu}(\ident{edna})\}}
       {\ident{flies}(\ident{edna})}\,? \ 
  Does the specificity of a defeasible reasoning
  step really reduce if we introduce intermediate literals?
\\According to human intuition, this question has a negative answer,
  as we have already explained
  at the end of \sectref{subsubsection Preference of  the ``More Concise''}.
  \hskip.3em
  Moreover, \examref{example poole 6} will exhibit a strong reason
  to deny it. \hskip.1em
\\Finally, 
  see \examref{variation 2 example poole 6} \hskip.1em
  for another example that makes even clearer
  why defeasible rules should be considered for their global semantical
  effect instead of their syntactical fine structure.%
\end{example}\pagebreak
\subsection{Monotonicity of Preference \wrt\ Conjunction}\label
{subsection Monotonicity of Preference}%
\begin{example}[\litexamref 6 of \cite{Poole-Preferring-Most-Specific-1985}]
\label{example poole 6}
\\[-3ex]\math{\begin{array}[t]{@{}l l l@{}}
\\\Pioffont^{\rm F}_{\ref{example poole 6}}
 &:=
 &\left\{\begin{array}{l}\ident a, 
    \\\ident d
    \\\end{array}\right\},
\\\Pioffont^{\rm G}_{\ref{example poole 6}}
 &:=
 &\left\{\begin{array}{l}\ident g_1\antiimplies\neg\ident c\tightund\neg\ident f,
    \\\ident g_2\antiimplies\ident c\tightund\ident f
    \\\end{array}\right\},
\\\Deltaoffont_{\ref{example poole 6}}
 &:=
 &\mathcal A_1\tightcup\mathcal A_2, 
\\\mathcal A_1
 &:=
 &\left\{\begin{array}{l}
      \neg\ident c\defeasibleantiimplies\ident a,
    \\\neg\ident f\defeasibleantiimplies\ident d
    \\\end{array}\right\},
\\\mathcal A_2
 &:=
 &\left\{\begin{array}{l}
      \ident b\defeasibleantiimplies\ident a,
    \\\ident c\defeasibleantiimplies\ident b,
    \\\ident e\defeasibleantiimplies\ident d,
    \\\ident f\defeasibleantiimplies\ident e
    \\\end{array}\right\}
\\\end{array}}\hfill
\begin{tikzpicture}[baseline=(current bounding box.north),
>=stealth,->,looseness=.5,auto]
\matrix [matrix of math nodes,
column sep={1.4cm,between origins},
row sep={1.3cm,between origins}]{
 &
 &
 & |(geins)| \math{\ident g_1}
\\
 &
 &
 & |(gzwei)| \math{\ident g_2}
\\ |(notc)|  \math{\neg\ident c} 
 &
 & |(c)|     \math{\ident c} 
 &
 & |(f)|     \math{\ident f}
 &
 & |(notf)|  \math{\neg\ident f} 
\\ 
 &
 & |(b)|     \math{\ident b}
 &
 & |(e)|     \math{\ident e}
\\
 & |(a)|     \math{\ident a}
 &
 & |(TRUE)|  \math{\TRUEpp}  
 &
 & |(d)|     \math{\ident d}
\\
};
\begin{scope}[every node/.style={midway,auto}]
\draw [] (b) -- node [right] 
      {$\scriptscriptstyle\!\!\mathcal A_2$} 
      (c);
\draw [] (a) -- node [right] 
      {$\scriptscriptstyle\!\!\mathcal A_2$} 
      (b);
\draw [] (a) -- node [right] 
      {$\scriptscriptstyle\!\mathcal A_1$} 
      (notc);
\draw [] (e) -- node [right] 
      {$\scriptscriptstyle\!\!\mathcal A_2$} 
      (f);
\draw [] (d) -- node [right] 
      {$\scriptscriptstyle\!\mathcal A_2$} 
      (e);
\draw [] (d) -- node [right] 
      {$\scriptscriptstyle\!\!\mathcal A_1$} 
      (notf);
\draw [double] (TRUE) -- (a);
\draw [double] (TRUE) -- (d);
\draw [double] (notc) -- (geins);
\draw [double] (notf) -- (geins);
\draw [double] (c)    -- (gzwei);
\draw [double] (f)    -- (gzwei);
\end{scope}
\end{tikzpicture}
\\\noindent Let us compare the specificity of the arguments
\pair{\mathcal A_1}{\ident g_1} and \pair{\mathcal A_2}{\ident g_2}.
\par\noindent\LINEmaths
{\mathfrak T_{\Pioffont_{\ref{example poole 6}}}=\{\ident a, \ident d\}
\hfill\mathfrak 
T_{\Pioffont_{\ref{example poole 6}}\cup\Deltaoffont_{\ref{example poole 6}}}=
\{\ident b,\ident c,\ident e,\ident f,
  \ident g_1,\ident g_2,\neg\ident c,\neg\ident f\}
\cup\mathfrak T_{\Pioffont_{\ref{example poole 6}}}}.
\par\noindent We have 
\maths{\pair{\mathcal A_1}{\ident g_1}\approx_{\rm CP}
\pair{\mathcal A_2}{\ident g_2}}{}
because \math{H\subseteq\mathfrak T_{\Pioffont_{\ref{example poole 6}}}}
is an activation set for \pair{\mathcal A_i}{\ident g_i}
if and only 
\linebreak if
\maths{H\tightequal\{\ident a,\ident d\}}. \hskip.2em
We have
\bigmaths{
\pair{\mathcal A_1}{\ident g_1}
\vartriangle_{\rm P3}
\pair{\mathcal A_2}{\ident g_2}
}{} for the following reasons: \
\math{\{\ident a,\neg\ident f\}\subseteq\mathfrak T_
{\Pioffont_{\ref{example poole 6}}\cup\Deltaoffont_{\ref{example poole 6}}}}
is a simplified activation set for \pair{\mathcal A_1}{\ident g_1},
but neither for \pair\emptyset{\ident g_1},
nor for \pair{\mathcal A_2}{\ident g_2}. \
\math{\{\ident a,\ident f\}\subseteq\mathfrak T_
{\Pioffont_{\ref{example poole 6}}\cup\Deltaoffont_{\ref{example poole 6}}}}
is a simplified activation set for \pair{\mathcal A_2}{\ident g_2},
but neither for \pair\emptyset{\ident g_2},
nor for \nlbmaths{\pair{\mathcal A_1}{\ident g_1}}.
\\\citet{Poole-Preferring-Most-Specific-1985} considers the same result for 
\math{\lesssim_{\rm P1}} 
as for
\math{\lesssim_{\rm P3}} 
to be 
``seemingly unintuitive\closequotecomma
because, as we have seen in the isomorphic sub-specification of 
\examref{example poole 3}, \hskip.2em
we have both 
\maths{
\pair{\mathcal A_1}{\neg\ident c}
<_{\rm P3}
\pair{\mathcal A_2}{\ident c}
}{} and
\maths{
\pair{\mathcal A_1}{\neg\ident f}
<_{\rm P3}
\pair{\mathcal A_2}{\ident f}
}. \hskip.3em
Indeed, as already listed as an essential requirement in \nlbsectref
{section Requirements Specification of Specificity in Logic Programming}, 
\hskip.2em the conjunction of two respectively 
more specific derivations should be
more specific, \mbox{shouldn't it?} \
On the other hand, \hskip.1em
considering \math{\lesssim_{\rm CP}} 
instead of \hskip.1em\math{\lesssim_{\rm P3}}, 
the conjunction of two equivalently specific derivations results
in an equivalently specific derivation
--- exactly as one intuitively expects.\end{example}
\begin{example}[\nth 1\,Variation of \examref{example poole 6}]
\label{example variation 1 poole 6}
\\[-5ex]\math{\begin{array}[t]{@{}l l l@{}}
\\[+2.1ex]\Pioffont^{\rm F}_{\ref{example variation 1 poole 6}}
 &:=
 &\Pioffont^{\rm F}_{\ref{example poole 6}},
\\\Pioffont^{\rm G}_{\ref{example variation 1 poole 6}}
 &:=
 &\left\{\begin{array}{l}\ident g_1\antiimplies\neg\ident c\tightund\neg\ident f,
    \\\ident g_2\antiimplies\ident c\tightund\ident f,
    \\\ident b\antiimplies\ident a
    \\\end{array}\right\},
\\\Deltaoffont_{\ref{example variation 1 poole 6}}
 &:=
 &\mathcal A_1\tightcup\mathcal A_2, 
\\\mathcal A_1
 &:=
 &\left\{\begin{array}{l}
      \neg\ident c\defeasibleantiimplies\ident a,
    \\\neg\ident f\defeasibleantiimplies\ident d
    \\\end{array}\right\},
\\\mathcal A_2
 &:=
 &\left\{\begin{array}{l}
      \ident c\defeasibleantiimplies\ident b,
    \\\ident e\defeasibleantiimplies\ident d,
    \\\ident f\defeasibleantiimplies\ident e
    \\\end{array}\right\}.
\\\end{array}}\hfill
\begin{tikzpicture}[baseline=(current bounding box.north),
>=stealth,->,looseness=.5,auto]
\matrix [matrix of math nodes,
column sep={1.4cm,between origins},
row sep={1.3cm,between origins}]{
 &
 &
 & |(geins)| \math{\ident g_1}
\\
 &
 &
 & |(gzwei)| \math{\ident g_2}
\\ |(notc)|  \math{\neg\ident c} 
 &
 & |(c)|     \math{\ident c} 
 &
 & |(f)|     \math{\ident f}
 &
 & |(notf)|  \math{\neg\ident f} 
\\ 
 &
 & |(b)|     \math{\ident b}
 &
 & |(e)|     \math{\ident e}
\\
 & |(a)|     \math{\ident a}
 &
 & |(TRUE)|  \math{\TRUEpp}  
 &
 & |(d)|     \math{\ident d}
\\
};
\begin{scope}[every node/.style={midway,auto}]
\draw [] (b) -- node [right] 
      {$\scriptscriptstyle\!\!\mathcal A_2$} 
      (c);
\draw [] (a) -- node [right] 
      {$\scriptscriptstyle\!\mathcal A_1$} 
      (notc);
\draw [] (e) -- node [right] 
      {$\scriptscriptstyle\!\!\mathcal A_2$} 
      (f);
\draw [] (d) -- node [right] 
      {$\scriptscriptstyle\!\mathcal A_2$} 
      (e);
\draw [] (d) -- node [right] 
      {$\scriptscriptstyle\!\!\mathcal A_1$} 
      (notf);
\draw [double] (a)    -- (b);
\draw [double] (TRUE) -- (a);
\draw [double] (TRUE) -- (d);
\draw [double] (notc) -- (geins);
\draw [double] (notf) -- (geins);
\draw [double] (c)    -- (gzwei);
\draw [double] (f)    -- (gzwei);
\end{scope}
\end{tikzpicture}
\\\noindent Let us compare the specificity of the arguments
\pair{\mathcal A_1}{\ident g_1} and \pair{\mathcal A_2}{\ident g_2}.
\par\noindent\LINEmaths{\mathfrak 
T_{\Pioffont_{\ref{example variation 1 poole 6}}}=\{\ident a,\ident b,\ident d\}
\hfill\mathfrak T_
{\Pioffont_{\ref{example variation 1 poole 6}}
 \cup\Deltaoffont_{\ref{example variation 1 poole 6}}}=
\{\ident c,\ident e,\ident f,\ident g_1,\ident g_2,\neg\ident c,\neg\ident f\}
\cup\mathfrak T_{\Pioffont_{\ref{example variation 1 poole 6}}}}.
\par\noindent We now have
\bigmaths{\pair{\mathcal A_2}{\ident g_2}
\not\lesssim_{\rm CP}
\pair{\mathcal A_1}{\ident g_1}
}{} 
because \math{\{\ident b,\ident d\}\subseteq
\mathfrak T_{\Pioffont_{\ref{example variation 1 poole 6}}}}
is an activation set for \pair{\mathcal A_2}{\ident g_2},
but not for \pair{\mathcal A_1}{\ident g_1}.
\\We still have
\bigmaths{
\pair{\mathcal A_1}{\ident g_1}
\lesssim_{\rm CP}
\pair{\mathcal A_2}{\ident g_2}
}{}
because, \ 
for any activation set
\math{H\subseteq\mathfrak T_{\Pioffont_{\ref{example variation 1 poole 6}}}}
for \pair{\mathcal A_1}{\ident g_1}, \ 
we have
\bigmaths{\{\ident a,\ident b\}\,\tightsubseteq\,H};
and so \math H is also an activation set for \pair{\mathcal A_2}{\ident g_2}.
\\We again have \bigmaths{
\pair{\mathcal A_1}{\ident g_1}
\vartriangle_{\rm P3}
\pair{\mathcal A_2}{\ident g_2}}, 
for the same reason as in \examref{example poole 6}.
Thus, 
the situation for \math{\lesssim_{\rm P3}} is just as in 
\examref{example poole 6}, \hskip.2em
and just as ``seemingly unintuitive'' for exactly the same reason.
\\We have \bigmaths{
\pair{\mathcal A_1}{\ident g_1}<_{\rm CP}\pair{\mathcal A_2}{\ident g_2}}, 
which is intuitive because
the conjunction of a more specific and an equivalently specific element,
respectively,
should be more specific. \hskip.3em
Indeed, \hskip.1em
from the isomorphic sub-specifications
in \examrefs{example poole 2}{example poole 3}, \hskip.1em
we know that \bigmaths{
\pair{\mathcal A_1}{\neg\ident c}
<_{\rm CP}
\pair{\mathcal A_2}{\ident c}}{}
and
\bigmaths{
\pair{\mathcal A_1}{\neg\ident f}
\approx_{\rm CP}
\pair{\mathcal A_2}{\ident f}}, 
respectively. \hskip.3em
\\All in all, 
\poole's specificity relation \nlbmath{\lesssim_{\rm P3}} 
fails in this example again, 
whereas the quasi-ordering
\math{\lesssim_{\rm CP}} \nolinebreak works according to human intuition
and satisfies the required monotonicity \wrt\ conjunction of \sectref
{section Requirements Specification of Specificity in Logic Programming}.
\end{example}
\subsection{Implementation of the Preference of the ``More Precise''}
\noindent
As primary sources of differences in specificity,
the previous examples illustrate only the effect of a chain of
implications.
As in our motivating discussion of 
\nlbsectref{section Toward an Intuitive Notion of Specificity}, \hskip.2em
we should also consider examples where the primary source is an 
essentially required condition that is a super-conjunction of the condition 
triggering another rule. 
We will do so in the following examples.
\begin{example}\label{example lovely one}
\\[-3ex]\math{\begin{array}[t]{@{}l l l@{}}
\\\Pioffont^{\rm F}_{\ref{example lovely one}}
 &:=
 &\left\{\begin{array}{l}\ident{somebody},
    \\\ident{noisy}
    \\\end{array}\right\},
\\\Pioffont^{\rm G}_{\ref{example lovely one}}
 &:=
 &\left\{\begin{array}{l}\ident{lovely}\antiimplies\ident{grandma},
    \\\neg\ident{lovely}\antiimplies\ident{grandpa}
    \\\end{array}\right\},
\\\Deltaoffont_{\ref{example lovely one}}
 &:=
 &\mathcal A_1\cup\mathcal A_2
\\\mathcal A_1
 &:=
 &\left\{\begin{array}{l}
      \ident{grandpa}
      \defeasibleantiimplies\ident{somebody}\tightund\ident{noisy}
    \\\end{array}\right\},
\\\mathcal A_2
 &:=
 &\left\{\begin{array}{l}
      \ident{grandma}\defeasibleantiimplies\ident{somebody}
    \\\end{array}\right\}.
\\\end{array}}\hfill
\begin{tikzpicture}[baseline=(current bounding box.north),
>=stealth,->,looseness=.5,auto]
\matrix [matrix of math nodes,
column sep={1.4cm,between origins},
row sep={1.3cm,between origins}]{
 &
 & |(notlovely)| \math{\neg\ident{lovely}} 
 &
 & |(lovely)|    \math{\ident{lovely}}
\\ 
 &
 & |(grandpa)|   \math{\ident{grandpa}}
 &
 & |(grandma)|   \math{\ident{grandma}}
\\ |(noisy)|     \math{\ident{noisy}}
 &
 & |(TRUE)|      \math{\TRUEpp}  
 &
 & |(somebody)|  \math{\ident{somebody}}
\\
};
\begin{scope}[every node/.style={midway,auto}]
\draw [] (noisy) -- node [right] 
      {$\scriptscriptstyle\,\mathcal A_1$} 
      (grandpa);
\draw [] (somebody) -- node [right] 
      {$\scriptscriptstyle\,\mathcal A_1$} 
      (grandpa);
\draw [] (somebody) -- node [right] 
      {$\scriptscriptstyle\!\!\mathcal A_2$} 
      (grandma);
\draw [double] (grandpa) -- (notlovely);
\draw [double] (grandma) -- (lovely);
\draw [double] (TRUE) -- (noisy);
\draw [double] (TRUE) -- (somebody);
\end{scope}
\end{tikzpicture}
\par\noindent Let us compare the specificity of the arguments
\pair{\mathcal A_1}{\neg\ident{lovely}} and 
\nlbmath{\pair{\mathcal A_2}{\ident{lovely}}}.
\par\noindent\LINEmaths{\mathfrak T_{\Pioffont_{\ref{example lovely one}}}
=\{\ident{somebody},
\ident{noisy}\},\hfill\mathfrak T_
{\Pioffont_{\ref{example lovely one}}\cup\Deltaoffont_{\ref{example lovely one}}}
=\{\ident{grandma},\ident{grandpa},\ident{lovely},\neg\ident{lovely}\}\cup
\mathfrak T_{\Pioffont_{\ref{example lovely one}}}}.
\par\noindent 
We have 
\maths{
\pair{\mathcal A_2}{\ident{lovely}}
\tight{\not\lesssim_{\rm CP}}
\pair{\mathcal A_1}{\neg\ident{lovely}}
}{}
because \maths
{\ident{lovely}\tightnotin\mathfrak T_{\Pioffont_{\ref{example lovely one}}}}{} 
and
because \mbox{\math{\{\ident{somebody}\}\subseteq
\mathfrak T_{\Pioffont_{\ref{example lovely one}}}}} is an activation set for
\pair{\mathcal A_2}{\ident{lovely}}, but not for 
\pair{\mathcal A_1}{\neg\ident{lovely}}.
\\We have \maths{\pair{\mathcal A_1}{\neg\ident{lovely}}
\tight{\lesssim_{\rm P3}}\pair{\mathcal A_2}{\ident{lovely}}}{}
because of \maths
{\ident{lovely}\tightnotin\mathfrak T_{\Pioffont_{\ref{example lovely one}}}}{} 
and because, 
if \math{H\subseteq\mathfrak T_{\Pioffont_{\ref{example lovely one}}
\cup\Deltaoffont_{\ref{example lovely one}}}}
is a simplified activation set for \pair{\mathcal A_1}{\neg\ident{lovely}}, \ 
but not for \pair\emptyset{\neg\ident{lovely}}, \
then we \nolinebreak have \bigmaths{
\{\ident{somebody},\ident{noisy}\}
\subseteq H},
and so \math H is also a simplified activation set for 
\pair{\mathcal A_2}{\ident{lovely}}.
\\All in all, by \theoref{theorem super quasi-ordering},
\math{\begin{array}[t]{@{}r r@{}c@{}l}\mbox{we get}
 &\pair{\mathcal A_1}{\neg\ident{lovely}}
 &<_{\rm CP}
 &\pair{\mathcal A_2}{\ident{lovely}}
\\\mbox{and}
 &\pair{\mathcal A_1}{\neg\ident{lovely}}
 &<_{\rm P3}
 &\pair{\mathcal A_2}{\ident{lovely}}.
\\\end{array}}
\\Note that we can nicely see here that the condition 
that \math H is not a simplified
activation set for \pair\emptyset{\neg\ident{lovely}}
is relevant in \defiref{DefSpec}. 
Without this condition we would have to consider
the simplified activation set \math{\{\ident{grandpa}\}} for 
\pair{\mathcal A_1}{\neg\ident{lovely}}, which is not an activation set for
\pair{\mathcal A_2}{\ident{lovely}}; \
and so, 
contrary to our intuition,
\pair{\mathcal A_1}{\neg\ident{lovely}}
would not be more specific than 
\pair{\mathcal A_2}{\ident{lovely}} 
\wrt\ \nlbmath{\lesssim_{\rm P3}} any more.%
\end{example}\par\halftop\noindent
So both relations \math{\lesssim_{\rm P3}} and \nlbmath{\lesssim_{\rm CP}} 
produce the intuitive result if the 
``more precise''
super-conjunction is directly the 
condition of a rule.
Let us see whether this is also the case if the condition of the rule
is only derived from a super-conjunction.
\begin{example}[\nth 2\,Variation of \examref{example poole 6}]
\label{variation 2 example poole 6}
\\[-4ex]\math{\begin{array}[t]{@{}l l l@{}}
\\[+1ex]\Pioffont^{\rm F}_{\ref{variation 2 example poole 6}}
 &:=
 &\Pioffont^{\rm F}_{\ref{example poole 6}},
\\\Pioffont^{\rm G}_{\ref{variation 2 example poole 6}}
 &:=
 &\left\{\begin{array}{l}\ident g_1\antiimplies\neg\ident c,
    \\\ident g_2\antiimplies\ident c\tightund\ident f
    \\\end{array}\right\},
\\\Deltaoffont_{\ref{variation 2 example poole 6}}
 &:=
 &\mathcal A_1\tightcup\mathcal A_2, 
\\\mathcal A_1
 &:=
 &\left\{\begin{array}{l}
      \neg\ident c\defeasibleantiimplies\ident a
    \\\end{array}\right\},
\\\mathcal A_2
 &:=
 &\left\{\begin{array}{l}
      \ident b\defeasibleantiimplies\ident a,
    \\\ident c\defeasibleantiimplies\ident b,
    \\\ident e\defeasibleantiimplies\ident d,
    \\\ident f\defeasibleantiimplies\ident e
    \\\end{array}\right\}
\\\end{array}}\hfill
\begin{tikzpicture}[baseline=(current bounding box.north),
>=stealth,->,looseness=.5,auto]
\matrix [matrix of math nodes,
column sep={1.4cm,between origins},
row sep={1.3cm,between origins}]{
 & |(geins)| \math{\ident g_1}
 &
 & |(gzwei)| \math{\ident g_2}
\\ |(notc)|  \math{\neg\ident c} 
 &
 & |(c)|     \math{\ident c} 
 &
 & |(f)|     \math{\ident f}
\\ 
 &
 & |(b)|     \math{\ident b}
 &
 & |(e)|     \math{\ident e}
\\
 & |(a)|     \math{\ident a}
 &
 & |(TRUE)|  \math{\TRUEpp}  
 &
 & |(d)|     \math{\ident d}
\\
};
\begin{scope}[every node/.style={midway,auto}]
\draw [] (b) -- node [right] 
      {$\scriptscriptstyle\!\!\mathcal A_2$} 
      (c);
\draw [] (a) -- node [right] 
      {$\scriptscriptstyle\!\!\mathcal A_2$} 
      (b);
\draw [] (a) -- node [right] 
      {$\scriptscriptstyle\!\mathcal A_1$} 
      (notc);
\draw [] (e) -- node [right] 
      {$\scriptscriptstyle\!\!\mathcal A_2$} 
      (f);
\draw [] (d) -- node [right] 
      {$\scriptscriptstyle\!\mathcal A_2$} 
      (e);
\draw [double] (TRUE) -- (a);
\draw [double] (TRUE) -- (d);
\draw [double] (notc) -- (geins);
\draw [double] (c)    -- (gzwei);
\draw [double] (f)    -- (gzwei);
\end{scope}
\end{tikzpicture}
\\\noindent Let us compare the specificity of the arguments
\pair{\mathcal A_1}{\ident g_1} and \pair{\mathcal A_2}{\ident g_2}.
\par\noindent\LINEmaths{\mathfrak T_{\Pioffont_{\ref{variation 2 example poole 6}}}=\{\ident a, \ident d\}
\hfill\mathfrak 
T_{\Pioffont_{\ref{variation 2 example poole 6}}\cup\Deltaoffont_{\ref{variation 2 example poole 6}}}=
\{\ident b,\ident c,\ident e,\ident f,
  \ident g_1,\ident g_2,\neg\ident c\}
\cup\mathfrak T_{\Pioffont_{\ref{variation 2 example poole 6}}}}.
\par\noindent We have 
\bigmaths{\pair{\mathcal A_1}{\ident g_1}
\not\lesssim_{\rm CP}
\pair{\mathcal A_2}{\ident g_2}}{}
because 
\math{\{\ident a\}\subseteq
\mathfrak T_{\Pioffont_{\ref{variation 2 example poole 6}}}}
is an activation set for \pair{\mathcal A_1}{\ident g_1}, \hskip.2em
but not for \pair{\mathcal A_2}{\ident g_2}. \
We have \bigmaths{\pair{\mathcal A_2}{\ident g_2}
\lesssim_{\rm CP}\pair{\mathcal A_1}{\ident g_1}}{}
because any activation set for \pair{\mathcal A_2}{\ident g_2}
that is a subset of 
\math{\mathfrak T_{\Pioffont_{\ref{variation 2 example poole 6}}}}
includes \nlbmath{\ident a}, and so is also an activation set for 
\pair{\mathcal A_1}{\ident g_1}. \ 
Considering 
\theoref{theorem super quasi-ordering} and the
the activation set \nlbmath{\{\ident b,\ident d\}}
for \pair{\mathcal A_2}{\ident g_2},
we see \bigmaths{\pair{\mathcal A_1}{\ident g_1}
\vartriangle_{\rm P3}\pair{\mathcal A_2}{\ident g_2}}.
\\All in all, \hskip.2em
only \nlbmath{\lesssim_{\rm CP}} 
realizes 
\ ---~~via~~\math{\pair{\mathcal A_2}{\ident g_2}<_{\rm CP}
\pair{\mathcal A_1}{\ident g_1}}~~--- \ 
the intuition that the super-conjunction \nlbmath{\ident a\tightund\ident d}
---~which is essential to derive \nlbmath{\ident c\tightund\ident f} 
according to \nlbmath{\mathcal A_2}~---
is more specific than the ``less precise'' \nlbmath{\ident a}.
\\Just like \examref{example poole 3}, \hskip.2em
this example shows again that \math{\lesssim_{\rm P3}} 
does not really implement the intuition that 
defeasible rules should be considered for their global semantical
effect instead of their syntactical fine structure.%
\end{example}
\halftop
\begin{example}[\litexamref{11} from 
{\cite[\p\,96]{Stolzenburg-etal-Computing-Specificity-2003}}]
\label{example stolzenburg 2}
\\[-3ex]\math{\begin{array}[t]{@{}l l@{}}\multicolumn{2}{@{}l@{}}{%
\begin{array}[t]{@{}l l l@{}}
    \\\Pioffont^{\rm F}_{\ref{example stolzenburg 2}}
     &:=
     &\left\{\begin{array}{l}\ident c,
        \\\ident d,
        \\\ident e
        \\\end{array}\right\},
    \\\Pioffont^{\rm G}_{\ref{example stolzenburg 2}}
     &:=
     &\left\{\begin{array}{l}\ident x\antiimplies\ident a\tightund\ident f
        \\\end{array}\right\},
    \\\Deltaoffont_{\ref{example stolzenburg 2}}
     &:=
     &\mathcal A^{1}\cup
      \mathcal A^{2}\cup
      \mathcal A^{3}\cup
      \mathcal A^{4}\cup
      \mathcal A^{5},
    \\\end{array}}
\\\begin{array}[b]{@{}l l l@{}}\mathcal A^{1}
     &:=
     &\left\{\begin{array}{l}\ident x
                \defeasibleantiimplies\ident a\tightund\ident b\tightund\ident c
        \\\end{array}\right\},
    \\\mathcal A^{2}
     &:=
     &\left\{\begin{array}{l}\neg\ident x
                             \defeasibleantiimplies\ident a\tightund\ident b
        \\\end{array}\right\},
    \\\mathcal A^{3}
     &:=
     &\left\{\begin{array}{l}
          \ident f\defeasibleantiimplies\ident e
        \\\end{array}\right\},
    \\\end{array}
 &\begin{array}[b]{@{}l l l@{}}\mathcal A^{4}
     &:=
     &\left\{\begin{array}{l}\ident a\defeasibleantiimplies\ident d
        \\\end{array}\right\},
    \\\mathcal A^{5}
     &:=
     &\left\{\begin{array}{l}\ident b\defeasibleantiimplies\ident e
        \\\end{array}\right\}.
    \\\end{array}
\\\end{array}}\hfill
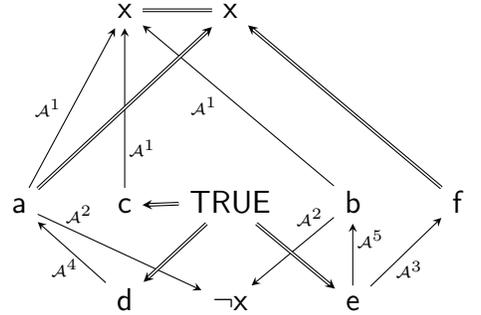
\begin{tikzpicture}[baseline=(current bounding box.north),
>=stealth,->,looseness=.5,auto]
\matrix [matrix of math nodes,
column sep={1.4cm,between origins},
row sep={1.3cm,between origins}]{
 & |(xeins)|  \math{\ident x}
 & |(xzwei)|  \math{\ident x}
\\ 
\\ |(a)|      \math{\ident a} 
 & |(c)|      \math{\ident c}  
 & |(TRUE)|   \math{\TRUEpp}  
 & |(b)|      \math{\ident b} 
 & |(f)|      \math{\ident f}
\\
 & |(d)|      \math{\ident d} 
 & |(notx)|   \math{\neg\ident x} 
 & |(e)|      \math{\ident e} 
\\ 
};
\begin{scope}[every node/.style={midway,auto}]
\draw [] (a) -- node [left] 
      {$\scriptscriptstyle\!\!\mathcal A^1$} 
      (xeins);
\draw [] (c) -- node [near start, right] 
      {$\scriptscriptstyle\!\!\mathcal A^1$} 
      (xeins);
\draw [] (b) -- node [left] 
      {$\scriptscriptstyle\mathcal A^1$\,\,} 
      (xeins);
\draw [] (a) -- node [at start, right] 
      {$\scriptscriptstyle\,\,\,\,\,\mathcal A^2$} 
      (notx);
\draw [] (b) -- node [at start, left] 
      {$\scriptscriptstyle\mathcal A^2$} 
      (notx);
\draw [] (e) -- node [near start, right] 
      {$\scriptscriptstyle\!\mathcal A^3$} 
      (f);
\draw [] (d) -- node [near start, left] 
      {$\scriptscriptstyle\mathcal A^4$} 
      (a);
\draw [] (e) -- node [near end, right] 
      {$\scriptscriptstyle\!\!\mathcal A^5$} 
      (b);
\draw [double] (a)    -- (xzwei);
\draw [double] (f)    -- (xzwei);
\draw [double] (TRUE) -- (c);
\draw [double] (TRUE) -- (d);
\draw [double] (TRUE) -- (e);
\draw [double, -] (xeins) -- (xzwei);
\end{scope}
\end{tikzpicture}
\\\noindent Let us compare the specificity of the arguments 
\pair{\mathcal A^1\tightcup\mathcal A^4\tightcup\mathcal A^5}{\ident x},
\pair{\mathcal A^2\tightcup\mathcal A^4\tightcup\mathcal A^5}{\neg\ident x}, 
\pair{\mathcal A^3\tightcup\mathcal A^4}{\ident x}.
\par\noindent\LINEmaths{
  \mathfrak T_{\Pioffont_{\ref{example stolzenburg 2}}}
 =
 \{\ident c,\ident d,\ident e\},
 \hfill
 \mathfrak T_{\Pioffont_{\ref{example stolzenburg 2}}
        \cup\Deltaoffont_{\ref{example stolzenburg 2}}}
 =
 \{\ident a,\ident b,\ident f,\ident x,\neg\ident x\}\cup
  \mathfrak T_{\Pioffont_{\ref{example stolzenburg 2}}}}.
\par\noindent 
We have \bigmaths{
\pair{\mathcal A^1\tightcup\mathcal A^4\tightcup\mathcal A^5}{\ident x}
<_{\rm CP}
\pair{\mathcal A^2\tightcup\mathcal A^4\tightcup\mathcal A^5}{\neg\ident x}
\approx_{\rm CP}
\pair{\mathcal A^3\tightcup\mathcal A^4}{\ident x}
},
because of \bigmaths
{\ident x,\neg\ident x\tightnotin
 \mathfrak T_{\Pioffont_{\ref{example stolzenburg 2}}}}, and
because any activation set \math{H\subseteq
\mathfrak T_{\Pioffont_{\ref{example stolzenburg 2}}}} for any of 
\pair{\mathcal A^1\tightcup\mathcal A^4\tightcup\mathcal A^5}{\ident x},
\pair{\mathcal A^2\tightcup\mathcal A^4\tightcup\mathcal A^5}{\neg\ident x},
\pair{\mathcal A^3\tightcup\mathcal A^4}{\ident x}
contains \nlbmath{\{\ident d,\ident e\}},
which is an activation set only for the latter two.
This matches our intuition well,
because the first of these arguments essentially requires
the ``more precise'' \math{\ident c\tightund\ident d\tightund\ident e} \hskip.2em
instead of the less specific 
\nlbmath{\ident d\tightund\ident e}.
\\We have
\bigmaths{
\pair{\mathcal A^1\tightcup\mathcal A^4\tightcup\mathcal A^5}{\ident x}
\vartriangle_{\rm P3}
\pair{\mathcal A^2\tightcup\mathcal A^4\tightcup\mathcal A^5}{\neg\ident x}
\vartriangle_{\rm P3}
\pair{\mathcal A^3\tightcup\mathcal A^4}{\ident x}
\vartriangle_{\rm P3}
\pair{\mathcal A^1\tightcup\mathcal A^4\tightcup\mathcal A^5}{\ident x}
}, however.\footnotemark\ 
This means that
\math{\lesssim_{\rm P3}} cannot compare these counterarguments
and cannot help us to pick the more specific argument.
\\What is most interesting under the computational aspect is that,
for realizing 
\par\noindent\LINEmaths{
\pair{\mathcal A^1\tightcup\mathcal A^4\tightcup\mathcal A^5}{\ident x}
\not\lesssim_{\rm P3}
\pair{\mathcal A^2\tightcup\mathcal A^4\tightcup\mathcal A^5}{\neg\ident x}},
\par\noindent we have to consider 
(implicitly via \math{\{\ident d,\ident f\}\,\tightsubseteq\,\mathfrak T_
{\Pioffont_{\ref{example stolzenburg 2}}\cup
 \Deltaoffont_{\ref{example stolzenburg 2}}}})
the defeasible rule of \nlbmaths{\mathcal A^3}, \hskip.2em
which is not part of any of the two arguments under comparison. \hskip.3em
Note that such considerations are not required, 
however, 
for realizing the properties of \nlbmaths{\lesssim_{\rm CP}},
because defeasible rules not in the given argument can be 
completely ignored when calculating the minimal activation sets
as subsets of \nlbmath{\mathfrak T_\Pioffont} instead of
\nlbmaths{\mathfrak T_{\Pioffont\cup\Deltaoffont}}. \hskip.3em
This means in particular that the complication of {\em pruning}
\ ---~~as discussed in detail in 
\cite[\litsectref{3.3}]{Stolzenburg-etal-Computing-Specificity-2003}~~--- \
does not have to be considered for the operationalization 
of \nolinebreak\hskip.2em\nlbmath{\lesssim_{\rm CP}}.
\end{example}
\footnotetext{%
 Because \math{\{\ident d,\ident f\}\subseteq\mathfrak T_
 {\Pioffont_{\ref{example stolzenburg 2}}\cup
  \Deltaoffont_{\ref{example stolzenburg 2}}}}
 is a simplified activation set for \pair{\mathcal A^4}{\ident x}, \hskip.2em
 but neither for \pair\emptyset{\ident x}, \hskip.2em
 nor for \pair{\mathcal A^2\tightcup\mathcal A^4\tightcup\mathcal A^5}{\neg\ident x},
 \hskip.2em 
 we have 
 \bigmaths
 {\pair{\mathcal A^1\tightcup\mathcal A^4\tightcup\mathcal A^5}{\ident x}
 \not\lesssim_{\rm P3}
 \pair{\mathcal A^2\tightcup\mathcal A^4\tightcup\mathcal A^5}{\neg\ident x}
 \not\gtrsim_{\rm P3}\pair{\mathcal A^3\tightcup\mathcal A^4}{\ident x}}.
 \\Because of 
 \\[-2.7ex]\linemaths{\pair{\mathcal A^3\tightcup\mathcal A^4}{\ident x}
 \not\lesssim_{\rm CP}
 \pair{\mathcal A^1\tightcup\mathcal A^4\tightcup\mathcal A^5}{\ident x}
 \not\gtrsim_{\rm CP}
 \pair{\mathcal A^2\tightcup\mathcal A^4\tightcup\mathcal A^5}{\neg\ident x}},
 we have
 \\[-2.7ex]\linemaths{\pair{\mathcal A^3\tightcup\mathcal A^4}{\ident x}
 \not\lesssim_{\rm P3}
 \pair{\mathcal A^1\tightcup\mathcal A^4\tightcup\mathcal A^5}{\ident x}
 \not\gtrsim_{\rm P3}
 \pair{\mathcal A^2\tightcup\mathcal A^4\tightcup\mathcal A^5}{\neg\ident x}}{}
 by \theoref{theorem super quasi-ordering}. \ 
 Because \math{\{\ident b,\ident c,\ident d\}\subseteq\mathfrak T_
 {\Pioffont_{\ref{example stolzenburg 2}}\cup
  \Deltaoffont_{\ref{example stolzenburg 2}}}}
 is a simplified activation set for 
 \pair{\mathcal A^2\tightcup\mathcal A^4\tightcup\mathcal A^5}{\neg\ident x}
 and \pair{\mathcal A^1\tightcup\mathcal A^4\tightcup\mathcal A^5}{\ident x},
 but for none of \pair\emptyset{\neg\ident x}, \pair\emptyset{\ident x},
 \pair{\mathcal A^3\tightcup\mathcal A^4}{\ident x}, \hskip.2em
 we have \\\LINEmaths{
 \pair{\mathcal A^2\tightcup\mathcal A^4\tightcup\mathcal A^5}{\neg\ident x}
 \not\lesssim_{\rm P3}
 \pair{\mathcal A^3\tightcup\mathcal A^4}{\ident x}
 \not\gtrsim_{\rm P3}
 \pair{\mathcal A^1\tightcup\mathcal A^4\tightcup\mathcal A^5}{\ident x} 
 }.%
}%
\halftop\begin{example}[Variation of \examref{example stolzenburg 2}]
\label{example stolzenburg 3}\sloppy
\\[-2.5ex]\math{\begin{array}[t]{@{}l@{}l@{}}\begin{array}[t]{@{}l l l@{}}
    \\[-.5ex]\Pioffont^{\rm F}_{\ref{example stolzenburg 3}}
     &:=
     &\left\{\begin{array}{l}\ident c,
        \\\ident d,
        \\\ident e
        \\\end{array}\right\},
    \\\Pioffont^{\rm G}_{\ref{example stolzenburg 3}}
     &:=
     &\left\{\begin{array}{l}\ident x\antiimplies\ident a\tightund\ident f,
        \\\ident f\antiimplies\ident e
        \\\end{array}\right\},
    \\\Deltaoffont_{\ref{example stolzenburg 3}}
     &:=
     &\mathcal A^1\cup
      \mathcal A^2\cup
      \mathcal A^4\cup
      \mathcal A^5,
    \\\end{array}
\\\begin{array}[b]{@{}l l l@{}}\mathcal A^1
     &:=
     &\left\{\begin{array}{l}\ident x
                \defeasibleantiimplies\ident a\tightund\ident b\tightund\ident c
        \\\end{array}\right\},
    \\\mathcal A^2
     &:=
     &\left\{\begin{array}{l}\neg\ident x
                             \defeasibleantiimplies\ident a\tightund\ident b
        \\\end{array}\right\},
    \\\mathcal A^4
     &:=
     &\left\{\begin{array}{l}\ident a\defeasibleantiimplies\ident d
        \\\end{array}\right\},
    \\\mathcal A^5
     &:=
     &\left\{\begin{array}{l}\ident b\defeasibleantiimplies\ident e
        \\\end{array}\right\}.
    \\\end{array}
\\\end{array}}\hfill
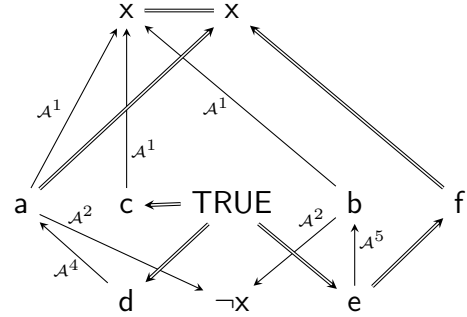
\begin{tikzpicture}[baseline=(current bounding box.north),
>=stealth,->,looseness=.5,auto]
\matrix [matrix of math nodes,
column sep={1.4cm,between origins},
row sep={1.3cm,between origins}]{
 & |(xeins)|  \math{\ident x}
 & |(xzwei)|  \math{\ident x}
\\ 
\\ |(a)|      \math{\ident a} 
 & |(c)|      \math{\ident c}  
 & |(TRUE)|   \math{\TRUEpp}  
 & |(b)|      \math{\ident b} 
 & |(f)|      \math{\ident f}
\\
 & |(d)|      \math{\ident d} 
 & |(notx)|   \math{\neg\ident x} 
 & |(e)|      \math{\ident e} 
\\ 
};
\begin{scope}[every node/.style={midway,auto}]
\draw [] (a) -- node [left] 
      {$\scriptscriptstyle\!\!\mathcal A^1$} 
      (xeins);
\draw [] (c) -- node [near start, right] 
      {$\scriptscriptstyle\!\!\mathcal A^1$} 
      (xeins);
\draw [] (b) -- node [left] 
      {$\scriptscriptstyle\mathcal A^1$} 
      (xeins);
\draw [] (a) -- node [at start, right] 
      {$\scriptscriptstyle\,\,\,\,\,\mathcal A^2$} 
      (notx);
\draw [] (b) -- node [at start, left] 
      {$\scriptscriptstyle\mathcal A^2$} 
      (notx);
\draw [] (d) -- node [near start, left] 
      {$\scriptscriptstyle\mathcal A^4$} 
      (a);
\draw [] (e) -- node [near end, right] 
      {$\scriptscriptstyle\!\!\mathcal A^5$} 
      (b);
\draw [double] (e)    -- (f);
\draw [double] (a)    -- (xzwei);
\draw [double] (f)    -- (xzwei);
\draw [double] (TRUE) -- (c);
\draw [double] (TRUE) -- (d);
\draw [double] (TRUE) -- (e);
\draw [double, -] (xeins) -- (xzwei);
\end{scope}
\end{tikzpicture}
\par\noindent Let us compare the specificity of the arguments
\pair{\mathcal A^1\tightcup\mathcal A^4\tightcup\mathcal A^5}{\ident x},
\pair{\mathcal A^2\tightcup\mathcal A^4\tightcup\mathcal A^5}{\neg\ident x},
\nlbmaths{\pair{\mathcal A^4}{\ident x}}.
\par\noindent\LINEmaths{
  \mathfrak T_{\Pioffont_{\ref{example stolzenburg 3}}}
 =
 \{\ident c,\ident d,\ident e,\ident f\},
 \hfill
 \mathfrak T_{\Pioffont_{\ref{example stolzenburg 3}}
        \cup\Deltaoffont_{\ref{example stolzenburg 3}}}
 =
 \{\ident a,\ident b,\ident x,\neg\ident x\}\cup
  \mathfrak T_{\Pioffont_{\ref{example stolzenburg 3}}}}{.~~~~~~~~~~\mbox{}}
\par\noindent 
 Obviously, \bigmaths
 {\ident x,\neg\ident x\tightnotin
  \mathfrak T_{\Pioffont_{\ref{example stolzenburg 3}}}}.
 Moreover, \math{\{\ident d\}\subseteq
 \mathfrak T_{\Pioffont_{\ref{example stolzenburg 3}}}} is an activation set 
 for 
 \pair{\mathcal A^4}{\ident x} 
 (but not a simplified one!) 
 and, 
 {\it\afortiori}\/ \hskip.1em
 (by \cororef{corollary sub-argument CP}), \hskip.2em
 for 
 \pair{\mathcal A^1\tightcup\mathcal A^4\tightcup\mathcal A^5}{\ident x},
 but not for 
 \pair{\mathcal A^2\tightcup\mathcal A^4\tightcup\mathcal A^5}{\neg\ident x}. \
 Furthermore, 
 every activation set 
 \math{H\subseteq\mathfrak T_{\Pioffont_{\ref{example stolzenburg 3}}}}
 for \pair{\mathcal A^2\tightcup\mathcal A^4\tightcup\mathcal A^5}{\neg\ident x}
 satisfies \bigmaths{\{\ident d,\ident e\}\,\tightsubseteq\,H},
 which is an activation set for \pair{\mathcal A^4}{\ident x} and
 \pair{\mathcal A^1\tightcup\mathcal A^4\tightcup\mathcal A^5}{\ident x}.
 \hskip.3em
 Furthermore, 
 every activation set 
 \math{H\,\tightsubseteq\,\mathfrak T_{\Pioffont_{\ref{example stolzenburg 3}}}}
 for \pair{\mathcal A^1\tightcup\mathcal A^4\tightcup\mathcal A^5}{\ident x}
 satisfies \bigmaths{\{\ident d\}\subseteq H}{}
 which is an activation set for \nlbmaths{\pair{\mathcal A^4}{\ident x}}.
\par\noindent All in all, 
we have \bigmaths{\pair{\mathcal A^4}{\ident x}\approx_{\rm CP}
\pair{\mathcal A^1\tightcup\mathcal A^4\tightcup\mathcal A^5}{\ident x}>_{\rm CP}
\pair{\mathcal A^2\tightcup\mathcal A^4\tightcup\mathcal A^5}{\neg\ident x}}. 
\par\noindent This is intuitively sound because 
\pair{\mathcal A^2\tightcup\mathcal A^4\tightcup\mathcal A^5}{\neg\ident x}
is activated only by the more specific \maths{\ident d\tightund\ident e}, 
whereas \pair{\mathcal A^4}{\ident x} 
is activated also by the 
``less precise'' \nlbmaths{\ident d}. \
Moreover, \math{\ident c\tightund\ident d\tightund\ident e} 
is not essentially required
for \pair{\mathcal A^1\tightcup\mathcal A^4\tightcup\mathcal A^5}{\ident x},
and so this argument is equivalent to \pair{\mathcal A^4}{\ident x}.
\\We have
\\[-1.7ex]\noindent\mbox{~~~~~~~~~~~~~~~~}\LINEmaths{
\pair{\mathcal A^4}{\ident x}<_{\rm P3}
\pair{\mathcal A^1\tightcup\mathcal A^4\tightcup\mathcal A^5}{\ident x}
\vartriangle_{\rm P3}
\pair{\mathcal A^2\tightcup\mathcal A^4\tightcup\mathcal A^5}{\neg\ident x}
\vartriangle_{\rm P3}
\pair{\mathcal A^4}{\ident x}},
\par\noindent however.\footnote{%
 The minimal simplified activation sets for \pair{\mathcal A^4}{\ident x} 
 that are no simplified activation sets for \pair\emptyset{\ident x} 
 are \math{\{\ident d,\ident e\}} and \math{\{\ident d,\ident f\}}. \
 The minimal simplified activation sets for 
 \pair{\mathcal A^1\tightcup\mathcal A^4\tightcup\mathcal A^5}{\ident x} 
 that are no simplified activation sets for \pair\emptyset{\ident x} 
 are \math{\{\ident d,\ident e\}},
 \math{\{\ident d,\ident f\}},
 \math{\{\ident a,\ident b,\ident c\}},
 and \math{\{\ident b,\ident c,\ident d\}}. \
 The minimal simplified activation sets for 
 \pair{\mathcal A^2\tightcup\mathcal A^4\tightcup\mathcal A^5}{\neg\ident x}
 that are no simplified activation sets for \pair\emptyset{\neg\ident x} 
 are 
 \maths{\{\ident a,\ident b\}},
 \maths{\{\ident a,\ident e\}},
 \maths{\{\ident b,\ident d\}}, and
 \maths{\{\ident d,\ident e\}}.%
} \hskip.2em
This means that \math{\lesssim_{\rm P3}} 
fails here completely \wrt\ \poole's intuition, 
as actually in most non-trivial examples. \hskip.2em
\end{example}
\vfill\pagebreak
\yestop
\subsection{Conflict between the ``More Concise'' and the ``More Precise''}
\yestop\begin{example}[Variation of \examref{example variation 1 poole 6}]
\label{example variation 3 poole 6}
\\[-3ex]\math{\begin{array}[t]{@{}l l l@{}}
\\\Pioffont^{\rm F}_{\ref{example variation 3 poole 6}}
 &:=
 &\Pioffont^{\rm F}_{\ref{example poole 6}},
\\\Pioffont^{\rm G}_{\ref{example variation 3 poole 6}}
 &:=
 &\left\{\begin{array}{l}\ident g_1\antiimplies\neg\ident c,
    \\\ident g_2\antiimplies\ident c\tightund\ident f,
    \\\ident b\antiimplies\ident a
    \\\end{array}\right\},
\\\Deltaoffont_{\ref{example variation 3 poole 6}}
 &:=
 &\mathcal A_1\tightcup\mathcal A_2, 
\\\mathcal A_1
 &:=
 &\left\{\begin{array}{l}\neg\ident c\defeasibleantiimplies\ident a
    \\\end{array}\right\},
\\\mathcal A_2
 &:=
 &\left\{\begin{array}{l}\ident c\defeasibleantiimplies\ident b,
    \\\ident e\defeasibleantiimplies\ident d,
    \\\ident f\defeasibleantiimplies\ident e
    \\\end{array}\right\}
\\\end{array}}\hfill
\begin{tikzpicture}[baseline=(current bounding box.north),
>=stealth,->,looseness=.5,auto]
\matrix [matrix of math nodes,
column sep={1.4cm,between origins},
row sep={1.3cm,between origins}]{
 & |(geins)| \math{\ident g_1}
 &
 & |(gzwei)| \math{\ident g_2}
\\ |(notc)|  \math{\neg\ident c} 
 &
 & |(c)|     \math{\ident c} 
 &
 & |(f)|     \math{\ident f}
\\ 
 &
 & |(b)|     \math{\ident b}
 &
 & |(e)|     \math{\ident e}
\\
 & |(a)|     \math{\ident a}
 &
 & |(TRUE)|  \math{\TRUEpp}  
 &
 & |(d)|     \math{\ident d}
\\
};
\begin{scope}[every node/.style={midway,auto}]
\draw [] (b) -- node [right] 
      {$\scriptscriptstyle\!\!\mathcal A_2$} 
      (c);
\draw [] (a) -- node [right] 
      {$\scriptscriptstyle\!\mathcal A_1$} 
      (notc);
\draw [] (e) -- node [right] 
      {$\scriptscriptstyle\!\!\mathcal A_2$} 
      (f);
\draw [] (d) -- node [right] 
      {$\scriptscriptstyle\!\mathcal A_2$} 
      (e);
\draw [double] (a)    -- (b);
\draw [double] (TRUE) -- (a);
\draw [double] (TRUE) -- (d);
\draw [double] (notc) -- (geins);
\draw [double] (c)    -- (gzwei);
\draw [double] (f)    -- (gzwei);
\end{scope}
\end{tikzpicture}
\par\noindent Let us compare the specificity of the arguments
\pair{\mathcal A_1}{\ident g_1} and \pair{\mathcal A_2}{\ident g_2}.
\par\noindent\LINEmaths{\mathfrak 
T_{\Pioffont_{\ref{example variation 3 poole 6}}}
=\{\ident a,\ident b,\ident d\},
\hfill\mathfrak T_
{\Pioffont_{\ref{example variation 3 poole 6}}
 \cup\Deltaoffont_{\ref{example variation 3 poole 6}}}=
\{\ident c,\ident e,\ident f,\ident g_1,\ident g_2,\neg\ident c\}\cup
\mathfrak T_{\Pioffont_{\ref{example variation 3 poole 6}}}}.
\par\noindent We now have
\bigmaths{\pair{\mathcal A_1}{\ident g_1}
\vartriangle_{\rm CP}\pair{\mathcal A_2}{\ident g_2}
}{} for the following reasons: \ 
\math{\{\ident{a}\}\subseteq
\mathfrak T_{\Pioffont_{\ref{example variation 3 poole 6}}}}
is an activation set for for \pair{\mathcal A_1}{\ident g_1}, 
but not for \pair{\mathcal A_1}{\ident g_1}; \
\math{\{\ident b,\ident d\}\subseteq
\mathfrak T_{\Pioffont_{\ref{example variation 3 poole 6}}}}
is an activation set for \pair{\mathcal A_2}{\ident g_2},
but not for \pair{\mathcal A_2}{\ident g_2}. \
By \theoref{theorem super quasi-ordering} we also get 
\bigmaths{
\pair{\mathcal A_1}{\ident g_1}
\vartriangle_{\rm P3}
\pair{\mathcal A_2}{\ident g_2}}.
\\
In this example the two intuitive reasons for specificity
\ ---~~super-conjunction (preference of the ``more precise'') \hskip.1em
and implication via a strict rule 
(preference of the ``more concise'')~~--- \
are in an
irresolvable conflict,
which goes well together with the fact that neither 
\math{\lesssim_{\rm CP}} nor \nlbmath{\lesssim_{\rm P3}}  
can compare the two arguments.%
\end{example}\vfill\pagebreak
\subsection{Why Global Effect matters more than Fine Structure}
\halftop\noindent
The following example nicely shows 
that a notion of specificity based only on single defeasible rules
(without considering the context of the strict rules as a whole)
cannot be intuitively adequate.\halftop\noindent
\begin{example}[Example from \litspageref{95} of 
\cite{Stolzenburg-etal-Computing-Specificity-2003}]
\label{example stolzenburg 1}
\\[-3ex]\math{\begin{array}[t]{@{}l l l@{}}
\\\Pioffont^{\rm F}_{\ref{example stolzenburg 1}}
 &:=
 &\left\{\begin{array}{l}\ident q(\ident a)
    \\\end{array}\right\},
\\\Pioffont^{\rm G}_{\ref{example stolzenburg 1}}
 &:=
 &\left\{\begin{array}{l}\ident s(x)\antiimplies\ident q(x)
    \\\end{array}\right\},
\\\Deltaoffont_{\ref{example stolzenburg 1}}
 &:=
 &\left\{\begin{array}{l}\ident p(x)\defeasibleantiimplies\ident q(x),
    \\\neg\ident p(x)\defeasibleantiimplies\ident q(x)\tightund\ident s(x)
    \\\end{array}\right\},
\\\mathcal A_1
 &:=
 &\left\{\begin{array}{l}
      \neg\ident p(\ident a)
      \defeasibleantiimplies\ident q(\ident a)\tightund\ident s(\ident a)
    \\\end{array}\right\},
\\\mathcal A_2
 &:=
 &\left\{\begin{array}{l}
      \ident p(\ident a)\defeasibleantiimplies\ident q(\ident a)
    \\\end{array}\right\}
\\\end{array}}\hfill
\begin{tikzpicture}[baseline=(current bounding box.north),
>=stealth,->,looseness=.5,auto]
\matrix [matrix of math nodes,
column sep={1.4cm,between origins},
row sep={1.3cm,between origins}]{
   |(pa)|     \math{\ident p(\ident a)} 
 &
 & |(notpa)|  \math{\neg\ident p(\ident a)} 
\\ 
 &
 & 
 & |(sa)|     \math{\ident s(\ident a)}
\\ |(TRUE)|   \math{\TRUEpp}  
 & 
 & |(qa)|     \math{\ident q(\ident a)}
\\
};
\begin{scope}[every node/.style={midway,auto}]
\draw [] (sa) -- node [right] 
      {$\scriptscriptstyle\mathcal A_1$} 
      (notpa);
\draw [] (qa) -- node [right] 
      {$\scriptscriptstyle\!\!\mathcal A_1$} 
      (notpa);
\draw [] (qa) -- node [right] 
      {$\scriptscriptstyle\!\mathcal A_2$} 
      (pa);
\draw [double] (qa)    -- (sa);
\draw [double] (TRUE) -- (qa);
\end{scope}
\end{tikzpicture}
\par\noindent Let us compare the specificity of the arguments 
\pair{\mathcal A_1}{\neg\ident p(\ident a)}
and 
\pair{\mathcal A_2}{\ident p(\ident a)}.
\par\noindent\LINEmaths{
  \mathfrak T_{\Pioffont_{\ref{example stolzenburg 1}}}
 =
 \{\ident q(\ident a),\ident s(\ident a)\},
 \hfill
 \mathfrak T_{\Pioffont_{\ref{example stolzenburg 1}}
        \cup\Deltaoffont_{\ref{example stolzenburg 1}}}
 =
 \{\ident p(\ident a),\neg\ident p(\ident a)\}\cup
  \mathfrak T_{\Pioffont_{\ref{example stolzenburg 1}}}}.
\par\noindent We have \bigmaths{\pair{\mathcal A_1}{\neg\ident p(\ident a)}
\approx_{\rm P3}\pair{\mathcal A_2}{\ident p(\ident a)}},
because of \bigmaths
{\ident p(\ident a),\neg\ident p(\ident a)\tightnotin
 \mathfrak T_{\Pioffont_{\ref{example stolzenburg 1}}}}, and
because, \hskip.2em
for \math{H\subseteq \mathfrak T_{\Pioffont_{\ref{example stolzenburg 1}}
\cup\Deltaoffont_{\ref{example stolzenburg 1}}}}, \ 
\maths{i\in\{1,2\}}, \bigmaths{L_1:=\neg\ident p(\ident a)},
and \bigmaths{L_2:=\ident p(\ident a)},
we have the logical equivalence of \hskip.1em
\bigmaths{H=\{\ident q(\ident a)\}}{} on the one hand, \hskip.1em
and of \math H being a minimal simplified activation set for  
\nlbmath{\pair{\mathcal A_i}{L_i}} \hskip.2em
but not for \nlbmaths{\pair\emptyset{L_i}}, \
on the other hand.
By \theoref{theorem super quasi-ordering}, \hskip.2em
we also get
\bigmaths{\pair{\mathcal A_1}{\neg\ident p(\ident a)}
\approx_{\rm CP}\pair{\mathcal A_2}{\ident p(\ident a)}}.
This makes perfect sense because
\bigmaths{\ident q(\ident a)\tightund\ident s(\ident a)}{}
is not at all strictly ``more precise'' than
\nlbmaths{\ident q(\ident a)}{} in the context of 
\nlbmaths{\Pioffont_{\ref{example stolzenburg 1}}}. 
\\
Note that nothing is changed here if 
\bigmaths{\ident s(x)\antiimplies\ident q(x)}{}
is replaced by setting \maths{\Pioffont^{\rm G}_{\ref{example stolzenburg 1}}
:=\{\ident s(\ident a)\}}. \hskip.4em
If \bigmaths{\ident s(x)\antiimplies\ident q(x)}{}
is replaced, \hskip.1em
however,     \hskip.1em
by setting 
\math{\Pioffont^{\rm G}_{\ref{example stolzenburg 1}}
:=\emptyset} \hskip.2em and \hskip.1em
\maths{\Pioffont^{\rm F}_{\ref{example stolzenburg 1}}
:=\{\ident q(\ident a), \ident s(\ident a)\}}, \hskip.35em
then we \nolinebreak get both 
\bigmaths{\pair{\mathcal A_1}{\neg\ident p(\ident a)}
<_{\rm P3}\pair{\mathcal A_2}{\ident p(\ident a)}}{}
and \bigmaths{\pair{\mathcal A_1}{\neg\ident p(\ident a)}
<_{\rm CP}\pair{\mathcal A_2}{\ident p(\ident a)}}.
\halftop\end{example}
\vfill\pagebreak

\section{Efficiency Considerations}\label{section Efficiency Considerations}

\halftop\noindent
The definitions of specificity we have presented in this paper (in particular,
\defiref{DefSpec} according to \cite{Simari-Loui-Defeasible-Reasoning-1992} and \defiref{definition CP specificity}) share several
features, which we \nolinebreak will highlight in this section. 

\subsection{A minor Gain in Efficiency}\label
{subsection A Minor Gain of Efficiency}
As 
a
straightforward procedure for 
deciding
the specificity orderings
\nlbmath{\lesssim_{\rm CP}} and \nlbmath{\lesssim_{\rm P3}}
between two
arguments has to consider all possible activation sets from the literals in
the sets
\nlbmath{\mathfrak T_\Pioffont}
\nolinebreak and 
\nlbmaths{\mathfrak T_{\Pioffont\cup\Deltaoffont}}, 
respectively, the effort
for computing $\lesssim_{\rm CP}$ is lower than that of $\lesssim_{\rm P3}$ 
because of \bigmathnlb
{\mathfrak T_\Pioffont\subseteq\mathfrak T_{\Pioffont\cup\Deltaoffont}},
though not \wrt\ asymptotic complexity: \hskip.1em
in both cases 
already 
the 
number of possible (simplified) activation sets 
is exponential in the number of literals in the
respective 
sets
\nlbmath{\mathfrak T_\Pioffont}
\nolinebreak and 
\nlbmaths{\mathfrak T_{\Pioffont\cup\Deltaoffont}}, \hskip.2em
because in principle each possible subset has to be tested. \hskip.3em

\subsection{Comparing Derivations}
To lower the computational complexity,
more syntactic 
criteria
for
computing specificity 
are 
introduced in
\cite{Stolzenburg-etal-Computing-Specificity-2003}. \hskip.3em
These criteria
refer 
to the {\em derivations}\/ for the
given arguments.

\yestop\noindent
A more formal definition of the {\em and-trees}\/ 
of \nlbsectref{subsubsection Precise Isolation in And-Trees} \hskip.1em
may be appropriate here:
\begin{definition}[Derivation Tree]\label{DefTree}
\\
Let a defeasible 
specification \nlbmath{\trip{\Pioffont^{\rm F}}{\Pioffont^{\rm G}}\Deltaoffont} 
and a literal $h$ be given. 
\\
A 
{\em
derivation tree}\/ $T$ for \nlbmath h
\hskip.2em \wrt\ the specification
is a finite, rooted tree 
where all nodes are labeled with literals,
satisfying the following conditions:\begin{enumerate}\noitem\item 
The root node of $T$ is labeled with 
$h$.
  \item For each node $N$ in $T$ that is labeled with the literal \nlbmaths L, 
    there is a strict or defeasible rule 
    in \nlbmath{\Pioffont\cup\Deltaoffont}
    with head $L_0$
    and  body \nlbmaths{L_1\tightund\cdots\tightund L_k}, 
    such that $L=L_0\sigma$ for some 
    substitution $\sigma$,
    and the node $N$ has exactly $k$ 
    child
    nodes, which are
    labeled with $L_1\sigma, \dots, L_k\sigma$, respectively.
\end{enumerate}
\end{definition}

\subsubsection{No Pruning Required}\label
{subsubsection No Pruning Required}

\noindent
The
concept of pruning derivation trees is introduced
in \cite[\litdefiref{12}]{Stolzenburg-etal-Computing-Specificity-2003} in this
context, 
because, for the case of \nlbmaths{\lesssim_{\rm P2}},
attention cannot be restricted to derivations
which 
make use 
only
of the 
instances of 
defeasible rules 
given in the
arguments. The reason
for this is that the specificity notions according to
\cite{Poole-Preferring-Most-Specific-1985} and
\cite{Simari-Loui-Defeasible-Reasoning-1992} \hskip.1em
admit 
literals $L$ 
in activation sets that cannot be derived
solely by strict rules, \ie\ \nlbmaths
{L\in\mathfrak T_{\Pioffont\cup\Deltaoffont}\tightsetminus\mathfrak T_\Pioffont}.
\hskip.3em
Because  
this is not possible with the relation \nlbmaths{\lesssim_{\rm CP}}, \hskip.2em
this problem 
vanishes 
with our new version of specificity.
See also \examref{example stolzenburg 2}.

\subsubsection{Sets of derivations have to be compared}

Yet still, the new relation $\lesssim_{\rm CP}$ inherits several properties
from  $\lesssim_{\rm P3}$. 
For instance, in general the syntactic criterion requires us to
compare sets of derivations. This is true for both versions of the specificity
relation, which can be seen from \examref{example frieder 4}.

\begin{example}%
\label{example frieder 4}%
\\[-3ex]\math{\begin{array}[t]{@{}l l l@{}}
\\\Pioffont^{\rm F}_{\ref{example frieder 4}}
 &:=
 &\left\{\begin{array}{l}\ident a,
   \\\ident e
   \\\end{array}\right\}, 
\\\Pioffont^{\rm G}_{\ref{example frieder 4}}
 &:=
 &\left\{\begin{array}{l}\ident d\antiimplies\ident c_1,
   \\\ident d\antiimplies\ident c_2,
   \\\ident c_1\antiimplies\ident b,
   \\\ident c_2\antiimplies\ident b
   \\\end{array}\right\}, 
\\\Deltaoffont_{\ref{example frieder 4}}
 &:=
 &\left\{\begin{array}{l}
     \ident b\defeasibleantiimplies\ident a,
   \\\neg\ident f\defeasibleantiimplies\ident d,
   \\\ident f\defeasibleantiimplies\ident d\tightund\ident e
   \\\end{array}\right\},
\\\mathcal A^1
 &:=
 &\left\{\begin{array}{l}\ident f\defeasibleantiimplies\ident d\tightund\ident e,
   \\\ident b\defeasibleantiimplies\ident a
   \\\end{array}\right\},
\\\mathcal A^2
 &:=
 &\left\{\begin{array}{l}\neg\ident f\defeasibleantiimplies\ident d,
   \\\ident b\defeasibleantiimplies\ident a
   \\\end{array}\right\}.
\\\end{array}}\hfill
\begin{tikzpicture}[baseline=(current bounding box.north),
>=stealth,->,looseness=.5,auto]
\matrix [matrix of math nodes,
column sep={1.4cm,between origins},
row sep={1.3cm,between origins}]{
 & |(f)|     \math{\ident f} 
&& |(notf)|  \math{\neg\ident f} 
\\ 
&& |(d1)|    \math{\ident d}
&& |(d2)|    \math{\ident d}
\\ |(e)|     \math{\ident e}
&& |(c1)|    \math{\ident c_1}
&& |(c2)|    \math{\ident c_2}
\\ 
 &
&& |(b)|     \math{\ident b}
\\ 
 & |(TRUE)|  \math{\ident{TRUE}}
&& |(a)|     \math{\ident a}
\\
};
\begin{scope}[every node/.style={midway,auto}]
\draw [] (e) -- node [right] 
      {$\scriptscriptstyle\!\!\mathcal A^1$} 
      (f);
\draw [] (d1) -- node [right] 
      {$\scriptscriptstyle\!\mathcal A^1$} 
      (f);
\draw [] (d1) -- node [right] 
      {$\scriptscriptstyle\!\mathcal A^2$} 
      (notf);
\draw [] (a) -- node [right] 
      {$\scriptscriptstyle\!\!\mathcal A^1\cap\mathcal A^2$} 
      (b);
\draw [double] (b) -- (c1);
\draw [double] (b) -- (c2);
\draw [double] (c1) -- (d1);
\draw [double] (c2) -- (d2);
\draw [double] (TRUE) -- (e);
\draw [double] (TRUE) -- (a);
\draw [double, -] (d1) -- (d2);
\end{scope}
\end{tikzpicture}
\par\noindent We have \bigmaths{\pair{\mathcal A^1}{\ident f}<_{\rm P3}
\pair{\mathcal A^2}{\neg\ident f}}{}
and \bigmaths{\pair{\mathcal A^1}{\ident f}<_{\rm CP}
\pair{\mathcal A^2}{\neg\ident f}}.
\\There are two derivation and-trees for each argument. 
Since either \math{\ident c_1} or \math{\ident c_2} is in the derivation tree, 
a one-to-one comparison of derivations does not 
suffice.
\end{example}

\yestop\noindent
The reason for this complication is that we consider a very general setting of defeasible 
reasoning in this paper, because we admit\begin{enumerate}\noitem\item 
more than one antecedent in rules, \ie\ bodies containing more than
    one (possibly negative) literal, and\noitem\item 
(possibly) non-empty sets of background knowledge, \ie\ strict rules,
    not only facts.\noitem\end{enumerate}
In the literature, often restricted cases are considered only: antecedents are
always singletons in \cite{Gelfond_Przymusinska_1990}, 
no background knowledge is allowed in
\cite{DS96}, and both restrictions are present in \cite{BG97b}. 

\yestop
\subsubsection{Path Criteria?}\label
{subsubsection Path Criteria}

As the computation of activations sets via goal-directed derivations
from \cite[\litsectref{3.3}]{Stolzenburg-etal-Computing-Specificity-2003}
is hardly changed by our step from \nlbmath{\lesssim_{\rm P3}}
to \nlbmaths{\lesssim_{\rm CP}}, \hskip.2em
let us have a closer look here only at the path criterion
of \cite[\litsectref{3.4}]{Stolzenburg-etal-Computing-Specificity-2003}.

\begin{definition}[Path]
For a leaf node 
\nlbmath N
in a derivation tree 
\nlbmaths T, 
we define  the {\em path}\/ in \nlbmath T
through $N$ as 
the empty set if \math N is the root, and otherwise as
the set consisting of the literal labeling 
\nlbmaths N, 
together with all
literals labeling its ancestors except the root node.
Let ${\sf Paths}(T)$ be the
set of all paths in $T$ through all leaf nodes $N$.
\end{definition}

\noindent
With this notion of paths, 
the quasi-ordering $\unlhd$ on and-trees
can be given as follows:

\begin{definition}
[{\cite[\litdefiref{23}]{Stolzenburg-etal-Computing-Specificity-2003}}]
\\
$T_1 \unlhd T_2$ \udiff\ 
$T_1$ and $T_2$ are two derivation trees, and  
for each $t_2\in{\sf Paths}(T_2)$ there 
is 
a path $t_1 \in {\sf Paths}(T_1)$ such that $t_1
\subseteq t_2$.
\end{definition}

\noindent
Two derivation trees can be compared \wrt\ $\unlhd$ 
efficiently. 
This requires the pairwise comparison of all
nodes in the trees for each path. Hence, the respective complexity is
polynomial, at most ${\rm O}(n^3)$, where $n$ \hskip.1em
is the overall number of nodes, which
makes the relation \nlbmath\unlhd\ attractive for practical use.

\begin{definition}
[{\cite[\litdefiref{24}]{Stolzenburg-etal-Computing-Specificity-2003}}]%
\label{better}\\
$(\mathcal A_1,h_1) \le (\mathcal A_2,h_2)$ \udiff\ 
$(\mathcal A_1,h_1)$ and $(\mathcal A_2,h_2)$ are two arguments in the given
specification and for
each derivation tree 
$T_1$ for $h_1$ there is a derivation tree
$T_2$ for $h_2$ such that $T_1 \unlhd T_2$.
\end{definition}

\noindent
It is shown in \cite[Theorem~25]{Stolzenburg-etal-Computing-Specificity-2003}
that in special cases, \math\leq\ and \nlbmath{\lesssim_{\rm P2}} are equivalent,
namely
if the arguments involved in the comparison correspond to exactly one derivation
tree. 
Let us try to adapt this result to our new relation
\nlbmaths{\lesssim_{\rm CP}},
in the sense that we try to establish a mutual subset relation
between \math\leq\ and \nlbmath{\lesssim_{\rm CP}}.

The forward direction is pretty straightforward,
but comes with the restriction to be expected:
From \cite[Theorem~25]{Stolzenburg-etal-Computing-Specificity-2003} \hskip.2em
we get \bigmaths{
\tight\leq\nottight{\nottight\subseteq}\tight{\lesssim_{\rm P2}}}.
By looking at the empty path,
we easily see that \nlbmath\leq\
satisfies the addition restriction of 
\defiref{DefSpec} as compared to \defiref{DefSpec2}; \hskip.2em
so we also get \bigmaths{
\tight\leq\nottight{\nottight\subseteq}\tight{\lesssim_{\rm P3}}}.
Finally, 
we can apply \theoref{theorem super quasi-ordering}
and so get the intended \bigmaths{
\tight\leq\nottight{\nottight\subseteq}\tight{\lesssim_{\rm CP}}},
but only with the strong restriction of the condition of 
\theoref{theorem super quasi-ordering}. \hskip.2em
We see no way yet to relax this restriction resulting from phase\,3
of \sectref{section where the phases are}.

It is even more unfortunate that
the backward direction does not hold at all
because of our change in phase\,1 of \sectref{section where the phases are}.
\hskip.3em
In particular, 
as shown in \examref{example failure}, \hskip.1em
it does not hold for the special case 
where it holds for \nlbmaths{\lesssim_{\rm P2}}, \hskip.2em
\ie\ in the case
that there are no general strict rules
and hence each argument corresponds to exactly
one derivation (\cfnlb\ the proof of \littheoref{25} in
\cite{Stolzenburg-etal-Computing-Specificity-2003}). 
\begin{example}\label{example failure}
\\[-3ex]\math{\begin{array}[t]{@{}l l l@{}}
\\\Pioffont^{\rm F}_{\ref{example failure}}
 &:=
 &\left\{\begin{array}{l}\ident a,
   \\\ident b
   \\\end{array}\right\}, 
\\\Pioffont^{\rm G}_{\ref{example failure}}
 &:=
 &\emptyset,
\\\Deltaoffont_{\ref{example failure}}
 &:=
 &\mathcal A^1_{\ref{example failure}}\cup\mathcal A^2_{\ref{example failure}}
\\\mathcal A^1_{\ref{example failure}}
 &:=
 &\left\{\begin{array}{l}
     \ident c_1\defeasibleantiimplies\ident a\tightund\ident b,
   \\\ident d\defeasibleantiimplies\ident c_1
   \\\end{array}\right\},
\\\mathcal A^2_{\ref{example failure}}
 &:=
 &\left\{\begin{array}{l}
     \ident c_2\defeasibleantiimplies\ident a,
   \\\neg\ident d\defeasibleantiimplies\ident c_2
   \\\end{array}\right\}.
\\\end{array}}\hfill\begin{tabular}[t]{@{}r@{}}
\begin{tikzpicture}[baseline=(current bounding box.north),
>=stealth,->,looseness=.5,auto]
\matrix [matrix of math nodes,
column sep={1.4cm,between origins},
row sep={1.3cm,between origins}]{
   |(negd)|  \math{\neg\ident d}
&& |(d)|     \math{\ident d} 
\\ |(c2)|    \math{\ident c_2}
&& |(c1)|    \math{\ident c_1}
\\ |(a)|     \math{\ident a}
&& |(TRUE)|  \math{\ident{TRUE}}
&& |(b)|     \math{\ident b}
\\   
};
\begin{scope}[every node/.style={midway,auto}]
\draw [] (c2) -- node [left] 
      {$\scriptscriptstyle\mathcal A^2_{\ref{example failure}}\!\!$} 
      (negd);
\draw [] (a) -- node [left] 
      {$\scriptscriptstyle\mathcal A^2_{\ref{example failure}}\!\!$} 
      (c2);
\draw [] (c1) -- node [left] 
      {$\scriptscriptstyle\mathcal A^1_{\ref{example failure}}\!\!$} 
      (d);
\draw [] (a) -- node [right] 
      {$\scriptscriptstyle\,\mathcal A^1_{\ref{example failure}}$} 
      (c1);
\draw [] (b) -- node [left] 
      {$\scriptscriptstyle\mathcal A^1_{\ref{example failure}}\,$} 
      (c1);
\draw [double] (TRUE) -- (a);
\draw [double] (TRUE) -- (b);
\end{scope}
\end{tikzpicture}
\\\end{tabular}
\par\noindent 
We have 
\bigmaths{\pair{\mathcal A^1_{\ref{example failure}}}{\ident d}
\vartriangle_{\rm P3}\pair{\mathcal A^2_{\ref{example failure}}}{\neg\ident d}}{}
and 
\bigmaths{\pair{\mathcal A^1_{\ref{example failure}}}{\ident d}
<_{\rm CP}\pair{\mathcal A^2_{\ref{example failure}}}{\neg\ident d}}.
Both arguments \pair{\mathcal A^1_{\ref{example failure}}}{\ident d}
and \pair{\mathcal A^2_{\ref{example failure}}}{\neg\ident d}
correspond to exactly one derivation tree,
say \nlbmath{T_1} and \nlbmaths{T_2}, respectively. 
All paths in \math{{\sf Paths}(T_1)} contain \nlbmaths{\ident c_1}, 
but not \nlbmaths{\ident c_2}, 
and all paths in \math{{\sf Paths}(T_2)} contain 
\nlbmaths{\ident c_2}, 
but not \nlbmaths{\ident c_1}. 
Hence, 
\bigmaths{\pair{\mathcal A^1_{\ref{example failure}}}{\ident d}
\leq\pair{\mathcal A^2_{\ref{example failure}}}{\neg\ident d}}{}
does not hold.
\end{example}

\clearpage
\section{Conclusion}\label
{section Conclusion}%
We would need further discussions on our 
utmost surprising new findings 
---~after all, 
defeasible reasoning with \poole's notion of specificity 
is being applied now for over a quarter of century,
and it was not to be expected that our 
investigations 
could shake an element of the field to the very foundations.

 One remedy for the discovered lack of transitivity 
 of \nolinebreak\hskip.1em\nlbmath{\lesssim_{\rm P3}}
 could be to consider the 
 transitive closure of the non-transitive relation \nlbmath{\lesssim_{\rm P3}}.
 \hskip.3em
 Only under the condition of \theoref{theorem super quasi-ordering}, \hskip.2em
 the transitive closure of 
 \nolinebreak\hskip.2em\nlbmath{\lesssim_{\rm P3}} \hskip.1em
 is a subset of 
 \nolinebreak\hskip.2em\nlbmath{\lesssim_{\rm CP}}, \hskip.2em
 and therefore a possible choice. \hskip.2em
 Moreover, it will still have all the intuitive shortcomings 
 made obvious in \sectref{section Putting Specificity to Test}. \hskip.3em
 We do not see how this transitive closure could be 
 decided efficiently. \hskip.2em
 Furthermore, 
 this transitive closure lacks a direct intuitive motivation,
 and after the first extension step from \nlbmath{\lesssim_{\rm P3}} 
 to its transitive closure, \hskip.1em 
 we had better take the 
 second extension step to the more intuitive 
 \nlbmath{\lesssim_{\rm CP}} immediately.

Finally,
contrary to the transitive closure of \maths{\lesssim_{\rm P3}}, \hskip.1em
our novel relation \nlbmath{\lesssim_{\rm CP}} 
also solves the problem of 
non-monotonicity of specificity \wrt\ conjunction
(\cfnlb\ \sectref{subsection Monotonicity of Preference}), \hskip.1em
which was already realized as a problem
of \math{\lesssim_{\rm P1}}
by \citet{Poole-Preferring-Most-Specific-1985}
(\cfnlb\ \examref{example poole 6}).

The present means to decide our novel specificity relation 
\nlbmaths{\lesssim_{\rm CP}}, \hskip.2em
however, \hskip.1em
show improvements 
(\cfnlb\ \sectrefs{subsection A Minor Gain of Efficiency}
 {subsubsection No Pruning Required})
and setbacks
(\cfnlb\ \sectref{subsubsection Path Criteria})
if compared to the known ones for \poole's relation.
Further work is needed to improve efficiency considerably.
Our plan is to narrow 
the concept of
\nlbmath{\lesssim_{\rm CP}} \hskip.1em
further down
according to 
\sectref{section Toward an Intuitive Notion of Specificity}, \hskip.1em
and then to develop strong methods 
for efficient operationalization,
which can hardly be found 
for any of 
\nlbmaths{\lesssim_{\rm P1}},
\nlbmaths{\lesssim_{\rm P2}},
\nlbmaths{\lesssim_{\rm P3}},
\nlbmath{\lesssim_{\rm CP}}.

It is just too early for a further conclusion,
and the further implications of the contributions of this paper
and the technical details of the operationalization of 
our correction of \poole's specificity
will have to be discussed in future work.

\section*{Acknowledgments}

\begin{sloppypar}%
This research has been supported by the DFG 
(German Research Community) grant 
\mbox{Sto\,421/5-1.}\par\end{sloppypar}

\halftop\noindent
To honor \poolename,
let us end this paper with the last sentence of
\cite{Poole-Preferring-Most-Specific-1985}:

This research was sponsored by no 
defence 
department.
\vfill\pagebreak

\nocite{writing-mathematics}
\addcontentsline{toc}{section}{References}
\bibliography{specificity}
\end{document}